\documentclass[review]{elsarticle}
\graphicspath{ {./figures/} }
\usepackage{hyperref}
\usepackage{float}
\usepackage{verbatim} 
\usepackage{apalike}
\restylefloat{figure}

\graphicspath{ {./figures/} }
\usepackage[table]{xcolor}
\usepackage{verbatim} 
\usepackage{apalike}
\usepackage{caption}
\usepackage{subcaption}
\usepackage[margin=1in]{geometry} 
\usepackage[]{graphicx}
\usepackage[font=small,skip=0pt]{caption}
\usepackage{enumitem}
\usepackage{booktabs}
\usepackage{longtable}
\usepackage{tabularx}
\usepackage{cancel}
\usepackage{multirow}
\usepackage{supertabular}
\usepackage{algorithmic}
\usepackage{algorithm}
\usepackage{amsmath}
\usepackage{rotating}
\usepackage[normalem]{ulem}
\usepackage[utf8]{inputenc}
\usepackage[T1]{fontenc}
\usepackage{lineno}
\usepackage{array}
\usepackage{makecell}
\usepackage{pdflscape}
\usepackage{afterpage}
\usepackage{capt-of}
\usepackage{lipsum}
\usepackage{changepage}
\usepackage{xcolor}
\usepackage{rotating} 
\usepackage{graphicx}
\newcommand{\blue}[1]{\textcolor{blue}{#1}}
\usepackage{amssymb}
\usepackage{natbib}
\bibpunct{(}{)}{,}{a}{}{;}
\renewcommand{\citep}[1]{(\citealp{#1})}

\journal{Expert Systems with Applications}

\biboptions{authoryear}

\begin{document}
\begin{frontmatter}

\begin{titlepage}
\begin{center}
\vspace*{1cm}

\textbf{ \large What is the  best RNN-cell structure to forecast each time series behavior?}

\vspace{1.5cm}

Rohaifa Khaldi$^{a}$ (rohaifa@ugr.es), Abdellatif El Afia$^b$ (abdellatif.elafia@ensias.um5.ac.ma), Raddouane Chiheb$^b$ (raddouane.chiheb@ensias.um5.ac.ma), Siham Tabik$^a$ (siham@ugr.es) \\

\hspace{10pt}

\begin{flushleft}
\small  
$^a$ Dept. of Computer Science and Artificial Intelligence, Andalusian Research Institute in Data Science and Computational Intelligence, DaSCI, University of Granada, 18071, Granada, Spain \\
$^b$ ENSIAS, Mohammed V University of Rabat, 10170, Rabat, Morocco

\vspace{1cm}
\textbf{Corresponding Author:} \\
Rohaifa Khaldi \\
Dept. of Computer Science and Artificial Intelligence, Andalusian Research Institute in Data Science and Computational Intelligence, DaSCI, University of Granada, 18071, Granada, Spain \\
Email: rohaifa@ugr.es

\end{flushleft}        
\end{center}
\end{titlepage}

\title{What is the  best RNN-cell structure to forecast each time series behavior?}
\author[label1]{Rohaifa Khaldi \corref{cor1}}
\ead{rohaifa@ugr.es}
\author[label2]{Abdellatif El Afia}
\ead{abdellatif.elafia@ensias.um5.ac.ma}
\author[label2]{Raddouane Chiheb}
\ead{raddouane.chiheb@ensias.um5.ac.ma}
\author[label1]{Siham Tabik}
\ead{siham@ugr.es}

\cortext[cor1]{Corresponding author.}
\address[label1]{Dept. of Computer Science and Artificial Intelligence, Andalusian Research Institute in Data Science and Computational Intelligence, DaSCI, University of Granada, 18071, Granada, Spain}
\address[label2]{ENSIAS, Mohammed V University of Rabat, 10170, Rabat, Morocco}

\begin{abstract}
It is unquestionable that time series forecasting is of paramount importance in many fields. The most used machine learning models to address time series forecasting tasks are Recurrent Neural Networks (RNNs). Typically, those models are built using one of the three most popular cells, ELMAN, Long–Short Term Memory (LSTM), or Gated Recurrent Unit (GRU) cells, each cell has a different structure and implies a different computational cost. However, it is not clear why and when to use each RNN-cell structure. Actually, there
is no comprehensive characterization of all the possible time series behaviors and no guidance on what RNN cell structure is the most suitable for each behavior. The objective of this study is two-fold: it presents a comprehensive taxonomy of almost all-time series behaviors (deterministic, random-walk, nonlinear, longmemory, and chaotic), and provides insights into the best RNN cell structure for each time series behavior.
We conducted two experiments: (1) The first experiment evaluates and analyzes the role of each component in the LSTM-Vanilla cell by creating 11 variants based on one alteration in its basic architecture (removing,
adding, or substituting one cell component). (2) The second experiment evaluates and analyzes the performance of 20 possible RNN-cell structures. To evaluate, compare, and select the best model, different statistical metrics were used: error-based metrics, information criterion-based metrics, naïve-based metric, and direction
change-based metric. To further improve our confidence in the models’ interpretation and selection, Friedman Wilcoxon–Holm signed-rank test was used.

Our results advocate the usage and the exploration of the newly created RNN variant, named SLIM, in
time series forecasting thanks to its high ability to accurately predict the different time series behaviors as well as its simple structural design that does not require expensive temporal and computing resources.

\end{abstract}


\end{frontmatter}

\section{Introduction}
\label{introduction}
Many real-world prediction problems involve a temporal dimension and  typically require the estimation of numerical sequential data referred to as time series forecasting. Time series forecasting is one of the major stones in data science playing a pivotal role in almost all domains, including meteorology \citep{murat2018forecasting}, natural disasters control \citep{erdelj2017wireless}, energy \citep{bourdeau2019modeling}, manufacturing \citep{wang2018direct}, finance \citep{liu2019novel}, econometrics \citep{siami2018forecasting}, telecommunication \citep{maeng2020demand}, healthcare \citep{khaldii2019forecasting} to name a few. 
Accurate time series forecasting requires robust forecasting models.

Currently, Recurrent Neural Network (RNN) models are one of the most popular machine learning models in sequential data modeling, including natural language, image/video captioning, and forecasting \citep{sutskever2014sequence, vinyals2015show, chimmula2020time}. 
Such RNN models are built as a sequence of the same cell structure, for example, ELMAN cell, Long-Short Term Memory (LSTM) cell or Gated Recurrent Unit (GRU) cell. The simplest RNN cell is ELMAN, it includes one layer of hidden neurons. While, LSTM and GRU cells incorporate a gating mechanism, three gates in LSTM and two gates in GRU, where each gate is a layer of hidden neurons.
Many other cell structures have been introduced in the literature \citep{zhou2016minimal, lu2017simplified, mikolov2014learning, pulver2017lstm}. However, to solve time series forecasting tasks, the building of RNN models is typically limited to the three aforementioned cell structures \citep{sezer2020financial, runge2021review, liu2021intelligent, rajagukguk2020review, alkhayat2021review}, as they provide very good accuracy \citep{runge2021review,sezer2020financial}.

Nevertheless, building  robust RNN models for time series forecasting  is still a  challenging task as there does not exist yet a clear  understanding of times series data itself and hence there exist very little knowledge about what cell structure is the most appropriate for each data type. 
In general, when facing a new problem,  practitioners select one of the most popular cells, usually LSTM, and use it as a building block for the RNN model without any guarantee on the appropriateness of this cell to the current data.
The objective of this work  is two-fold. It presents a comprehensive characterization of time series behaviors and provides guidelines on the best RNN cell structure for each behavior. As far as we know, this is the first work providing such insights.
The main contributions of this study can be summarized as follows:
\begin{itemize}
\setcounter{enumi}{0}

\item Provide a better understanding of times series data by presenting a comprehensive characterization of their behaviors. 

\item Determine the most appropriate cell structure for each time series behavior (i.e., whether a specific cell structure should be avoided for certain behaviors).

\item Identify differences in predictability between behaviors (i.e., whether certain behaviors are easier or harder to predict across all cell models).

\item Provide useful guidelines that can assist decision-makers and scholars in the process of  selecting  the most suitable RNN-cell structure from both, a computational and performance point of view.
\end{itemize}

The remainder of this study is organized as follows: Section 2 states the related works. Section 3 presents a taxonomy of time series behaviors.  Section 4 presents a taxonomy of RNN cells. Section 5 describes the experiment. Section 6 exhibits and discusses the obtained results. Finally, the last section concludes the findings and spots light on future research directions.

\section{Related works}

The last decades have known an explosion of time series data acquired by automated data collection devices such as monitors, IoT devices, and sensors \citep{murat2018forecasting, erdelj2017wireless, bourdeau2019modeling}.
The collected time series describes different quantitative values: stock price, amount of sales, electricity load demand, weather temperature, etc.
In parallel, a large number of comparative studies have been carried out in the forecasting area \citep{parmezan2019evaluation, godahewa2021monash, athiyarath2020comparative, divina2019comparative, choubin2018precipitation, sagheer2019time, bianchi2017overview}. These studies can be divided into two categories:  (1) the first category tries to find a universal forecasting model for any field \citep{parmezan2019evaluation, godahewa2021monash, athiyarath2020comparative}. They  compare multiple models on a set of datasets from different fields to conclude which model is the universal predictor. The selection of the used datasets is not based on any clear criteria. Whereas, (2) the second category focuses on selecting the most performing forecasting model for a specific field \citep{divina2019comparative, choubin2018precipitation, sagheer2019time, bianchi2017overview}.
They compare a set of models on one or multiple datasets coming from the same field. Nevertheless, the best predictor does not ensure stable performance over different datasets even if they come from the same field. Actually, after an extensive analysis of highly diverse datasets, \citep{keogh2003need} demonstrated that there is a need for more comprehensive time series benchmarks and more careful evaluations in the data mining community. In addition, datasets should have a large size to train and test the models and incorporate specific behaviors that challenge their modeling. Such knowledge of the dataset properties is required to facilitate a better interpretation of the modeling results.

The natural approach to create such benchmarking time series datasets is to collect data from real applications. For instance, the NN5 dataset \citep{crone2008nn5}, the CIF 2016 dataset \citep{godahewa_rakshitha_2020_3904073}, the M4 dataset \citep{makridakis2018m4}, and the Monash archive that gather 20 publicly available time series datasets \citep{godahewa2021monash}. Although, real time series are always business-oriented which make them either proprietorial or expensive to obtain \citep{dau2019ucr}, they can take decades to become mature and ready to be used for machine learning purposes, their diversity testing is tedious \citep{dau2019ucr, spiliotis2020forecasting}, and most importantly, their Data Generation Processes (DGPs) are unknown which make the interpretation of the models and the explanation of their decisions challenging.

An alternative solution is to generate synthetic time series datasets with known embedded patterns \citep{olson2017pmlb}. For instance, \citep{zhang2005neural} investigated the issue of how to effectively model artificial time series with deterministic behavior due to the existence of trend and seasonality using Artificial Neural Networks (ANNs). \citep{lopez2016mackey} examined ANNs on time series with noiseless and noisy chaotic behavior generated by Mackey-Glass series. \citep{li2016new} applied the Self-Constructing Fuzzy Neural Network (SCFNN) on chaotic time series including Logistic and Henon data. \citep{yeo2017model} evaluated the performance of LSTM on three different time series with chaotic behavior (delay-time chaotic dynamical systems, Mackey-Glass and Ikeda equations). \citep{fischer2018machine} presented an experimental evaluation of seven machine learning models applied on a set of eight DGPs reflecting linear and nonlinear behaviors. \citep{parmezan2019evaluation} used 40 synthetic datasets of deterministic, stochastic, and chaotic time series to compare eleven parametric and non-parametric models. \citep{kang2020gratis} used the Mixture Auto-Regressive (MAR) models to create the GRATIS dataset based on which they compared different statistical models. 
However, all the aforementioned studies remain non-comprehensive of the main time series behaviors that can be faced in real datasets. 

Another categorization of these comparative studies can be made based on the types of the evaluated models. Here, three categories can be set: (1) Studies based on parametric models where scientists try to evaluate the forecasting performance of different statistical models \citep{godahewa2021monash, yu2020hybrid, kim2018time}. (2) Studies based on non-parametric models where they assess the performance of machine learning models \citep{granata2019evapotranspiration, dudek2016neural, sagheer2019time}. (3) Studies based on parametric and non-parametric models where both types of models are compared \citep{parmezan2019evaluation, khaldi2019forecasting, yamak2019comparison, siami2018forecasting}. RNNs are one of the most used machine learning models in time series forecasting \citep{sezer2020financial}. Nevertheless, their usage is limited to three RNN variants (ELMAN, LSTM, and GRU). In the financial field,  \citep{sezer2020financial} reported that from 2005 to 2019, 52.5\% of publications used RNN models to perform time series forecasting, where the LSTM model represents 60.4\%, ELMAN (vanilla RNN) represents 29.7\%, and GRU represents 9.89\%. In the energy field, \citep{runge2021review} stated in their review that ELMAN, GRU, and LSTM are the main applied deep learning models to building energy forecasting. In the environmental field, \citep{liu2021intelligent} asserted in their review, from 2015 to 2020, that LSTM and GRU are the most practical RNN models in air quality forecasting. In the renewable energy field, \citep{rajagukguk2020review} reviewed, from 2005 to 2020, that in photovoltaic power forecasting, 60\% of publications used LSTM, 20\% used ELMAN, and 13\% used GRU. While, in the solar irradiance forecasting, they reported that LSTM represents 44\% of the used deep learning models, ELMAN represents 25\% and GRU 19\%. Similarly, \citep{alkhayat2021review} reported in their review, from 2016 to 2020, that ELMAN, LSTM, and GRU are the only RNN models used in wind and solar energy forecasting. They outlined that the usage of RNNs within this field increased from 2\% in 2016 to 25\% in 2020.

Therefore, there is a strong need for a comprehensive analysis of different types of RNN models, including the above three variants, in forecasting the main time series behaviors, and a strong need for an RNN-based model guide to assist practitioners in their process of selection and structure of the best RNN cell for each time series behavior. 


\section{Taxonomy of times series behaviors} \label{S3}

As far as we know, this is the first work introducing a complete formal characterization of real-world time series. Time series emerging from real-world applications can either follow a stochastic mechanism or a chaotic mechanism and are usually contaminated by white noise \citep{wales1991calculating, cencini2000chaos, zunino2012distinguishing, box2015time, boaretto2021discriminating}.

\subsection{Stochastic behavior}
\begin{figure}[H]
     \centering
     \begin{subfigure}[b]{0.3\textwidth}
         \centering
         \includegraphics[height=3.5cm,width=6cm]{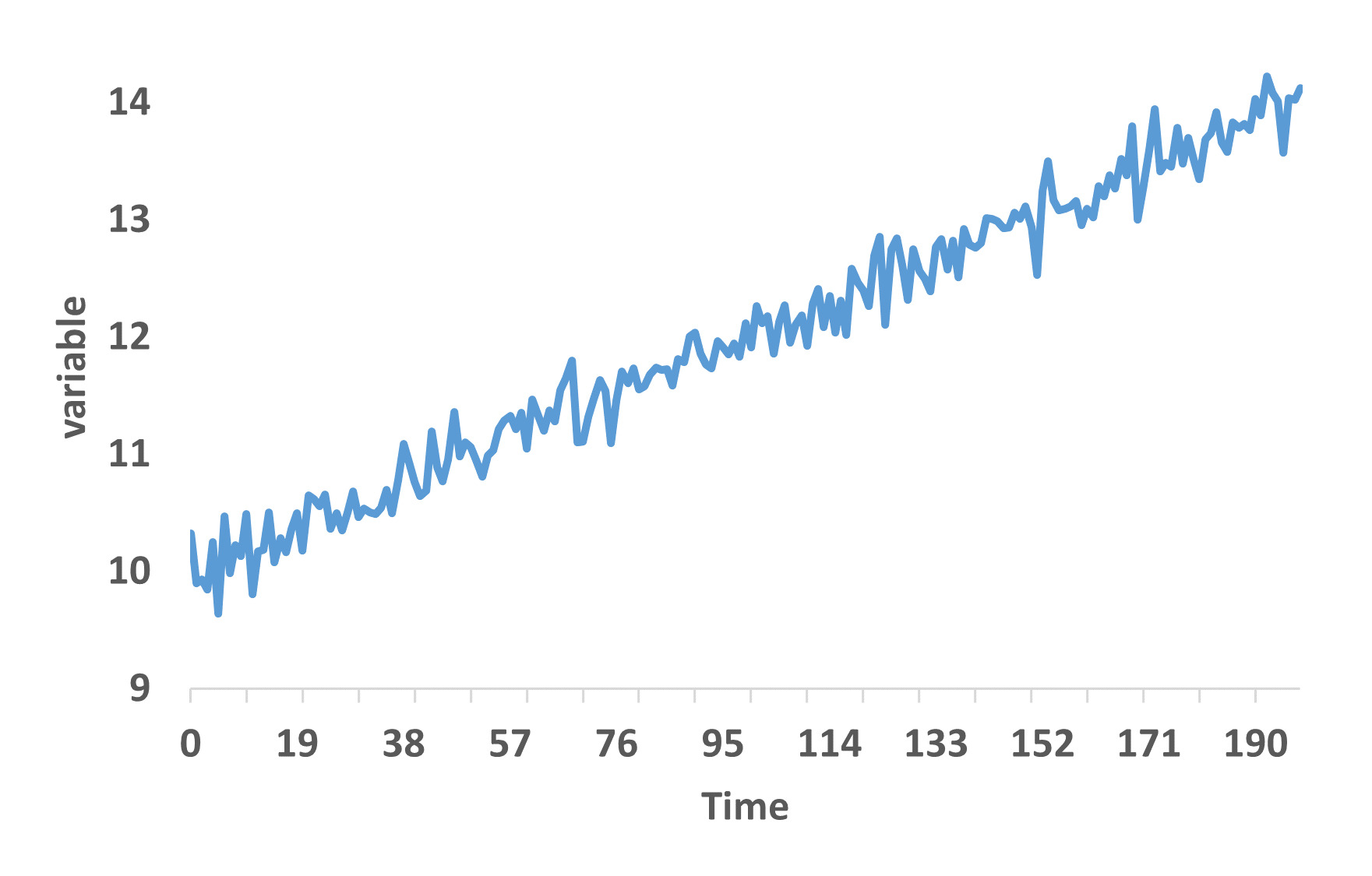}
         \caption{}
         \label{T}
     \end{subfigure}
     \hfill
     \begin{subfigure}[b]{0.3\textwidth}
         \centering
         \includegraphics[height=3.5cm,width=6cm]{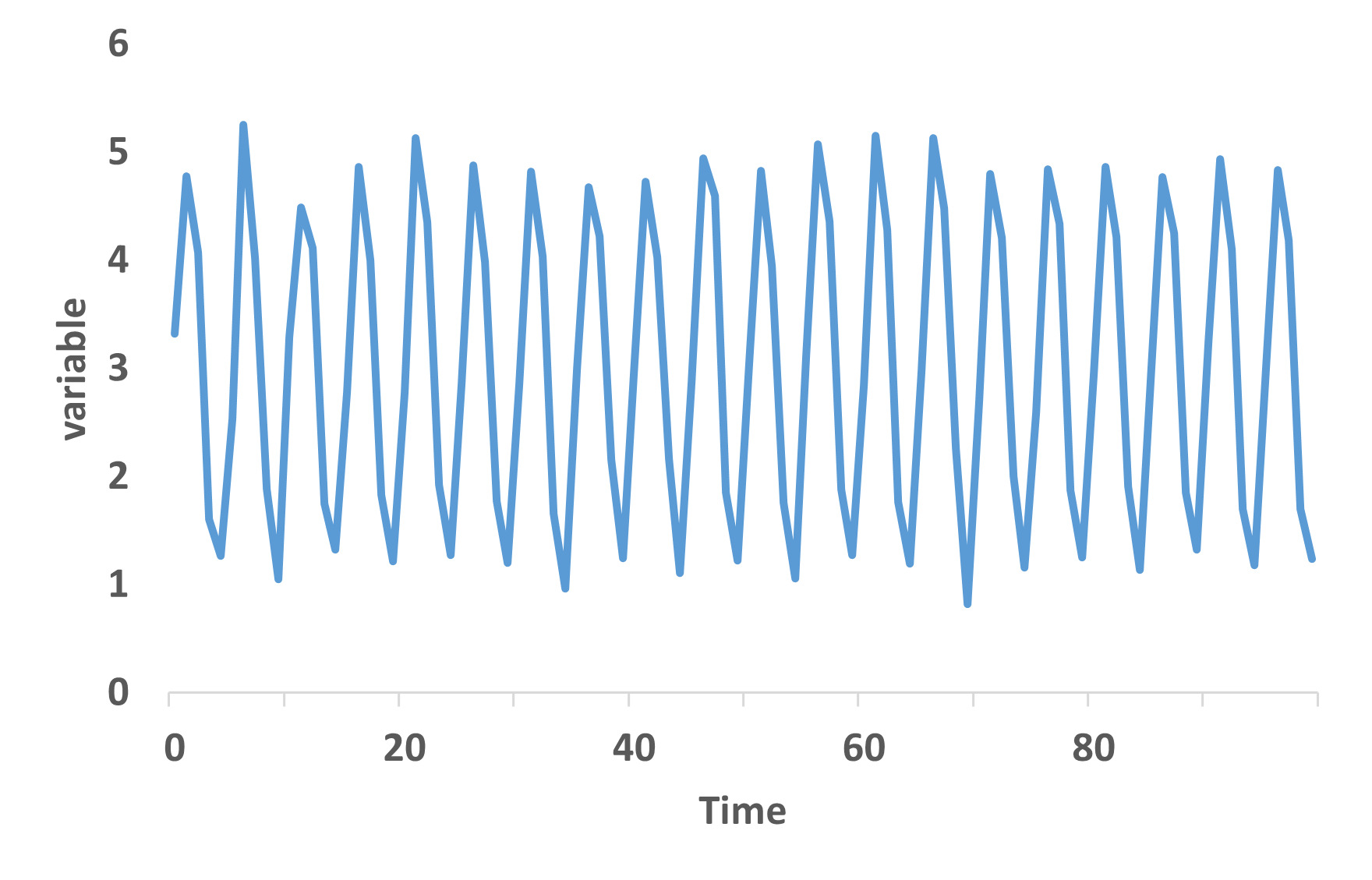}
         \caption{}
         \label{SS}
     \end{subfigure}
     \hfill
     \begin{subfigure}[b]{0.3\textwidth}
         \centering
         \includegraphics[height=3.5cm,width=6cm]{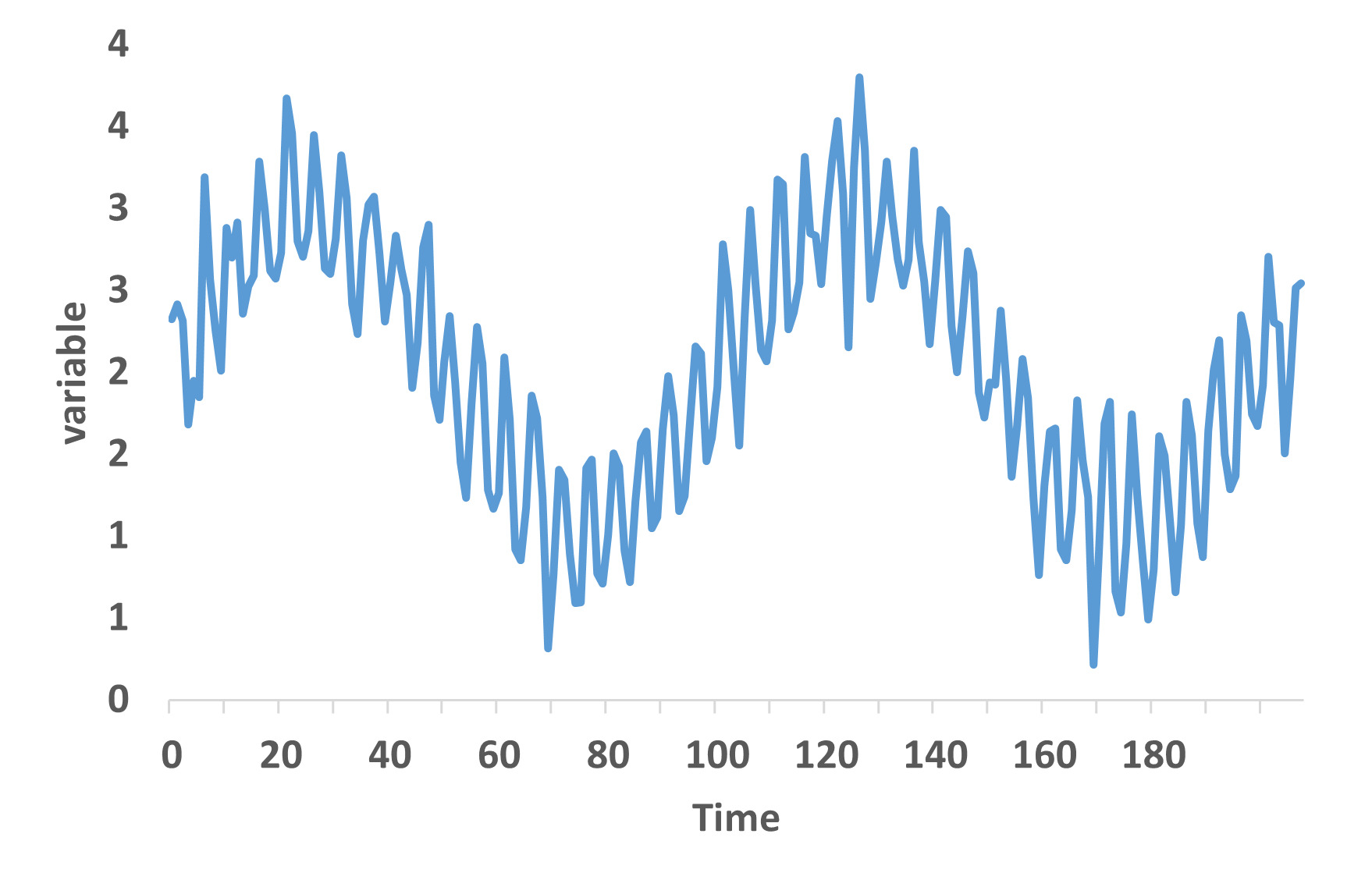}
         \caption{}
         \label{CS}
     \end{subfigure}
     \hfill
     \begin{subfigure}[b]{0.3\textwidth}
         \centering
         \includegraphics[height=3.5cm,width=6cm]{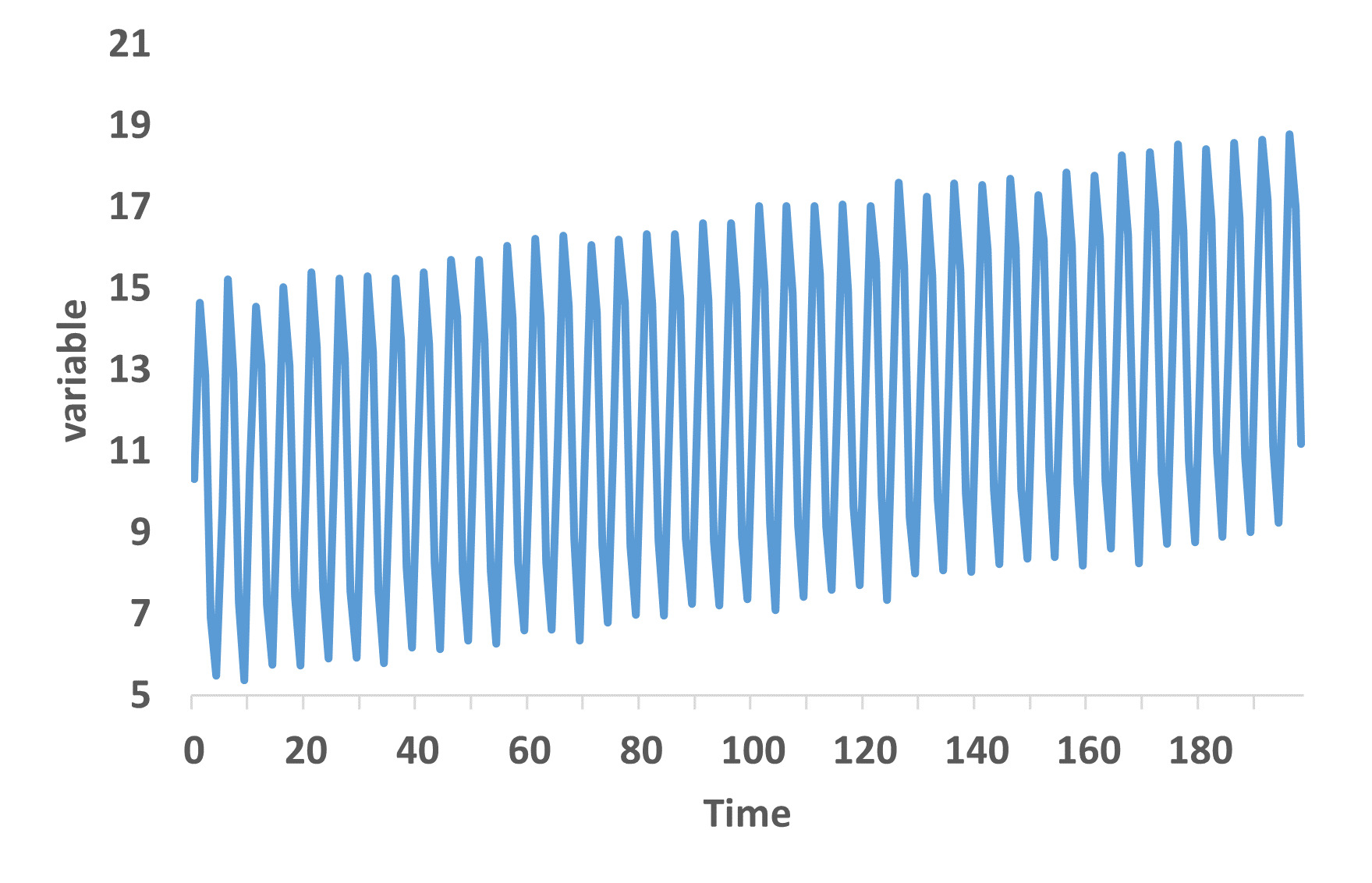}
         \caption{}
         \label{TSS}
     \end{subfigure}
     \begin{subfigure}[b]{0.3\textwidth}
         \centering
         \includegraphics[height=3.5cm,width=6cm]{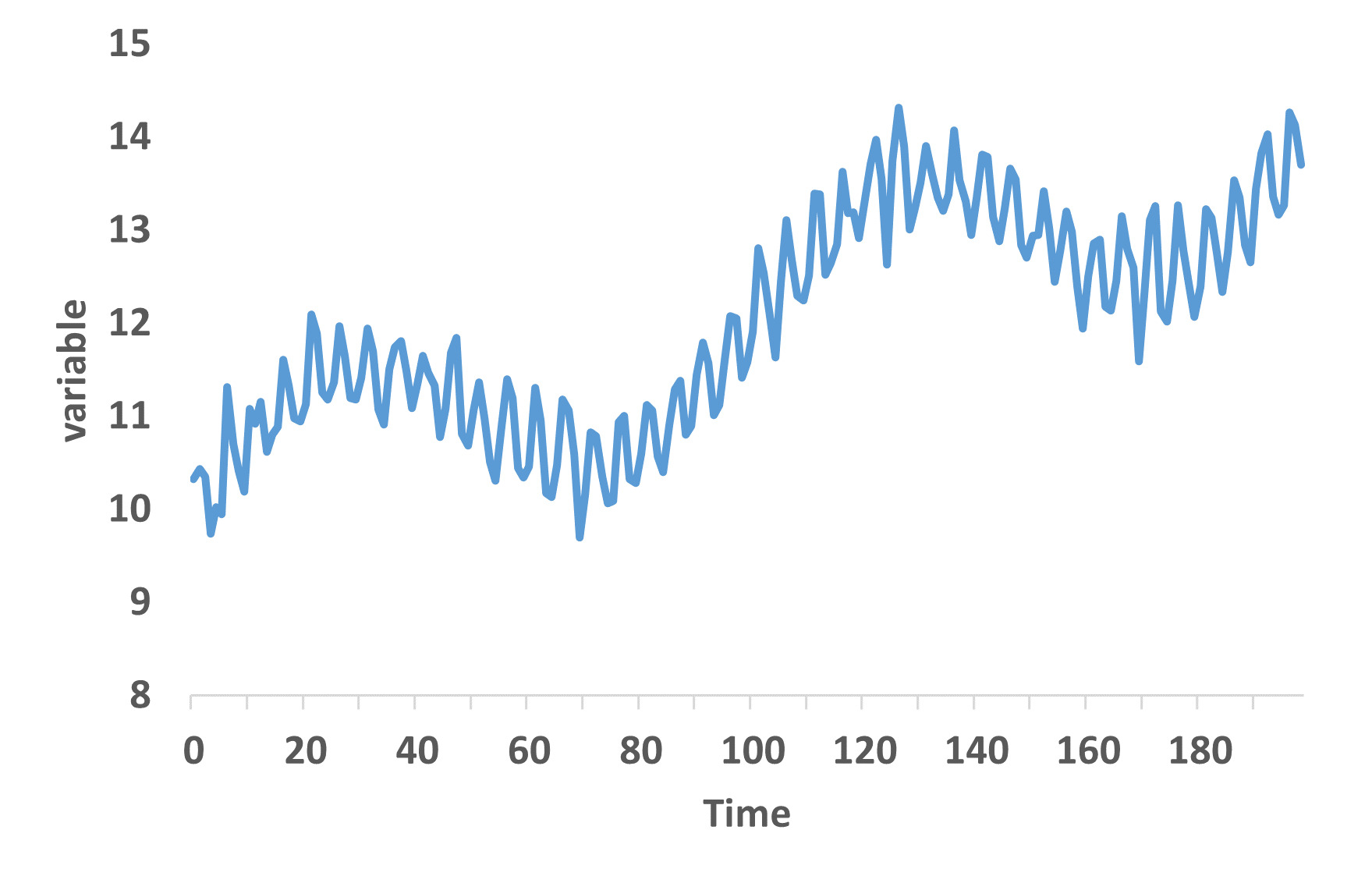}
         \caption{}
         \label{TCS}
     \end{subfigure}
        \caption{Time series with deterministic behavior: (a)  increasing trend, (b)  simple seasonality, (c) complex seasonality, (d) increasing trend and simple seasonality and (e) increasing trend and complex seasonality.}
        \label{fig:deterministic}
\end{figure}

\begin{figure}[H]
     \centering
     \begin{subfigure}[b]{0.3\textwidth}
         \centering
         \includegraphics[height=3.8cm,width=6cm]{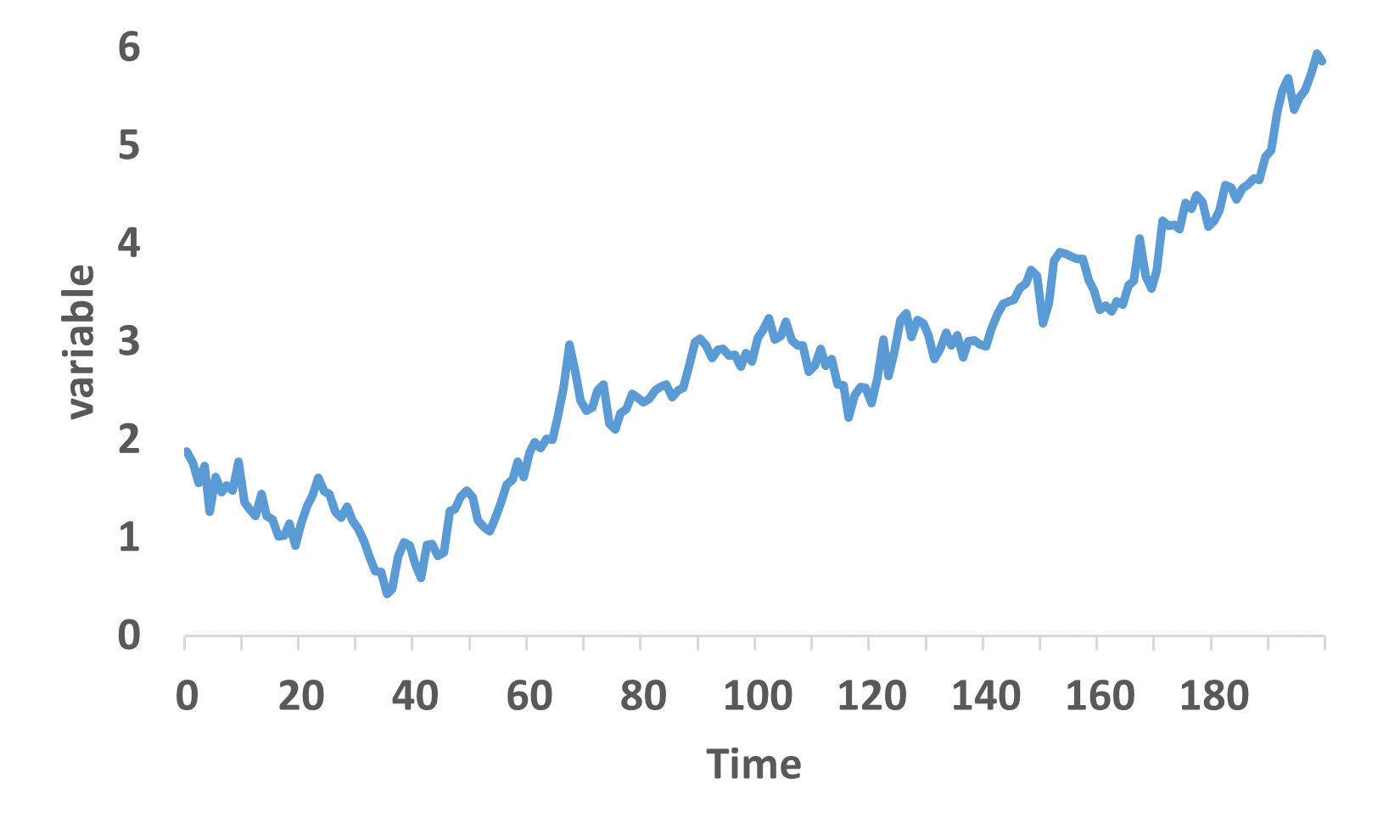}
         \caption{}
         \label{TRW}
     \end{subfigure}
     \hfill
     \begin{subfigure}[b]{0.3\textwidth}
         \centering
         \includegraphics[height=3.8cm,width=6cm]{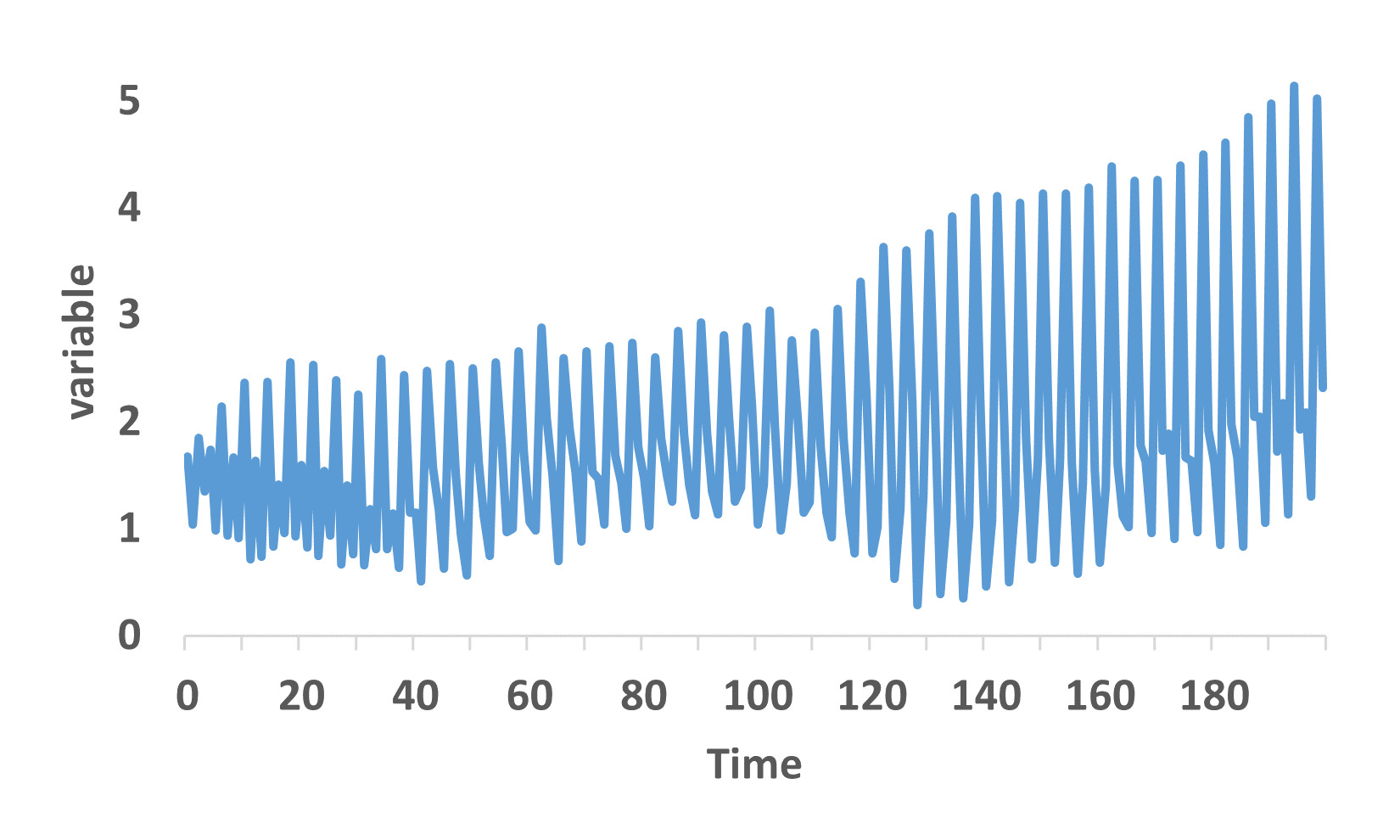}
         \caption{}
         \label{SRW}
     \end{subfigure}
     \hfill
     \begin{subfigure}[b]{0.3\textwidth}
         \centering
         \includegraphics[height=3.8cm,width=6cm]{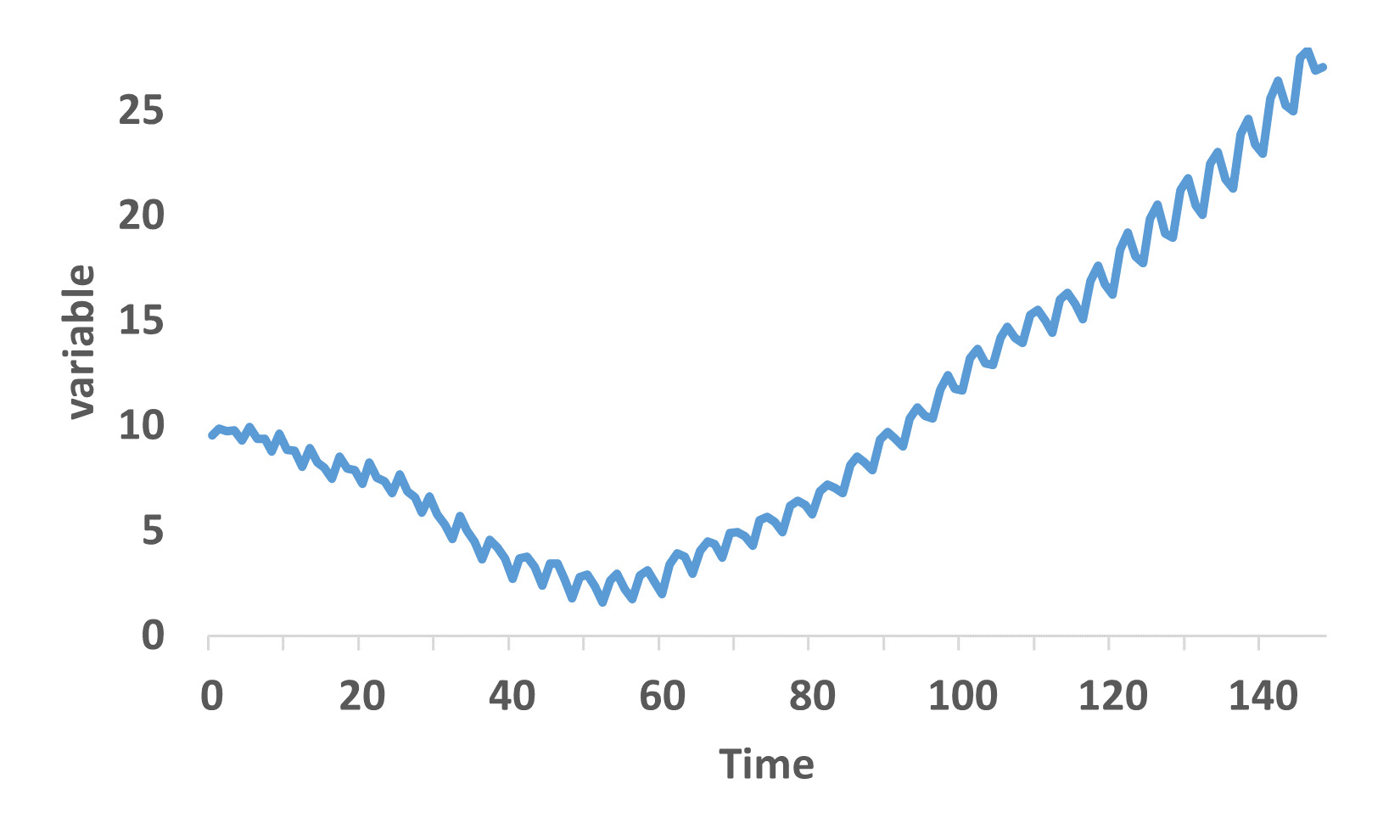}
         \caption{}
         \label{TSRW}
     \end{subfigure}
        \caption{Time series with random-walk behavior: (a) trend random-walk, (b)  seasonal random-walk, (c)  trend and seasonal random-walk.}
        \label{fig:unitroot}
\end{figure}

\begin{figure}[H]
     \centering
     \begin{subfigure}[b]{0.3\textwidth}
         \centering
         \includegraphics[height=3.8cm,width=6cm]{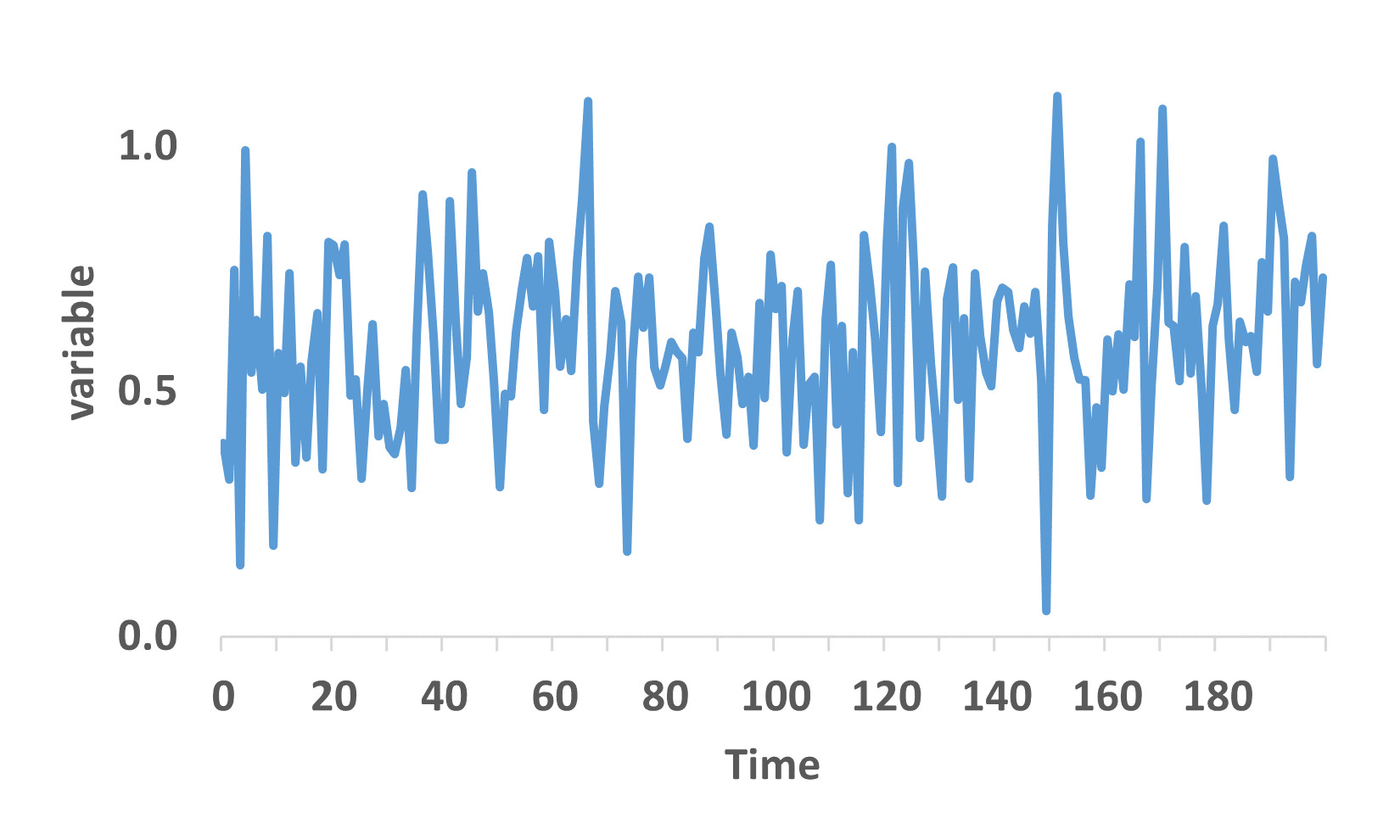}
         \caption{}
         \label{NAR}
     \end{subfigure}
     \hfill
     \begin{subfigure}[b]{0.3\textwidth}
         \centering
         \includegraphics[height=3.8cm,width=6cm]{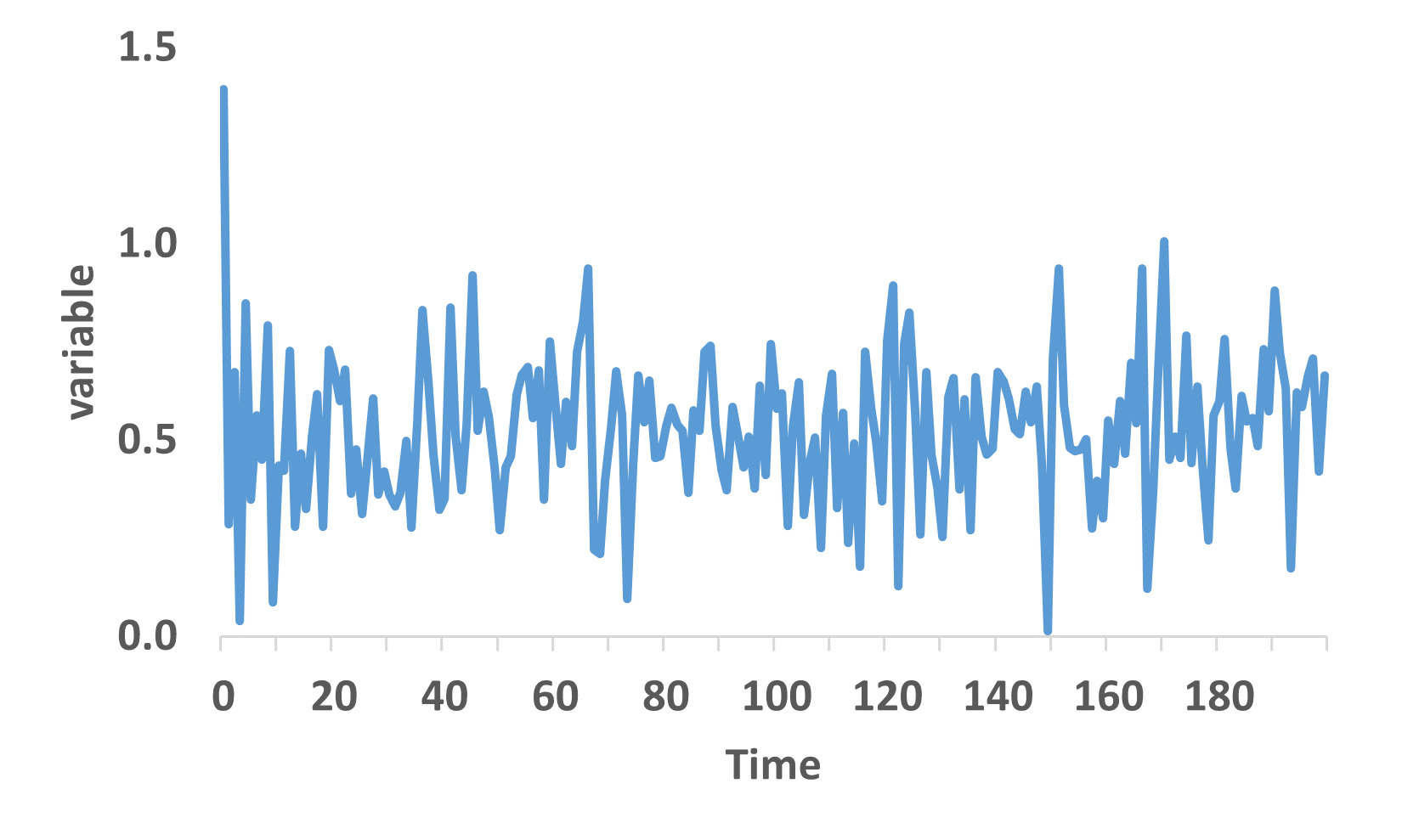}
         \caption{}
         \label{STAR}
     \end{subfigure}
     \hfill
     \begin{subfigure}[b]{0.3\textwidth}
         \centering
         \includegraphics[height=3.8cm,width=6cm]{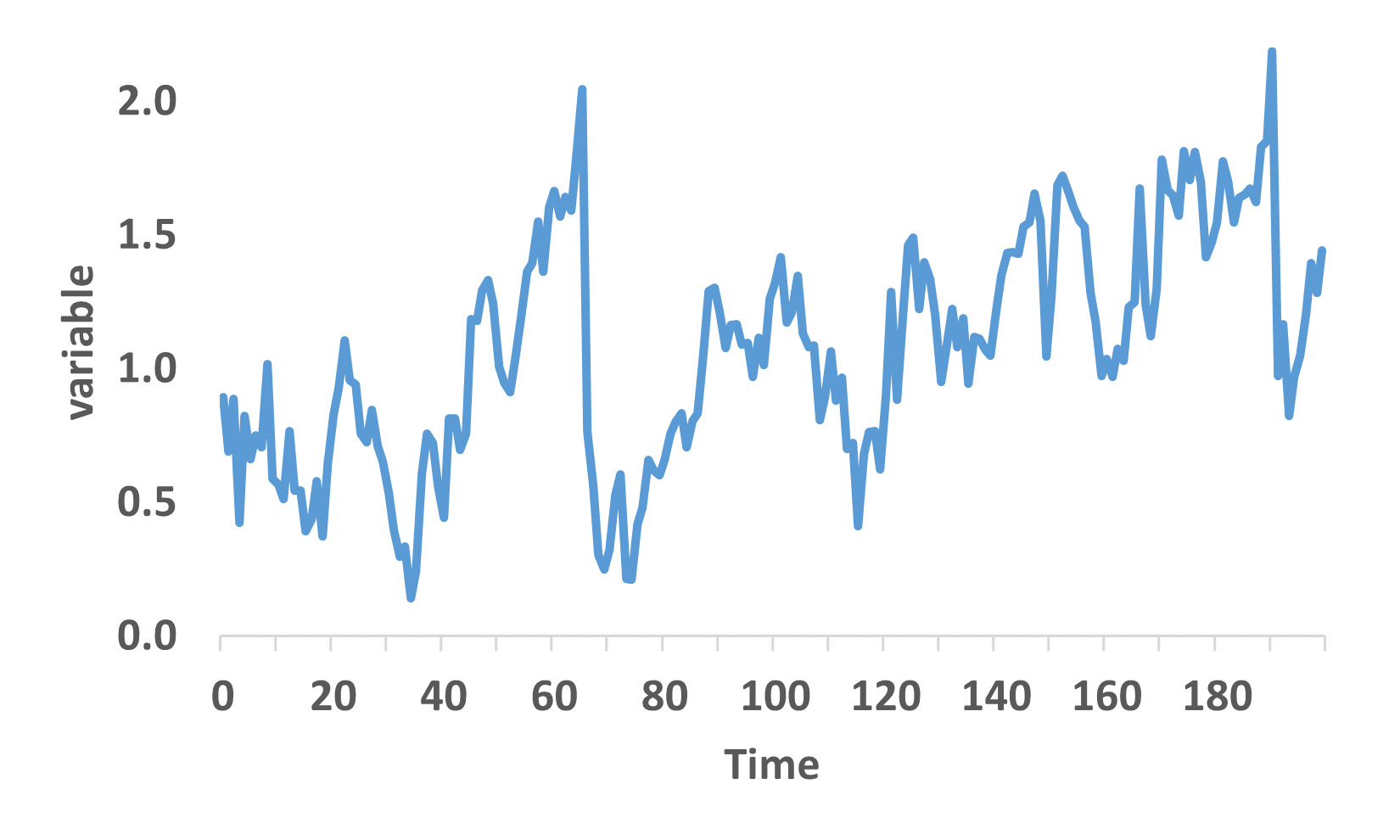}
         \caption{}
         \label{TAR}
     \end{subfigure}
        \caption{Time series with nonlinear behavior generated by: (a) Nonlinear Auto-Regressive (NAR) process, (b) Smooth Transition Auto-Regressive (STAR) process, (c) Threshold Auto-Regressive (TAR) process.}
        \label{fig:nonlinear}
\end{figure}

\begin{figure}[H]
     \centering
     \begin{subfigure}[b]{0.3\textwidth}
         \centering
         \includegraphics[height=3.8cm,width=6cm]{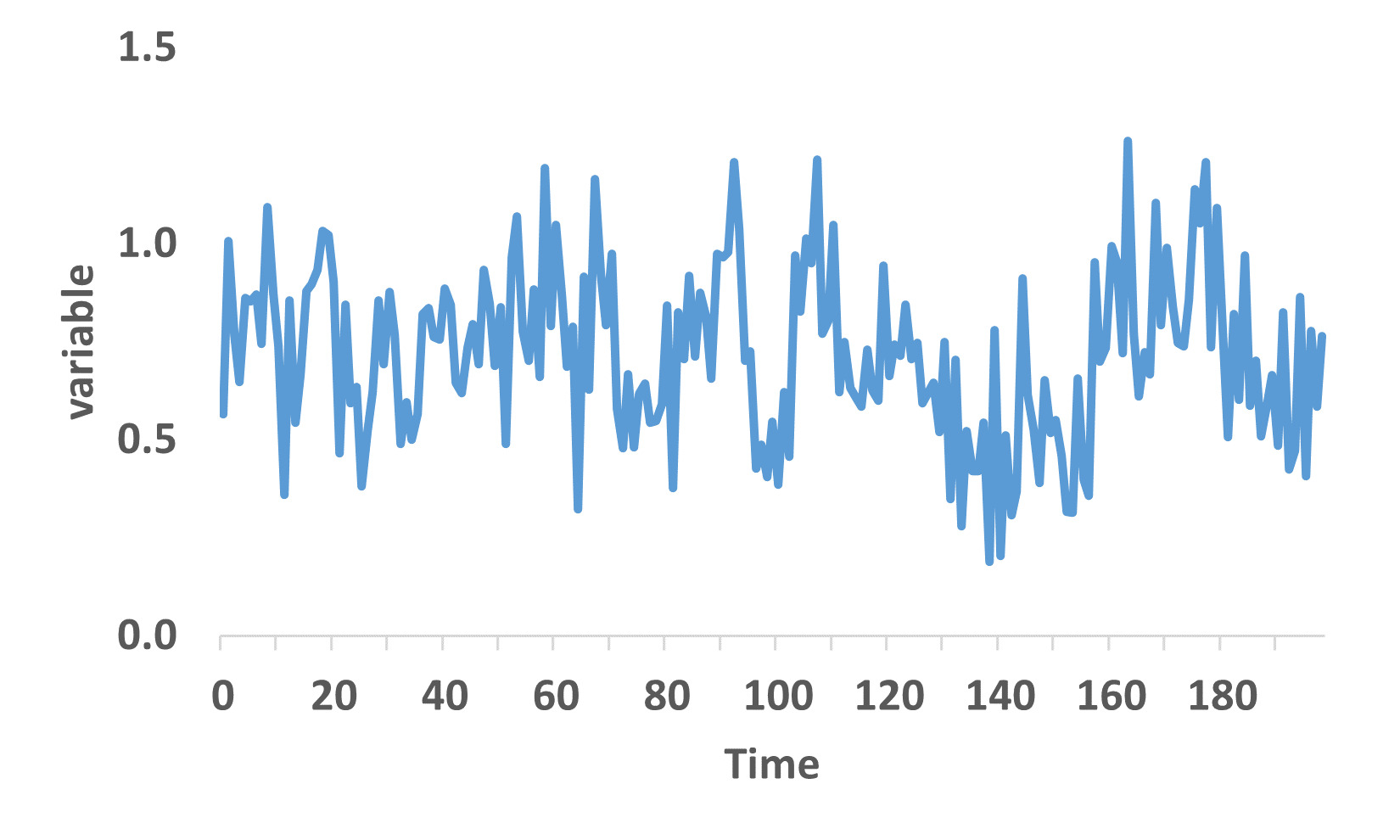}
         \caption{}
         \label{D00}
     \end{subfigure}
     \hfill
     \begin{subfigure}[b]{0.3\textwidth}
         \centering
         \includegraphics[height=3.8cm,width=6cm]{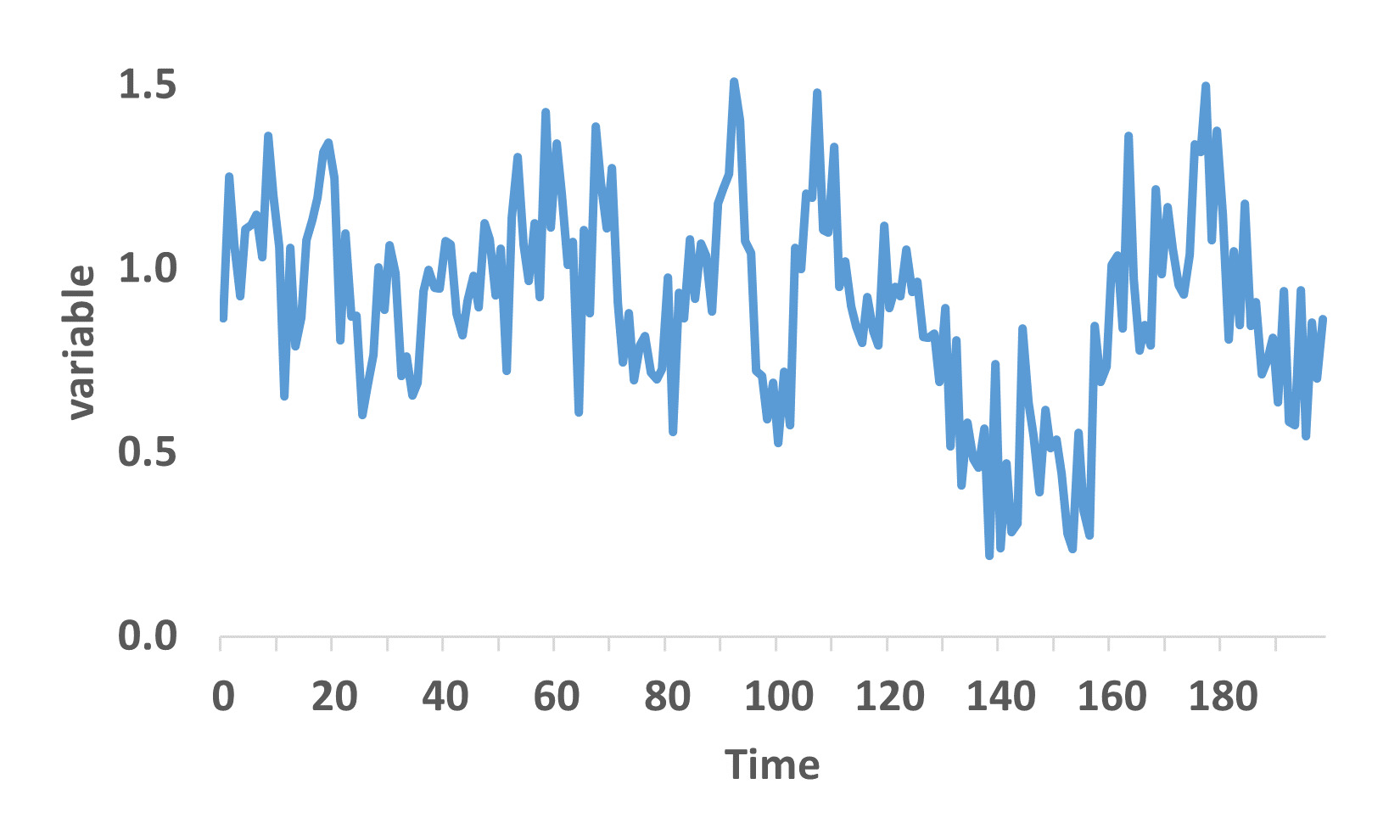}
         \caption{}
         \label{D02}
     \end{subfigure}
     \hfill
     \begin{subfigure}[b]{0.3\textwidth}
         \centering
         \includegraphics[height=3.8cm,width=6cm]{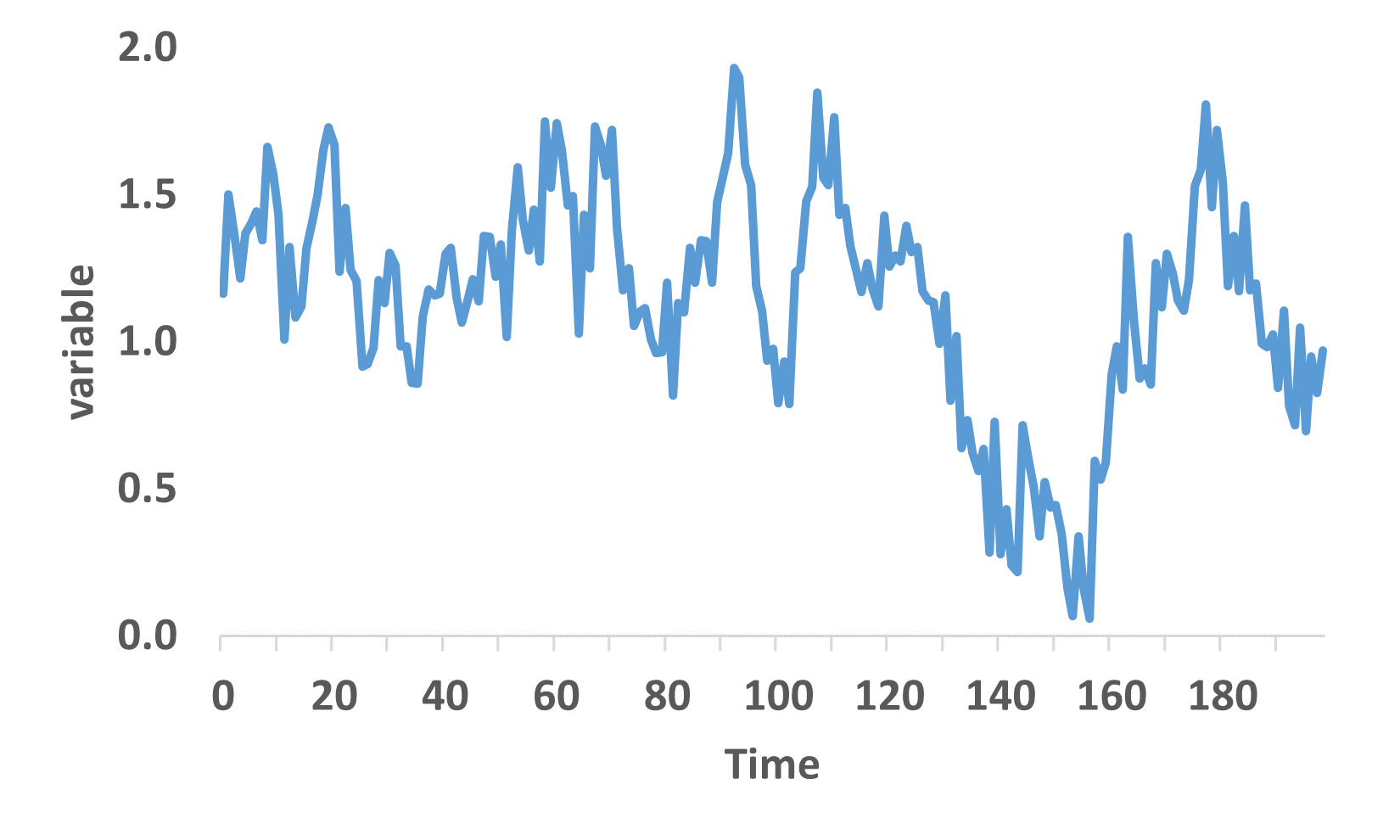}
         \caption{}
         \label{D04}
     \end{subfigure}
        \caption{Time series  with long-memory behavior: (a) short memory generated by ARFIMA(2,0.0,2), (b)  long memory generated by ARFIMA(2,0.2,2) and (c)  long memory generated by ARFIMA(2,0.4,2).}
        \label{fig:longmemory}
\end{figure}

In stochastic behavior, real times series are generated by a random stable system, and can exhibit the following behaviors: 

\begin{enumerate}
   
 \item  Deterministic behavior. These time series are characterized by the existence of deterministic patterns. They usually incorporate at least one of the following patterns: increasing or decreasing deterministic trend, simple deterministic seasonality, and complex deterministic seasonality (Figure \ref{fig:deterministic}). The trend pattern is a long-term evolution in the data, it can be increasing or decreasing, and it can have different forms (linear, exponential, and damped) \citep{montgomery2015introduction}. An increasing trend can appear in the demand for technologies in the social fields, while a decreasing trend is related to epidemics, mortality rates, and unemployment \citep{parmezan2019evaluation}. The seasonality pattern can be described as the occurrence of short-term regular variation that repeats at known and relatively constant time intervals. This type of pattern can occur in different types of data including time series of sales, e.g., the increase in sales of warm clothing in winter and air conditioners in summer \citep{parmezan2019evaluation}.

 \item  Random-Walk behavior. The time series of this behavior are characterized by the existence of unit-root patterns. This behavior appears when the time series have a stochastic increasing or decreasing trend and/or stochastic seasonality (Figure \ref{fig:unitroot}). Here the current observation is equal to the previous observation plus a random step. The lag between these two observations is equal to one in the case of trend and equal to the seasonality period in the case of seasonality.  
 
The presence of deterministic or random-walk behavior induces the non-stationarity in time series. The stationarity is a relevant feature in time series which basically implies that the mean, the variance, and the covariance do not depend on time \citep{montgomery2015introduction}. 
These two types of behaviors are the most important features in time series, and usually characterize business and macro-economic data \citep{salles2019nonstationary, liu2019nonpooling}.

 \item  Nonlinear behavior. Time series observations can often exhibit correlations with different degrees of non-linearity (Figure \ref{fig:nonlinear}). This type of behavior is present in almost all real-world time series data, such as in stream-flow forecasting \citep{wang2020monthly}, and in financial markets forecasting \citep{bukhari2020fractional}.
 
 \item  Long-memory behavior. Some time series may present properties of long-range dependencies in time which implies a strong coupling effect between the values of observations at different time steps, i.e. the correlations between observations has a slower exponential decay compared to short-range dependencies (Figure \ref{fig:longmemory}). This type of behavior can occur in hydrology forecasting \citep{papacharalampous2020hydrological}, network traffic forecasting \citep{ramakrishnan2018network}, financial market forecasting \citep{bukhari2020fractional}, etc.
 
\end{enumerate}

\subsection{Chaotic behavior}
\begin{figure}[H]
     \centering
     \begin{subfigure}[b]{0.3\textwidth}
         \centering
         \includegraphics[height=3.8cm,width=6cm]{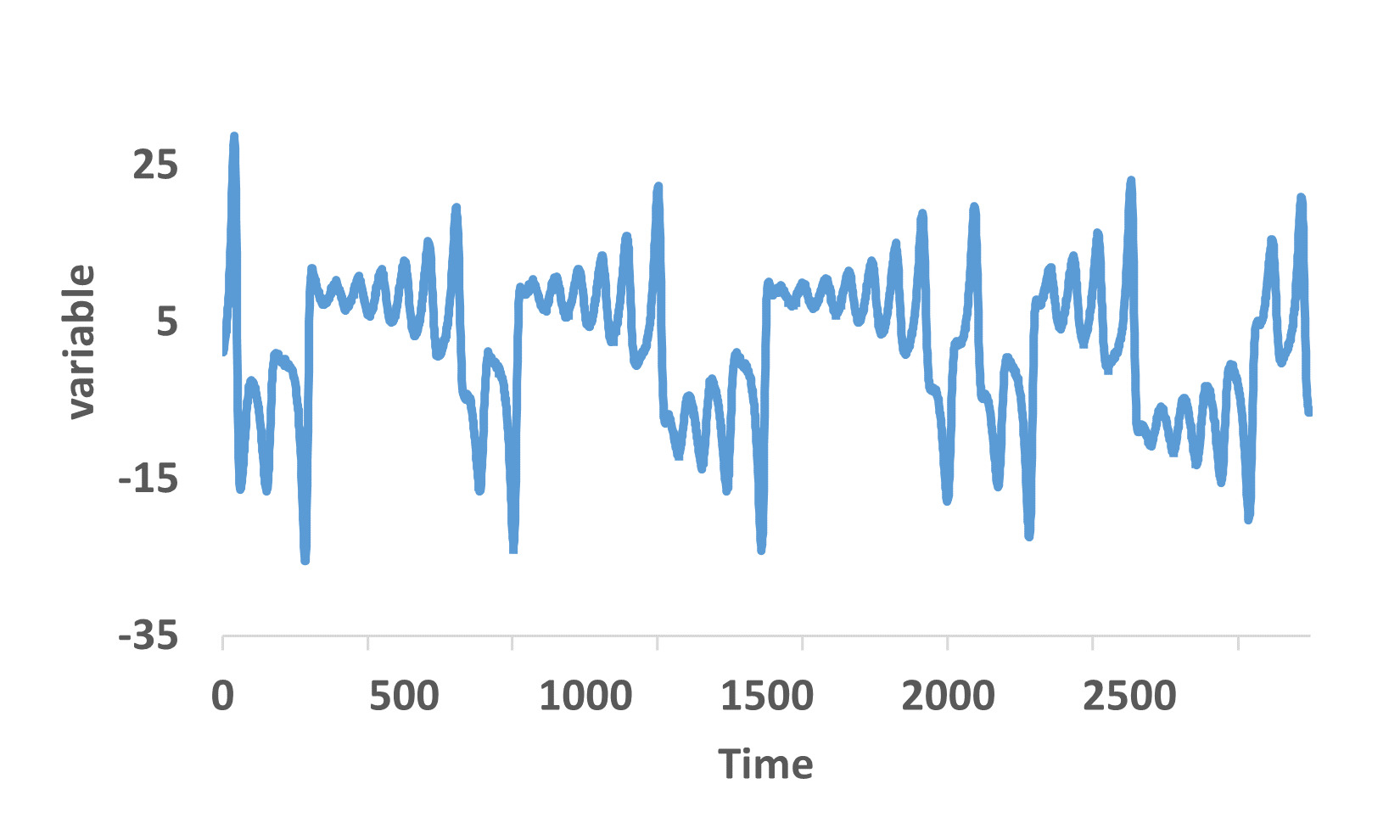}
         \caption{}
         \label{Lorenz}
     \end{subfigure}
     \hfill
     \begin{subfigure}[b]{0.3\textwidth}
         \centering
         \includegraphics[height=3.8cm,width=6cm]{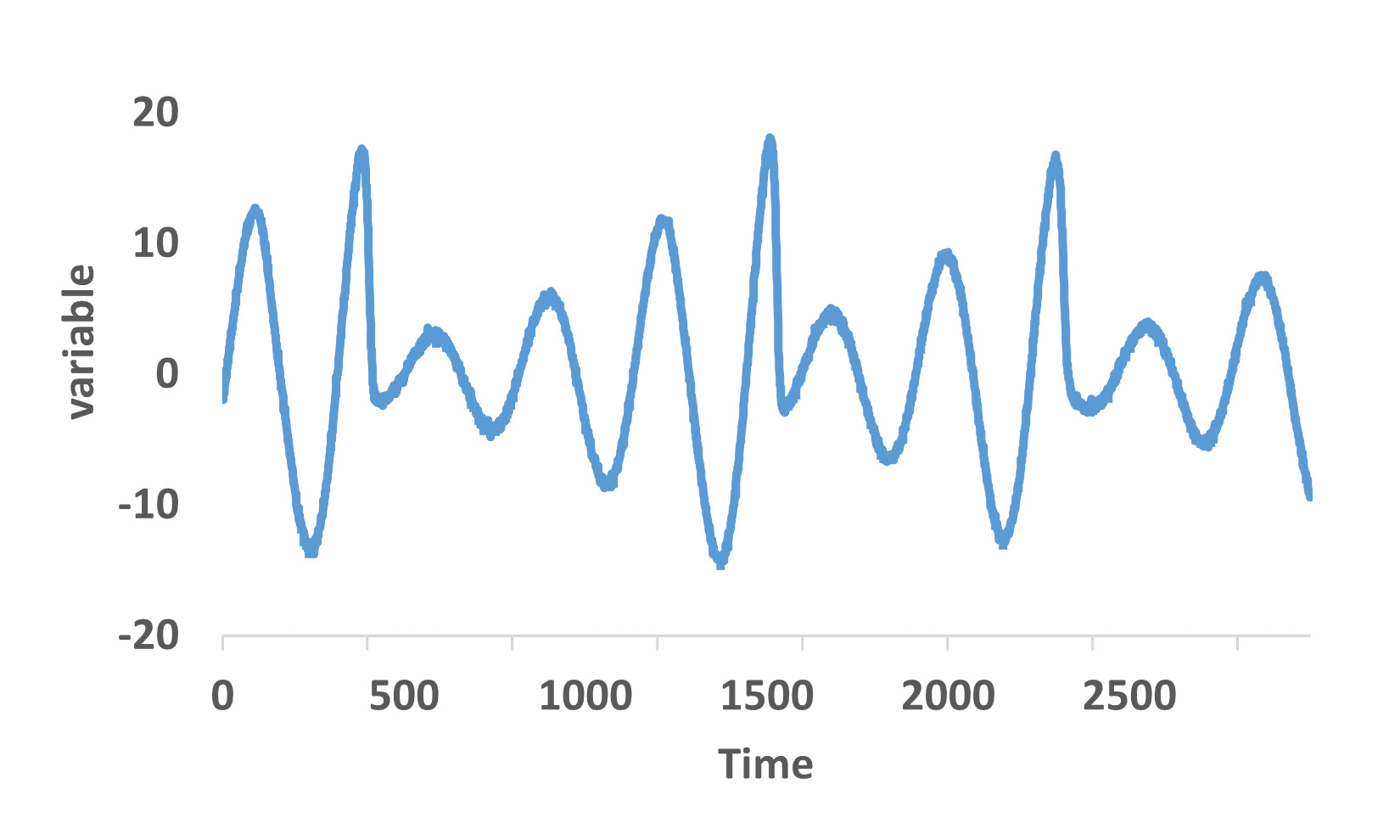}
         \caption{}
         \label{Rossler}
     \end{subfigure}
     \hfill
     \begin{subfigure}[b]{0.3\textwidth}
         \centering
         \includegraphics[height=3.8cm,width=6cm]{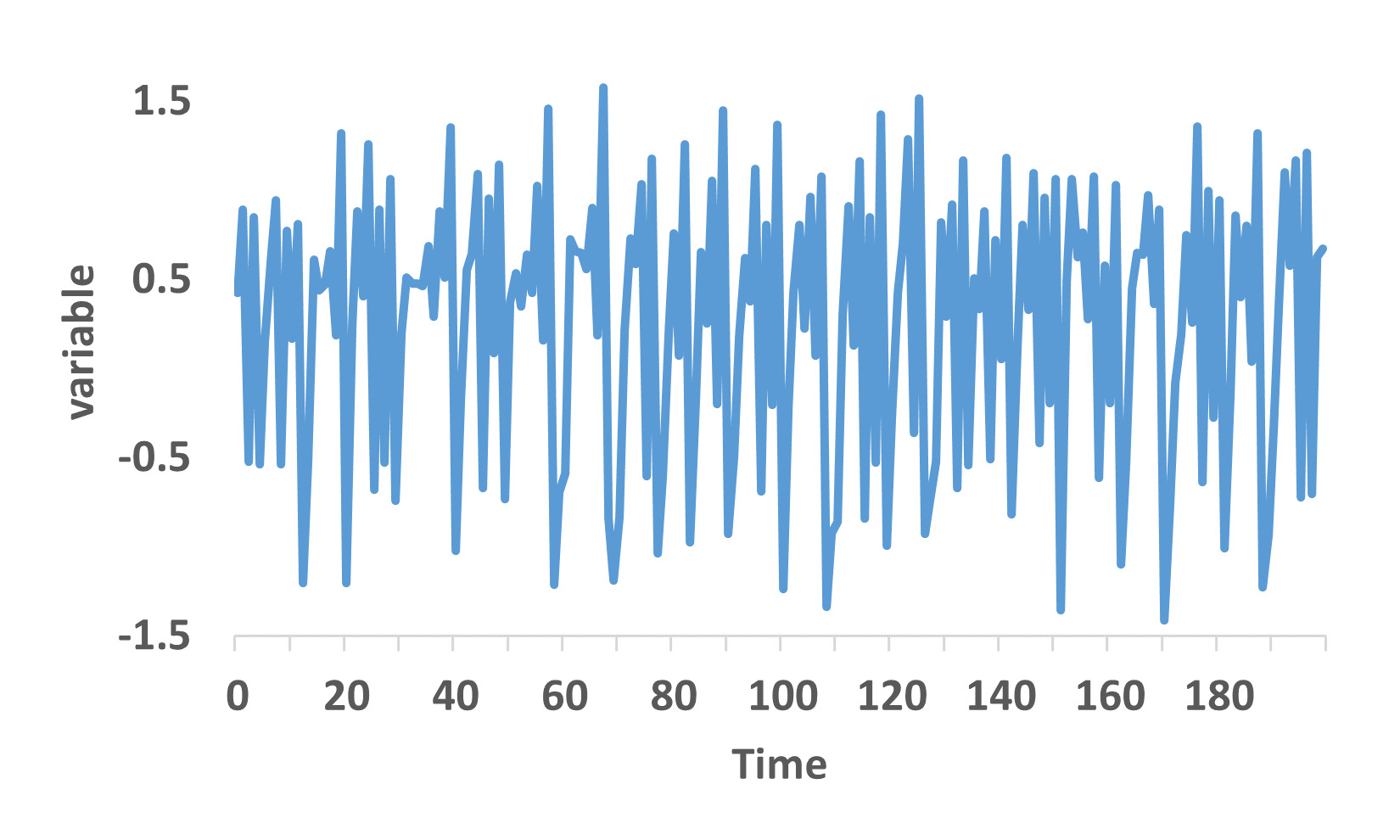}
         \caption{}
         \label{Henon}
     \end{subfigure}
     \hfill
     \begin{subfigure}[b]{0.4\textwidth}
         \centering
         \includegraphics[height=3.8cm,width=6cm]{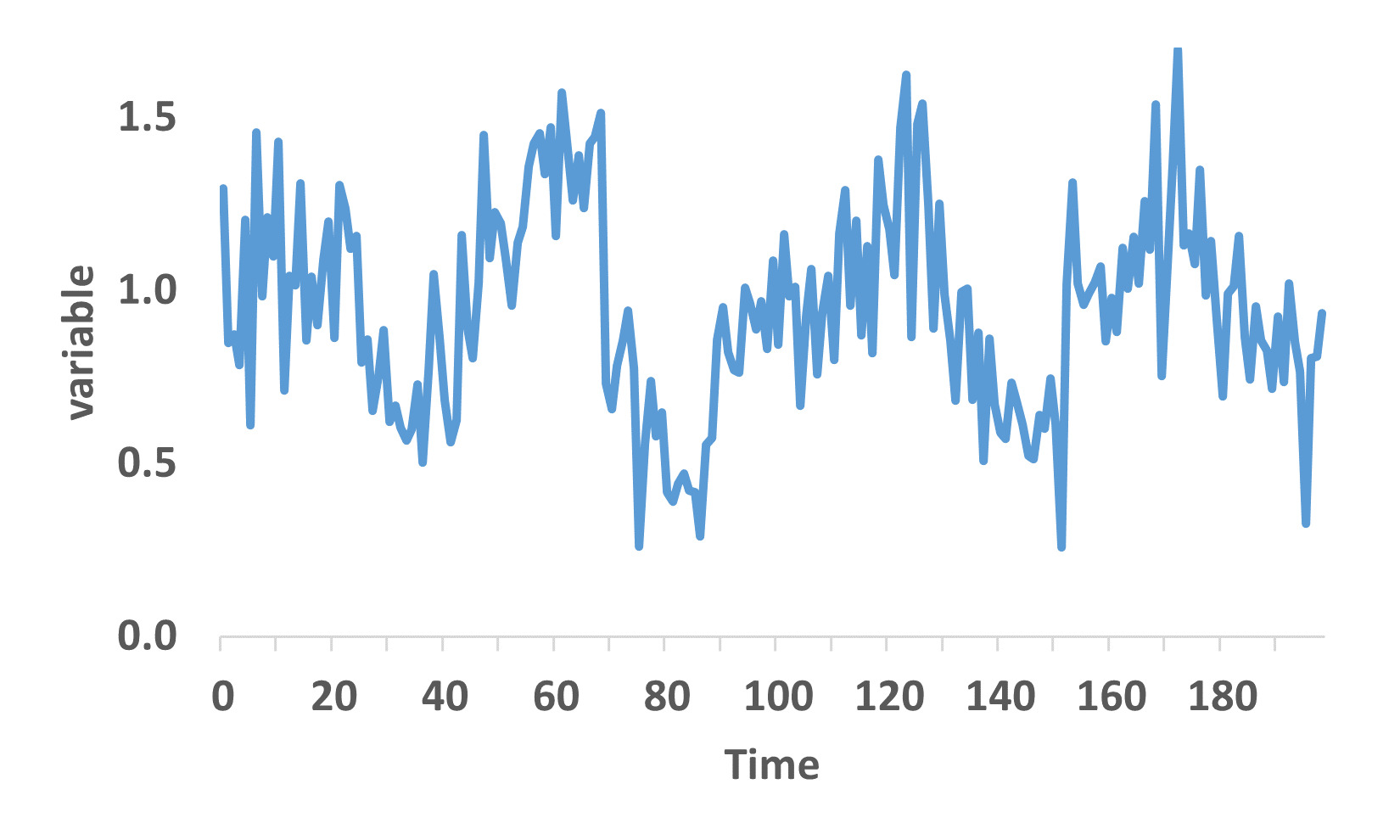}
         \caption{}
         \label{Mackey}
     \end{subfigure}
        \caption{Time series  with noisy chaotic behavior: (a) TS generated by Lorenz process. (b) TS generated by Rössler process. (c) TS generated by Hénon map process. (d) TS generated by Mackey-Glass process.}
        \label{fig:chaotic}
\end{figure}

The chaotic mechanism can be expressed as a nonlinear deterministic dynamical system that is often unknown or incompletely understood \citep{li2016new}. Real-time series can exhibit a noisy chaotic behavior (Figure \ref{fig:chaotic}), these time series are sensitive to initial conditions (butterfly effect) where small smooth perturbations in the system or measurement errors generate an abrupt change in the behavior of the time series (bifurcation). This type of behavior is unstable since it tends to be deterministic in short term but random in long term. Such kind of time series are usually present in many sciences and engineering fields such as  weather forecasting \citep{tian2019chaotic}, financial markets forecasting \citep{bukhari2020fractional}, energy forecasting \citep{bourdeau2019modeling}, intelligent transport and trajectory forecasting \citep{giuliari2021transformer}, etc.

Based on the literature \citep{chandra2012cooperative, montgomery2015introduction, liu2017evaluation, fischer2018machine}, the five aforementioned behaviors (deterministic, random-walk, nonlinear, long-memory, and chaotic) are the main behaviors encountered in real applications. 
Real-world time series can either express an individual behavior or an aggregation of more than one behavior. To identify which type of behavior a real-world time series can include, different statistical preprocessing tools and tests can be applied \citep{grau2005tests, inglada2020comprehensive}.
In Table \ref{Tools}, we present a set of tools that can be used to identify the existence of the five aforementioned behaviors in time series data.

\begin{table}[]
\centering
\caption{A set of tools used to identify the five different time series behaviors.}
\label{Tools}
\resizebox{\textwidth}{!}{
\begin{tabular}{llc}
\hline
\textbf{Tool} & \textbf{Time series behavior} & \textbf{Reference} \\\hline
 
Data visualization & Deterministic behavior & \citep{chatfield2013analysis} \\
Correlogram & Deterministic behavior & \citep{box2015time} \\
Time series decomposition & Deterministic behavior & \citep{chatfield2013analysis} \\
Smoothing & Deterministic behavior & \citep{montgomery2015introduction} \\
Data visualization & Random-Walk behavior & \citep{montgomery2015introduction} \\
Augmented Dickey Fuller (ADF) test & Random-Walk behavior & \citep{dickey1979distribution} \\
Phillips–Perron (PP) test & Random-Walk behavior & \citep{phillips1988testing} \\
Kwiatkowski–Phillips–Schmidt–Shin (KPSS) test & Random-Walk behavior & \citep{kwiatkowski1992testing} \\
Kaplan test & Nonlinear behavior & \citep{kaplan1994exceptional} \\
Keenan test & Nonlinear behavior & \citep{keenan1985tukey} \\
Tsay test & Nonlinear behavior & \citep{tsay1986nonlinearity} \\
Teräsvirta test & Nonlinear behavior & \citep{terasvirta1993power} \\
White test & Nonlinear behavior & \citep{white1989additional} \\
Correlogram & Long-memory behavior & \citep{palma2007long}\\
Qu test  & Long-memory behavior & \citep{qu2011test}\\
R/S analysis & Long-memory behavior & \citep{mandelbrot1968noah} \\
Modified R/S & Long-memory behavior & \citep{lo1991long} \\
Geweke and Porter-Hudak (GPH) test & Long-memory behavior & \citep{geweke1983estimation}\\
Detrended Fluctuation Analysis (DFA) & Long-memory behavior & \citep{peng1994mosaic} \\
Correlation Dimension & Chaotic behavior & \citep{grassberger1984dimensions} \\
Lyapunov Exponent & Chaotic behavior & \citep{bensaida2013high} \\
MGRM test & Chaotic behavior & \citep{matilla2010new} \\
Recurrence Plots & Chaotic behavior & \citep{eckmann1985ergodic} \\
0/1 test & Chaotic behavior & \citep{gottwald2004new} \\
\hline
\end{tabular}}
\end{table}

\section{Taxonomy of RNN cells}
\begin{figure}[H]
  \begin{center}
    \includegraphics[height=2.3cm,width=10cm]{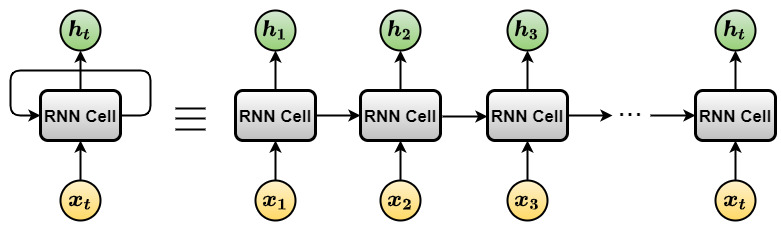} 
  \end{center}
  \caption{The folded (left) and unfolded (right) architecture of RNN model.}
  \label{rnn}
\end{figure}

Humans do not start their thinking from zero every second, our thoughts have persistence in the memory of our brains. For example, as the reader reads this paper, he/she understands each word based on his/her understanding of the words before. 
The absence of memory is the major shortcoming in traditional machine learning models, particularly in feed-forward neural networks (FNNs). To overcome this limitation, RNNs integrate the concept of feedback connections in their structure (Figure \ref{rnn}, where $x_t$ and $h_t$ are the input state and the hidden state at time step $t$, respectively). This mechanism enables RNNs to have a certain memory capable of capturing the dynamics in sequential data by conveying information through time. 

\begin{table}[H]
\centering
\caption{A collection of RNN cell structures historically sorted.}
\label{RNN cells}
\begin{tabular}{lrc}
\hline
\textbf{RNN cell structure} & \textbf{Year} & \textbf{Reference} \\\hline
JORDAN & 1989 & \citep{jordan1989serial} \\
ELMAN & 1990 & \citep{elman1990finding} \\
LSTM-NFG & 1997 & \citep{hochreiter1997long} \\
LSTM-Vanilla & 2000 & \citep{gers2000learning} \\
LSTM-PC & 2000 & \citep{gers2000recurrent} \\
SCRN & 2014 & \citep{mikolov2014learning} \\
GRU & 2014 & \citep{cho2014learning} \\
IRNN & 2015 & \citep{le2015simple} \\
LSTM-FB1 & 2015 & \citep{jozefowicz2015empirical} \\
MUT & 2015 & \citep{jozefowicz2015empirical} \\
LSTM-CIFG & 2015 & \citep{nina2015simplified} \\
Differential LSTM & 2015 & \citep{veeriah2015differential} \\
MRNN & 2016 & \citep{abdulkarim2016time} \\
MGU & 2016 & \citep{zhou2016minimal} \\
Phased LSTM & 2016 & \citep{neil2016phased} \\
Highway Connections & 2016 & \citep{irie2016lstm} \\
LSTM with Working Memory & 2017 & \citep{pulver2017lstm} \\
SLIM & 2017 & \citep{lu2017simplified, dey2017gate, heck2017simplified} \\
GORO & 2019 & \citep{jing2019gated} \\

\hline
\end{tabular}
\end{table}

\begin{figure}[H]
     \centering
     \begin{subfigure}[b]{0.3\textwidth}
         \centering
         \includegraphics[height=3.8cm,width=4.2cm]{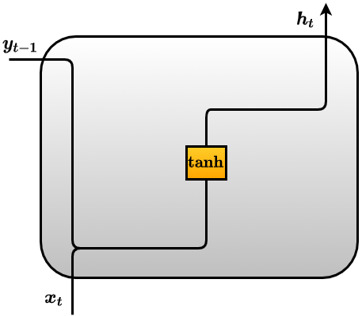}
         \caption{}
         \label{JORDAN}
     \end{subfigure}
     \hfill
     \begin{subfigure}[b]{0.3\textwidth}
         \centering
         \includegraphics[height=3.8cm,width=4.2cm]{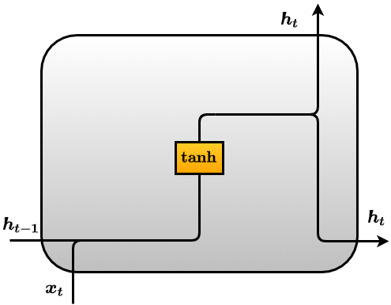}
         \caption{}
         \label{ELMAN}
     \end{subfigure}
     \hfill
     \begin{subfigure}[b]{0.3\textwidth}
         \centering
         \includegraphics[height=3.8cm,width=4.2cm]{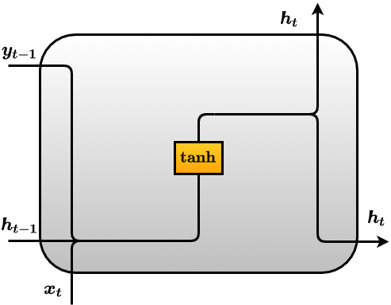}
         \caption{}
         \label{MRNN}
     \end{subfigure}
        \caption{(a) JORDAN cell structure. (b) ELMAN cell structure. (c) MRNN cell structure.}
        \label{srn}
\end{figure}

RNN models are built based on one specific cell structure which is the core of all computations that occur in the network. Multiple cell structures have been created since 1989 (Table \ref{RNN cells}).
The early cell structure is named JORDAN \citep{jordan1989serial}, where at each time step the previous output state is fed into the cell (Figure \ref{JORDAN}). Later,  the ELMAN cell was proposed by \citep{elman1990finding}. Unlike the JORDAN cell, each time step in the ELMAN cell calls the previous hidden state (Figure \ref{ELMAN}). In 2016, a combination of both JORDAN and ELMAN cells in one cell named multi-recurrent neural network (MRNN) was evaluated by \citep{abdulkarim2016time}. In this cell structure, at each time step, both previous output and hidden states are presented to the cell (Figure \ref{MRNN}).

\begin{figure}[H]
     \centering
     \begin{subfigure}[b]{0.3\textwidth}
         \centering
         \includegraphics[height=3.8cm,width=4.8cm]{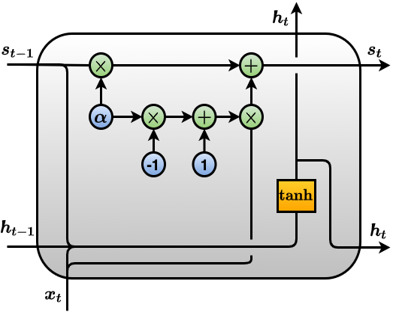}
         \caption{}
         \label{SCRN}
     \end{subfigure}
     \begin{subfigure}[b]{0.3\textwidth}
         \centering
         \includegraphics[height=3.8cm,width=4.8cm]{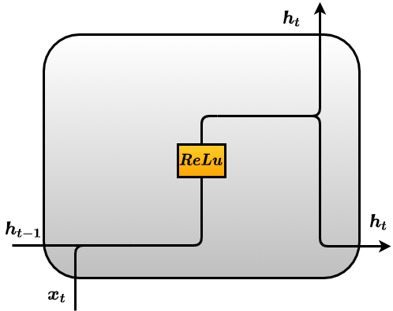}
         \caption{}
         \label{IRNN}
     \end{subfigure}
        \caption{(a) SCRN cell structure. (b) IRNN cell structure.}
        \label{scrn-irnn}
\end{figure}

It was proved that the ELMAN cell suffers from the vanishing and exploding gradient problems that impede the capturing of long-term dependencies \citep{pascanu2013difficulty}. To overcome the memory limitation of this cell, novel cell structures have been proposed. In 2014, the Structurally Constrained Recurrent Network (SCRN) was proposed by \citep{mikolov2014learning}. They integrated a slight structural modification in the ELMAN cell that consists in adding a new slowly changing state at each time step called context state $s_{t-1}$ (Figure \ref{SCRN}). In 2015, \citep{le2015simple} created a new cell called Identity Recurrent Neural Network (IRNN) as a modification of ELMAN by setting the ReLu as the activation function, the identity matrix as an initialization of the hidden states weight matrix, and zero as an initialization of the bias (Figure \ref{IRNN}).

A different way to handle the vanishing and exploding gradient problems resulted in creating different cell structures characterized by the gating mechanism that regulates the flowing of the information flux. The gates can be seen as filters that only hold useful information and selectively remove any irrelevant information from passing through. To perform this control of information (i.e., which information to pass and which information to discard), the gates are equipped with parameters that need to be trained through the model learning process using the back-propagation through time algorithm \citep{werbos1990backpropagation}. Thus, this mechanism provides the RNN cell with an internal permanent memory able to store information for long time periods \citep{weston2014memory, graves2014neural}. 

\begin{figure}[H]
     \centering
     \begin{subfigure}[b]{0.3\textwidth}
         \centering
         \includegraphics[height=3.8cm,width=4.8cm]{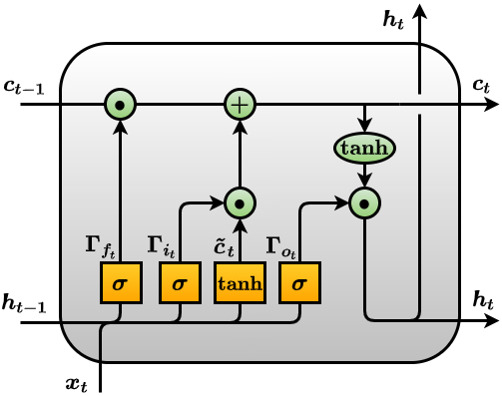}
         \caption{}
         \label{vanilla}
     \end{subfigure}
     \begin{subfigure}[b]{0.3\textwidth}
         \centering
         \includegraphics[height=3.8cm,width=4.8cm]{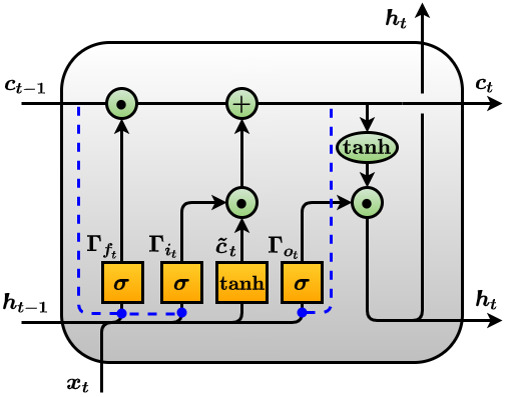}
         \caption{}
         \label{PC}
     \end{subfigure}
        \caption{(a) LSTM-Vanilla cell structure. (b) LSTM-PC cell structure.}
        \label{lstm}
\end{figure}

In 1997, the first version of this type of cell named Long-Short Term Memory with No Forget Gate (LSTM-NFG) was created by \citep{hochreiter1997long}. This cell contains two gates: the input gate $\Gamma_{i_t}$ and the output gate $\Gamma_{o_t}$. Later in 2000, the concept of the forget gate $\Gamma_{f_t}$ was introduced by \citep{gers2000learning} creating LSTM-Vanilla that has been widely used in most applications (Figure \ref{vanilla}). In the same year, the LSTM cell with peephole connections (LSTM-PC) was proposed by \citep{gers2000recurrent}. The peephole connections connect the previous cell state $c_{t-1}$ with the input, forget, and output gates (Figure \ref{PC}). These connections enable the LSTM cell to inspect its current cell states \citep{gers2001lstm}, and to learn precise and stable timing without teacher forcing \citep{gers2002learning}. 

\begin{figure}[H]
     \centering
     \begin{subfigure}[b]{0.3\textwidth}
         \centering
         \includegraphics[height=3.8cm,width=4.9cm]{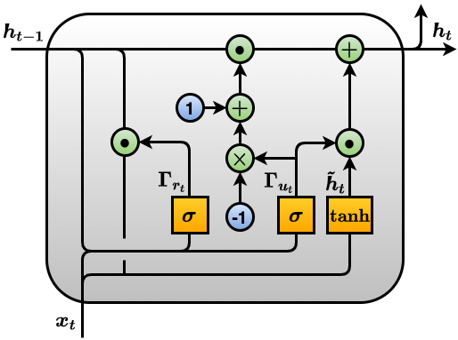}
         \caption{}
         \label{gru}
     \end{subfigure}
     \hfill
     \begin{subfigure}[b]{0.3\textwidth}
         \centering
         \includegraphics[height=3.8cm,width=4.8cm]{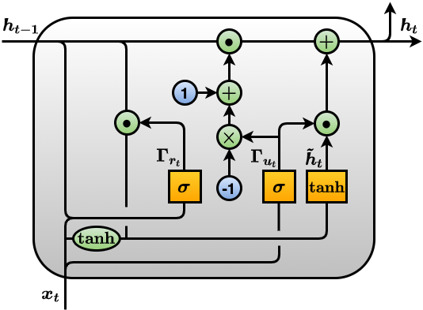}
         \caption{}
         \label{mut1}
     \end{subfigure}
     \hfill
     \begin{subfigure}[b]{0.3\textwidth}
         \centering
         \includegraphics[height=3.8cm,width=4.8cm]{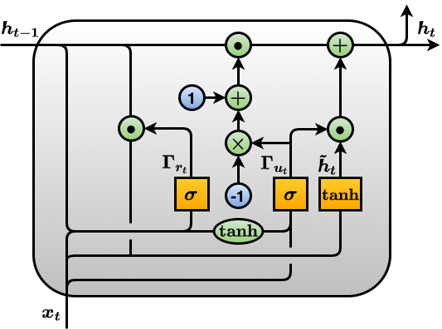}
         \caption{}
         \label{mut3}
     \end{subfigure}
        \caption{(a) GRU cell structure. (b) MUT1 cell structure. (c) MUT3 cell structure.}
        \label{gru-mut}
\end{figure}

\begin{figure}[H]
     \centering
          \begin{subfigure}[b]{0.3\textwidth}
         \centering
         \includegraphics[height=3.8cm,width=4.8cm]{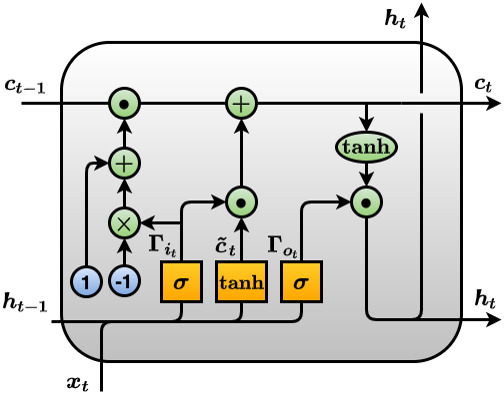}
         \caption{}
         \label{cifg}
     \end{subfigure}
     \begin{subfigure}[b]{0.3\textwidth}
         \centering
         \includegraphics[height=3.6cm,width=4.9cm]{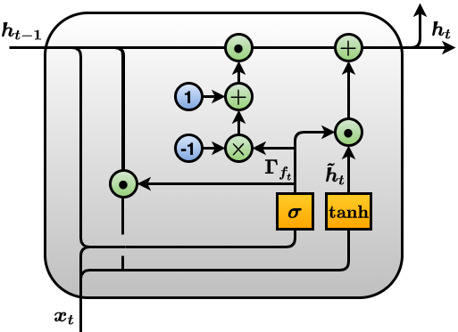}
         \caption{}
         \label{mgu}
     \end{subfigure}
        \caption{(a) LSTM-CIFG cell structure. (b) MGU cell structure.}
        \label{mgu-cifg}
\end{figure}

In 2014, the Gated Recurrent Unit (GRU) cell was proposed by \citep{cho2014learning} as a simpler variant of LSTM that shares many of the same properties. The idea behind GRU cell was to reduce the gating mechanism of LSTM cell from three gates to two gates (relevance gate $\Gamma_{r_t}$ and update gate $\Gamma_{u_t}$) in order to decrease the number of parameters and to improve the learning velocity (Figure \ref{gru}). In 2015, ten thousand RNN cell structures were evaluated by \citep{jozefowicz2015empirical} using a mutation-based search process. They identified a new cell architecture that outperforms both LSTM and GRU on some tasks. This cell consists in adding a bias of 1 to LSTM forget gate creating the LSTM-FB1 cell. Further, they discovered three optimal cell architectures named MUT1, MUT2, and MUT3 that are similar to GRU but have some modifications in their gating mechanism and in their candidate hidden state $\tilde{h_t}$ (Figure \ref{mut1} and \ref{mut3}). 
During that year, coupling both the input and the forget gates into one gate was proposed by \citep{nina2015simplified} creating the LSTM-CIFG cell (Figure \ref{cifg}). Further, the differential LSTM cell was proposed by \citep{veeriah2015differential} to solve the impact of spatial-temporal dynamics by introducing the differential gating scheme in LSTM cell.

In 2016, the Minimal Gate Unit (MGU) cell was created by \citep{zhou2016minimal} to further reduce the number of parameters by decreasing the gating mechanism to one forget gate. This variant has a simpler structure and fewer parameters compared to LSTM and GRU cells (Figure \ref{mgu}). 
In the same year, eight variants of LSTM-PC cell (based on modifying, adding, or removing one cell component at each time) were evaluated by \citep{greff2016lstm} on three different types of tasks: speech recognition, polyphonic music modeling, and handwritten recognition. They demonstrated that the forget and the output gates are the most critical components in LSTM cell. In addition, their results show that none of the evaluated variants can overcome the LSTM-PC cell. 
During that year, the phased LSTM cell was introduced by \citep{neil2016phased}, where they added a time gate that updates the cell sparsely, and makes it converge faster than the basic LSTM. 
Further, highway connections were added to GRU and LSTM cells by  \citep{irie2016lstm}. 
In 2017, LSTM with working memory was created by \citep{pulver2017lstm}, where they substituted the forget gate with a functional layer whose input depends on the previous cell state. 
In 2019, a Gated Orthogonal Recurrent Unit (GORO) was introduced by \citep{jing2019gated}, where they added to the GRU cell an orthogonal matrix that replaced the hidden state loop matrix.

\begin{figure}[H]
     \centering
     \begin{subfigure}[b]{0.3\textwidth}
         \centering
         \includegraphics[height=3.8cm,width=4.8cm]{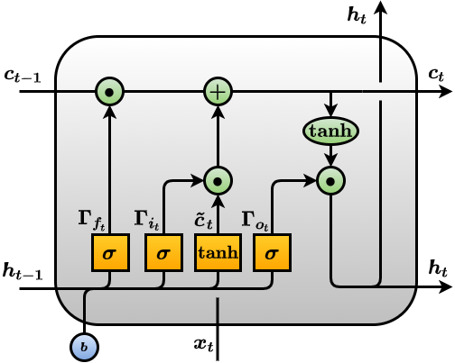}
         \caption{}
         \label{slim1}
     \end{subfigure}
     \hfill
     \begin{subfigure}[b]{0.3\textwidth}
         \centering
         \includegraphics[height=3.8cm,width=4.8cm]{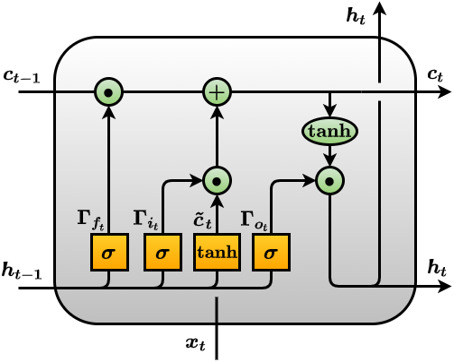}
         \caption{}
         \label{slim2}
     \end{subfigure}
     \hfill
     \begin{subfigure}[b]{0.3\textwidth}
         \centering
         \includegraphics[height=3.8cm,width=4.8cm]{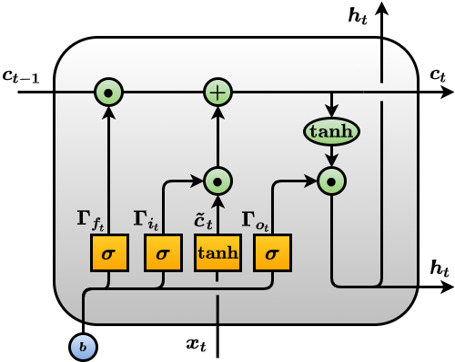}
         \caption{}
         \label{slim3}
     \end{subfigure}
        \caption{(a) SLIM1 cell structure. (b) SLIM2 cell structure. (c) SLIM3 cell structure.}
        \label{slim}
\end{figure}

While very powerful in long-term dependencies, the basic cells (LSTM, GRU, and MGU) have a complex structure with a relatively large number of parameters. In 2017, the concept of parameter reduction was differently tackled through the creation of new cells called SLIM \citep{lu2017simplified, dey2017gate, heck2017simplified} while maintaining the same number of gates in the basic cells. These variants aim to reduce aggressively the parameters in order to achieve memory and time savings while necessarily retaining a performance comparable to the basic cells.  
The new parameter-reduced variants of the basic cells eliminate the combinations of the input state, the hidden state, and the bias from the individual gating signals, creating SLIM1, SLIM2, and SLIM3, respectively. The SLIM1 cell consists in removing from the mechanism of all the gates the input state and its associated parameter matrix (Figure \ref{slim1}). The SLIM2 cell consists in maintaining only the hidden state and its associated parameter matrix (Figure \ref{slim2}). Whereas, the SLIM3 cell consists in removing the input state, the hidden state, and their associated parameters matrices (Figure \ref{slim3}). The cellular calculations within the displayed RNN cells along with the evaluated ones are provided in Appendix.

\section{Experimental structure}
 Two experiments have been carried out in this study.  The first experiment analyzes the utility of each LSTM-Vanilla cell component in forecasting time series behaviors. The second experiment evaluates different variants of RNN cell structures in forecasting time series behaviors. In this section, we first describe the process we followed to generate the dataset for each time series behavior (Section \ref{section_DGP}). Then, we present the selected models for the first and the second experiment (Section \ref{section_ARCH}). Finally, we provide the setup of the used models (Sections \ref{Data preparation}, \ref{Parameter optimization}, \ref{Hyperparameter selection}, and \ref{Performance evaluation}). 
 
\subsection{Synthetic data generation}
\label{section_DGP}
To simulate the five aforementioned time series behaviors, we used 21 different DGPs with white Gaussian noise $\epsilon_t \sim \mathcal{N}(\mu=0, \sigma=0.2) $. From each DGP we created time series of length 3000 observations replicated 30 times through a Monte Carlo simulation experiment using different initial random seeds for the white noise term $\epsilon_t$ (Table \ref{tab4}).

To generate time series with deterministic behavior, 5 DGPs were used (Table \ref{tab5}): Trend process (T), Simple Seasonality process (SS), Complex Seasonality process (CS), Trend and Simple Seasonality process (TSS), and Trend and Complex Seasonality process (TCS). 
To simulate the random-walk behavior, 3 DGPs were used (Table \ref{tab6}): Trend Random-Walk process (TRW), Seasonal Random-Walk process (SRW), and Trend and Seasonal Random-Walk process (TSRW). 
To simulate time series with nonlinear behavior, we used 6 most popular nonlinear models commonly used in the forecasting literature having an increasing level of non-linearity \citep{zhang2001simulation} (Table \ref{tab8}): Sign Auto-Regressive process (SAR), Nonlinear Moving Average process (NMA), Nonlinear Auto-Regressive process (NAR), Bilinear process (BL), Smooth Transition Auto-Regressive process (STAR), and Threshold Auto-Regressive process (TAR). These models are motivated by many nonlinear characteristics commonly observed in practice. 
To artificially generate time series with long memory behavior, we used the Auto-Regressive Fractionally Integrated Moving Average process ARFIMA(p,d,q) since it is one of the best-known long memory processes \citep{liu2017evaluation}. In order to evaluate the performance of RNN cell structures with respect to DGPs with an increasing memory structure, 2 DGPs were created based on the variation of the fractional order of ARFIMA process $d=\{0.2, 0.4\}$. A higher fractional order $d$ implies longer dependency structure (Table \ref{Long DGP}). To ensure the stationarity of the generated time series, we set the values of the fractional order strictly less than 0.5. 
Finally, to simulate the noisy chaotic behavior, the 4 most known chaotic DGPs were used (Table \ref{tab7}): Mackey-Glass process, Lorenz process, Rössler process, and Hénon-Map process. Then, we added the white Gaussian noise $\epsilon_t$ to the deterministic signals to create noisy chaotic time series \citep{sangiorgio2021forecasting}. 


\begin{table}[H]
\centering
\caption{Number of data generation processes and time series in each behavior, and the number of hyperparameter selection experiments per model in each behavior.}
\label{tab4}
\begin{tabular}{lrrr}
\hline
\textbf{Behavior} & \textbf{\# DGPs} & \textbf{\# Time series} & \makecell[c]{\bf \# Experiments \\ \bf per model} \\\hline
Deterministic behavior & 5 & 150 & 50\\
Random-walk behavior & 3 & 90 & 30\\
Nonlinear behavior & 6 & 180 & 60\\
Long-memory behavior & 2 & 60 & 20\\
Chaotic behavior & 4 & 120 & 40\\
Total & 20 & 600 & 200\\

\hline
\end{tabular}
\end{table}

\begin{table}[H]
\centering
\caption{The DGPs used to simulate time series with deterministic behavior.}
\label{tab5}
\begin{tabular}{ll}
\hline
\textbf{Process name} & \textbf{Process mathematical model} \\\hline
Trend process & $(T) : z_t=10+0.02t+\epsilon_t$ \\
Simple Seasonality process & $(SS) : z_t = 2\sin(2\pi t/5) +\epsilon_t$  \\
Complex Seasonality process & $(CS) : z_t = \sin(2\pi t/100)+0.5\sin(2\pi t/5) +\epsilon_t$  \\
Trend and Simple Seasonality process & $(TSS) :  z_t = 10+0.02t+5\sin(2\pi t/5)+\epsilon_t$  \\
Trend and Complex Seasonality process & $(TCS) : z_t = 10+0.02t+\sin(2\pi t/100)+0.5\sin(2\pi t/5)+\epsilon_t$ \\\hline
\end{tabular}
\end{table}

\begin{table}[H]
\centering
\caption{The DGPs used to simulate time series with nonlinear behavior.}
\label{tab8}
\begin{tabular}{ll}
\hline
\textbf{Process name} & \textbf{Process mathematical model} \\\hline
\multirow{3}{*}{Sign Auto-Regressive process} & 
$SAR(2) : z_t = sign(z_{t-1} + z_{t-2}) + \epsilon_{t}$ \\
& $sign(x) = \begin{cases}
               1 \;\; if \;\; x>0\\
                0 \;\; for \;\; x=0\\
                -1 \;\; for \;\; x<0
            \end{cases} $\\
Nonlinear Moving Average process  & $ NMA(2) : z_t = \epsilon_{t} - 0.3\epsilon_{t-1} + 0.2\epsilon_{t-2} + 0.4\epsilon_{t-1}\epsilon_{t-2} -0.25\epsilon_{t-2}^{2}$ \\
Nonlinear Auto-Regressive process & $ NAR(2) : z_t = \frac{0.7|z_{t-1}|}{|z_{t-1}|+2}+ \frac{0.35|z_{t-2}|}{|z_{t-2}|+2} + \epsilon_{t} $  \\
Bilinear process  & $ BL(2) : z_t = 0.4z_{t-1} - 0.3z_{t-2}+ 0.5z_{t-1}\epsilon_{t-1} + \epsilon_{t} $ \\
Smooth Transition Auto-Regressive process  & $ STAR(2) : z_t = 0.3z_{t-1} + 0.6z_{t-2}+\frac{0.1-0.9z_{t-1}+0.8z_{t-2}}{1+e^{-10z_{t-1}}}+\epsilon_{t} $ \\
Threshold Auto-Regressive process  & $TAR(2): z_t =
            \begin{cases}
                0.9z_{t-1}+0.05z_{t-2} +\epsilon_{t} \;\; for \;\; |z_{t-1}|\leq1\\
                -0.3z_{t-1}+0.65z_{t-2} -\epsilon_{t} \;\; for \;\; |z_{t-1}|>1
            \end{cases}$ \\
\hline
\end{tabular}
\end{table}

\begin{table}[H]
\centering
\caption{The DGPs used to simulate time series with random-walk behavior.}
\label{tab6}
\begin{tabular}{ll}
\hline
\textbf{Process name} & \textbf{Process mathematical model} \\\hline
Trend Random-Walk process & $ (TRW) : z_t=z_{t-1}+\epsilon_t $ \\
Seasonal Random-Walk process & $ (SRW) : z_t=z_{t-4}+\epsilon_t $ \\
Trend and Seasonal Random-Walk process & $ (TSRW) : z_t=z_{t-1}+z_{t-4}-z_{t-5}+\epsilon_t $  \\\hline
\end{tabular}
\end{table}

\begin{table}[H]
\centering
\caption{The DGPs used to simulate time series with long-memory behavior.}
\label{Long DGP}
\begin{tabular}{ll}
\hline
\textbf{Process name} & \textbf{Process mathematical model} \\\hline
$ARFIMA(p=2, d=0.2, q=2)$ & $z_{t}^{(0.2)}= 0.7z_{t-1}^{(0.2)} - 0.1z_{t-2}^{(0.2)} - 0.5\epsilon_{t-1} + 0.4\epsilon_{t-2} + \epsilon_{t}$  \\
$ARFIMA(p=2, d=0.4, q=2)$ & $z_{t}^{(0.4)}= 0.7z_{t-1}^{(0.4)} - 0.1z_{t-2}^{(0.4)} - 0.5\epsilon_{t-1} + 0.4\epsilon_{t-2} + \epsilon_{t}$  \\\hline
\end{tabular}
\end{table}

\begin{table}[H]
\centering
\caption{The DGPs used to simulate time series with chaotic behavior.}
\label{tab7}
\begin{tabular}{ll}
\hline
\textbf{Process name} & \textbf{Process mathematical model} \\\hline
\multirow{1}{*}{Mackey-Glass process} &
$\frac{dx} {dt}=a \frac{x(t-\pi)} {1+x^{c}(t-\pi)} - bx(t)$ such that $(\tau=17, a=0.2, b=0.1, c=10)$ \citep{ma2007chaotic}\\

\multirow{1}{*}{Hénon-Map process} &
$x_{t+1} = 1+y_{t}-a{x_{t}}^{2} \; ;\; y_{t+1} = bx_{t}$ such that $(a=1.4, b=0.3)$ \citep{li2016new} \\

\multirow{2}{*}{Rössler process} &
$\dot{x} = -y-z\; ;\; \dot{y} = x+ay\; ;\; \dot{z} = b+z(x-c)$ \\
 & $ (a=0.15, b=0.2, c=10, x_{0}=10, y_{0}=z_{0}=0) $ \citep{lim2007chaotic} \\
 
\multirow{2}{*}{Lorenz process} &
$\dot{x} = \sigma(y-x)\; ;\; \dot{y} = -xz+rx-y\; ;\; \dot{z} = xy-bz$ \\
 & $(\sigma=16, r=45.92, b=4, x_{0}=y_{0}=z_{0}=1)$ \citep{lim2007chaotic}\\
\hline
\end{tabular}
\end{table}

\subsection{RNN-cells used for experiments 1 and 2}
\label{section_ARCH}

To evaluate each cell structure with respect to each times series behavior, we conducted two experiments as summarized in Table \ref{RNNs evaluated}. The first experiment evaluates LSTM-Vanilla and 11 of its variants created based on one alteration in the basic Vanilla architecture that consists of (1) removing, (2) adding, or (3) substituting one cell component (Table \ref{tab1}):
(1) The first three variants NIG (No Input Gate), NFG (No Forget Gate), and NOG (No Output Gate) were created through the deactivation of the input gate, the forget gate, and the output gate, respectively. The four subsequent variants NIAF (No Input Activation Function), NFAF (No Forget Activation Function), NOAF (No Output Activation Function), and NCAF (No Candidate Activation Function) were constructed through the elimination of the input, forget, output, and candidate activation function, respectively. 
(2) The two subsequent variants PC (Peephole Connections), and FGR (Full Gate Recurrence) were designed through the creation of new connections between the cell state and the gates, and between the current states and the previous states of the gates, respectively. 
(3) Eventually, FB1 (Forget Gate Bias 1) and CIFG (Coupled Input Forget Gate) was conceived by setting the forget gate bias to one, and by coupling the input and the forget gate into one gate, respectively. 

The second experiment  evaluates and analyzes the performance of 20 possible RNN-cell structures: \\ JORDAN, ELMAN, MRNN, SCRN, IRNN, LSTM-Vanilla, GRU, MGU, MUT1, MUT2, MUT3, and 9 SLIM variants mapping LSTM, GRU, and MGU. A summary of the evaluated cells related to each experiment is presented in Table \ref{RNNs evaluated}, and the cellular calculations inside each cell are presented in Table \ref{tab2}.

\begin{table}[H]
\centering
\caption{List of evaluated RNN models with regard to each experiment with their theoretic complexity, number of weight matrices, and number of bias vectors. $n_I$ is the number of inputs, $n_H$ is the number of hidden nodes, $n_S$ is the number of context nodes, and $n_O$ is the number of outputs.}
\label{RNNs evaluated}
\resizebox{0.9\textwidth}{!}{%
\begin{tabular}{cllccc}
\cline{2-6}
\multicolumn{1}{l}{} & \multicolumn{5}{c}{\bf RNN models} \\ 
\cline{2-6} 
\multicolumn{1}{l}{} & \multicolumn{1}{c}{\bf Short name} & \multicolumn{1}{c}{\bf Full name} &
\multicolumn{1}{c}{\bf Theoretic complexity} &
\makecell[c]{\bf \# Weight\\ \bf matrices} &
\makecell[c]{\bf \# Bias \\ \bf vectors}
\\ \hline
\multicolumn{1}{c}{\multirow{12}{*}{\bf Experiment 1}} & \multicolumn{1}{l}{NIG} & LSTM with No Input Gate & $3n_In_H + 3n_H^2 + 3n_H$ & 6 & 3 \\ 
\cline{2-6} 
\multicolumn{1}{c}{} & \multicolumn{1}{l}{NFG} & LSTM with No Forget Gate & $3n_In_H + 3n_H^2 + 3n_H$ & 6 & 3\\ 
\cline{2-6} 
\multicolumn{1}{c}{} & \multicolumn{1}{l}{NOG} & LSTM with No Output Gate & $3n_In_H + 3n_H^2 + 3n_H$ & 6 & 3\\ 
\cline{2-6} 
\multicolumn{1}{c}{} & \multicolumn{1}{l}{CIFG} & LSTM with Coupled Input Forget Gate & $3n_In_H + 3n_H^2 + 3n_H$ & 6 & 3\\ 
\cline{2-6} 
\multicolumn{1}{c}{} & \multicolumn{1}{l}{FB1} & LSTM with Forget Gate Bias 1 & $4n_In_H + 4n_H^2 + 3n_H$ & 6 & 3\\ 
\cline{2-6} 
\multicolumn{1}{c}{} & \multicolumn{1}{l}{NIAF} & LSTM with No Input Activation Function & $4n_In_H + 4n_H^2 + 4n_H$ & 8 & 4\\ 
\cline{2-6} 
\multicolumn{1}{c}{} & \multicolumn{1}{l}{NFAF} & LSTM with No Forget Activation Function & $4n_In_H + 4n_H^2 + 4n_H$ & 8 & 4\\ 
\cline{2-6} 
\multicolumn{1}{c}{} & \multicolumn{1}{l}{NOAF} & LSTM with No Output Activation Function & $4n_In_H + 4n_H^2 + 4n_H$ & 8 & 4\\ 
\cline{2-6} 
\multicolumn{1}{c}{} & \multicolumn{1}{l}{NCAF} & LSTM with No Candidate Activation Function & $4n_In_H + 4n_H^2 + 4n_H$ & 8 & 4\\ 
\cline{2-6} 
\multicolumn{1}{c}{} & \multicolumn{1}{l}{Vanilla} & LSTM Vanilla & $4n_In_H + 4n_H^2 + 4n_H$ & 8 & 4\\ 
\cline{2-6} 
\multicolumn{1}{c}{} & \multicolumn{1}{l}{PC} & LSTM with Peephole Connections & $4n_In_H + 7n_H^2 + 4n_H$ & 11 & 4\\ 
\cline{2-6} 
\multicolumn{1}{c}{} & \multicolumn{1}{l}{FGR} & LSTM with Full Gate Recurrence & $4n_In_H + 13n_H^2 + 4n_H$ & 17 & 4 \\ 
\hline
\multicolumn{1}{c}{\multirow{20}{*}{\bf Experiment 2}} & \multicolumn{1}{l}{ELMAN} & ELMAN & $n_In_H + n_H^2 + n_H$ & 2 & 1\\ 
\cline{2-6} 
\multicolumn{1}{c}{} & \multicolumn{1}{l}{IRNN} & Identity Recurrent Neural Network & $n_In_H + n_H^2 + n_H$ & 2 & 1 \\ 
\cline{2-6} 
\multicolumn{1}{c}{} & \multicolumn{1}{l}{JORDAN} & JORDAN & $n_In_H + n_On_H + n_H$ & 2 & 1\\ 
\cline{2-6} 
\multicolumn{1}{c}{} & \multicolumn{1}{l}{MRNN} & Multi-Recurrent Neural Network & $n_In_H + n_H^2 + n_On_H + n_H$ & 3 & 1\\ 
\cline{2-6} 
\multicolumn{1}{c}{} & \multicolumn{1}{l}{SCRN} & Structurally Constrained Recurrent Network & $n_In_S + n_In_H + n_H^2 + n_Sn_H + n_H$ & 4 & 1\\ 
\cline{2-6} 
\multicolumn{1}{c}{} & \multicolumn{1}{l}{MGU-SLIM3} & Minimal Gate Unit SLIM3 & $n_In_H + n_H^2 + 2n_H$ & 2 & 2\\ 
\cline{2-6} 
\multicolumn{1}{c}{} & \multicolumn{1}{l}{MGU-SLIM2} & Minimal Gate Unit SLIM2 & $n_In_H + 2n_H^2 + n_H$ & 3 & 1\\ 
\cline{2-6} 
\multicolumn{1}{c}{} & \multicolumn{1}{l}{MGU-SLIM1} & Minimal Gate Unit SLIM1 & $n_In_H + 2n_H^2 + 2n_H$ & 3 & 2\\ 
\cline{2-6} 
\multicolumn{1}{c}{} & \multicolumn{1}{l}{MGU} & Minimal Gate Unit & $2n_In_H + 2n_H^2 + 2n_H$ & 4 & 2\\ 
\cline{2-6} 
\multicolumn{1}{c}{} & \multicolumn{1}{l}{GRU-SLIM3} & Gated Recurrent Unit SLIM3 & $n_In_H + n_H^2 + 3n_H$ & 2 & 3\\ 
\cline{2-6} 
\multicolumn{1}{c}{} & \multicolumn{1}{l}{GRU-SLIM2} & Gated Recurrent Unit SLIM2 & $n_In_H + 3n_H^2 + n_H$ & 4 & 1\\ 
\cline{2-6} 
\multicolumn{1}{c}{} & \multicolumn{1}{l}{GRU-SLIM1} & Gated Recurrent Unit SLIM1 & $n_In_H + 3n_H^2 + 3n_H$ & 4 & 3\\ 
\cline{2-6} 
\multicolumn{1}{c}{} & \multicolumn{1}{l}{MUT1} & Gated Recurrent Unit Mutation 1 & $2n_In_H + 2n_H^2 + 3n_H$ & 4 & 3\\ 
\cline{2-6} 
\multicolumn{1}{c}{} & \multicolumn{1}{l}{MUT2} & Gated Recurrent Unit Mutation 2 & $2n_In_H + 3n_H^2 + 3n_H$ & 5 & 3\\ 
\cline{2-6} 
\multicolumn{1}{c}{} & \multicolumn{1}{l}{MUT3} & Gated Recurrent Unit Mutation 3 & $3n_In_H + 3n_H^2 + 3n_H$ & 6 & 3\\ 
\cline{2-6} 
\multicolumn{1}{c}{} & \multicolumn{1}{l}{GRU} & Gated Recurrent Unit & $3n_In_H + 3n_H^2 + 3n_H$ & 6 & 3\\ \cline{2-6} 
\multicolumn{1}{c}{} & \multicolumn{1}{l}{LSTM-SLIM3} & LSTM SLIM3 & $n_In_H + n_H^2 + 4n_H$ & 2 & 4\\ 
\cline{2-6} 
\multicolumn{1}{c}{} & \multicolumn{1}{l}{LSTM-SLIM2} & LSTM SLIM2 & $n_In_H + 4n_H^2 + n_H$ & 5 & 1\\ 
\cline{2-6} 
\multicolumn{1}{c}{} & \multicolumn{1}{l}{LSTM-SLIM1} & LSTM SLIM1 & $n_In_H + 4n_H^2 + 4n_H$ & 5 & 4\\ 
\cline{2-6} 
\multicolumn{1}{c}{} & \multicolumn{1}{l}{LSTM-Vanilla} & LSTM Vanilla & $4n_In_H + 4n_H^2 + 4n_H$ & 8 & 4\\ 
\hline
\end{tabular}%
}
\end{table}

The studied RNN cells have different degrees of complexity, which is referred to as theoretic complexity (Table \ref{RNNs evaluated}). This complexity is defined by the number of parameters inside each cell which depends on the number of inputs, the number of hidden nodes, the number of context nodes (in the case of the SCRN model), and the number of outputs. During the hyperparameter tuning the complexity of the cell may increase or decrease depending on the optimal number of hidden nodes found. Thus, the complexity of the cell defined after the hyperparameter tuning is called empirical complexity.  

\subsection{Data preparation}
\label{Data preparation} 
\begin{figure}[H]
  \begin{center}
    \includegraphics[height=4cm,width=10cm]{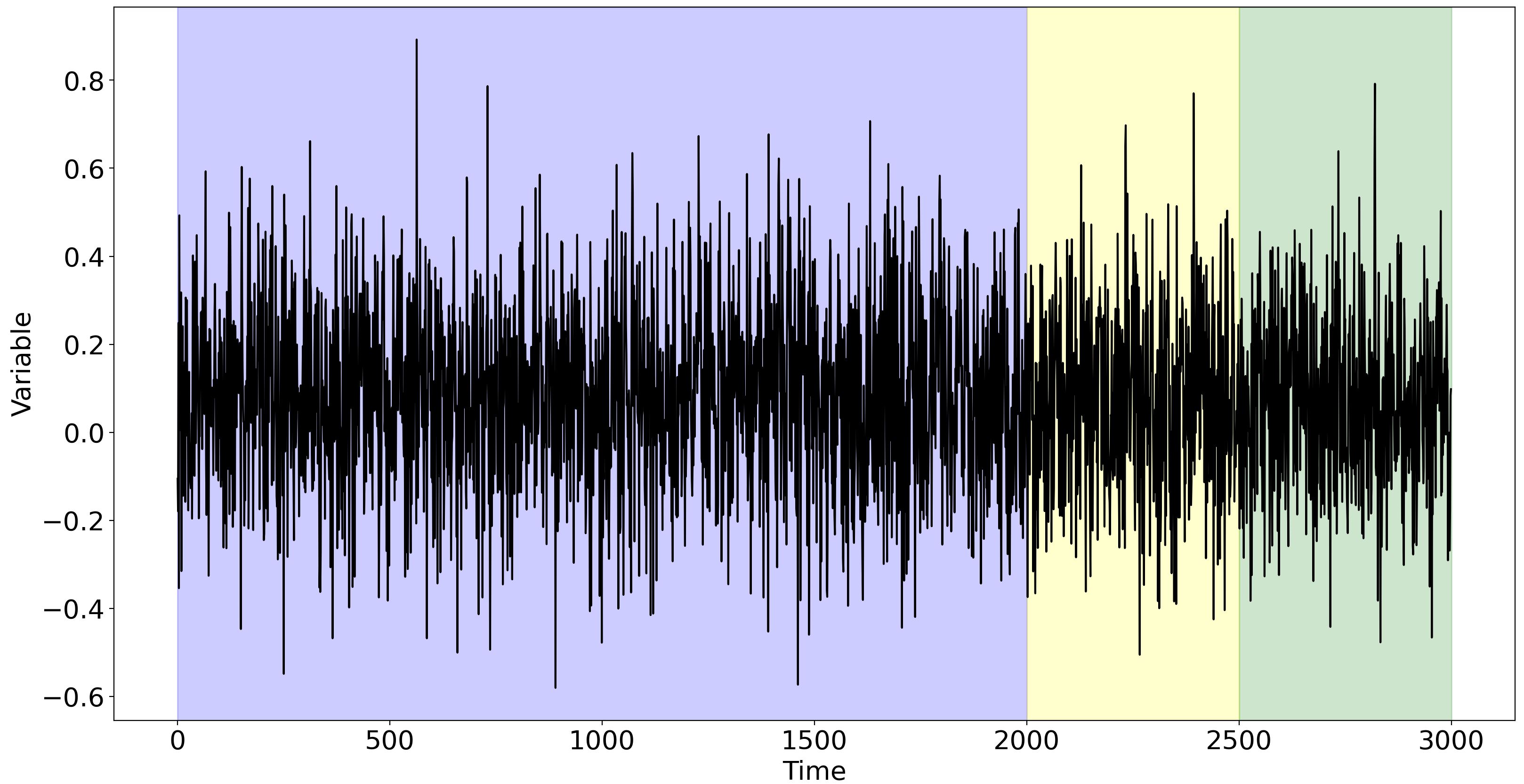}
  \end{center}
  \caption{Visualization of data partitioning into a training set (purple color), validation set (yellow color), and test set (green color). The plotted time series is a sample of NMA process.}
  \label{data_partition}
\end{figure}

\begin{figure}[H]
  \begin{center}
    \includegraphics[height=4cm,width=15cm]{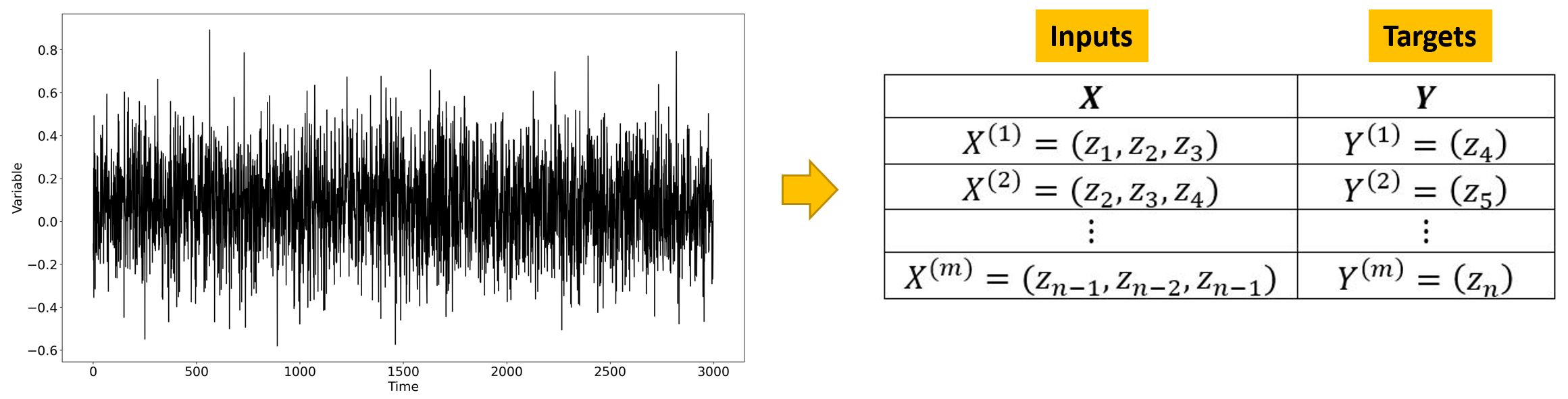}
  \end{center}
  \caption{Visualization of data structuring from raw to tabular data. In this case, the number of time steps $T=3$ and the number of horizons $H=1$.}
  \label{data_structuring}
\end{figure}

Before starting the modeling process, each time series data was partitioned into three subsets: the first 2000 observations were used to train the models in order to find the best parameters, the next 500 observations were used to select the best configuration of hyperparameters of each model, and the last 500 observations were used to test the out-of-sample performance of these models (Figure \ref{data_partition}). 
Each of these partitions was normalized, then converted from unstructured into structured data (Figure \ref{data_structuring}) by reshaping it based on the number of time steps $T$ (estimation window size) and the number of horizons $H$ (forecasting window size). Thus, we converted the raw time series data $(Z_t)_{t\in \mathbb{N}} = (z_1, z_2, ..., z_n)$
into a structured data $S = \{(X^{(i)}, Y^{(i)}) , i \in \{1, ..., m\}\} = S_{train} \cup S_{validation} \cup S_{test}$.
Where $X^{(i)} = (z_i, z_{(i+1)}, \ldots , z_{(i+T-1)})$ and $Y^{(i)} = (z_{(i+T)})$ because we are handling one-time-step-ahead forecasting.

\subsection{Parameter optimization}
\label{Parameter optimization} 
To find the best set of parameters (weight $W$ and bias $b$), first, the model parameters were initialized by Kaiming method \citep{he2015delving}, then optimized using a stochastic gradient descent-based algorithm named Adam \citep{kingma2014adam} (Algorithm \ref{alg1}) by minimizing the cost function $J$ that describes a Mean Square Error (MSE) using a mini-batch size of 100 (Table \ref{tab10}). 
\[J = \frac{1}{|S_{train}|} \sum_{i=1}^{|S_{train}|} (Y^{(i)} - \hat{Y}^{(i)})^{2} \]

\[ \hat{Y}^{(i)} =  h_{T}.W_{hy} + b_{y} \]

Where $\hat{Y}^{(i)}$ is the predicted value of instance $i$ computed using the identity output activation function, $W_{hy}$ is the weight matrix between the hidden and the output layer, and $b_{y}$ is the output layer bias.

\begin{center}
\begin{minipage}{1.0\linewidth}
    \begin{algorithm}[H]
    \caption{Description of the parameter optimization process using Adam algorithm.}\label{alg1}
    \begin{algorithmic}[1]
        \REQUIRE The learning rate $\alpha=0.01$
        \REQUIRE The exponential decay rates for the moments estimates $\beta_1=0.9$, $\beta_2=0.999$, and $\epsilon=10^{-8}$
        \REQUIRE The cost function $J$ with parameters $W$ and $b$ to be minimized.
        \STATE $V_{dW}=0$, $V_{db}=0$, $S_{dW}=0$, and $S_{db}=0$
        \FOR {each iteration $t$}
            \STATE Compute the $dW$ and $db$:
            \STATE $dW=\frac{\partial J}{\partial W}$ and $db=\frac{\partial J}{\partial b}$
            \STATE Update the biased first moments estimates $V_{dW}$ and $V_{db}$:
            \STATE $V_{dW} \leftarrow \beta_{1} V_{dW} + (1-\beta_{1})d_W$ and $V_{db} \leftarrow \beta_{1} V_{db} + (1-\beta_{1})d_b$
            \STATE Update the biased second moments estimates $S_{dW}$ and $S_{db}$:
            \STATE $S_{dW} \leftarrow \beta_{2} S_{dW} + (1-\beta_{2})(d_W)^2$ and $S_{db} \leftarrow \beta_{2} S_{db} + (1-\beta_{2})(d_b)^2$
            \STATE Update the bias-corrected first moments estimates $v^{corr}_{dW}$ and $v^{corr}_{db}$:
            \STATE $v^{corr}_{dW} = \frac{V_{dW}}{(1-\beta^{t}_{1})}$ and $v^{corr}_{db} = \frac{V_{db}}{(1-\beta^{t}_{1})}$
            \STATE Update the bias-corrected second moments estimates $S^{corr}_{dW}$ and $S^{corr}_{db}$:
            \STATE $S^{corr}_{dW} = \frac{S_{dW}}{(1-\beta^{t}_{2})}$ and $S^{corr}_{db} = \frac{S_{db}}{(1-\beta^{t}_{2})}$
            \STATE Update the parameters $W$ and $b$:
            \STATE $W \leftarrow W - \alpha \times \frac{v^{corr}_{dW}}{\sqrt{S^{corr}_{dW}}+\epsilon}$ and $b \leftarrow b - \alpha \times \frac{v^{corr}_{db}}{\sqrt{S^{corr}_{db}}+\epsilon}$
        \ENDFOR
    \end{algorithmic}
    \end{algorithm}
\end{minipage}
\par
\end{center}

\subsection{Hyperparameter tuning}
\label{Hyperparameter selection}
Within this section, we present the approach used to select the best architecture for each model. This architecture is defined by two hyperparameters: 1) the number of time steps $T$, and 2) The number of hidden neurons $n_H$. To find the best combination $\{T, n_H\}$ we used a Grid Search algorithm (Algorithm \ref{alg2}) that loops over each possible combination in a grid created in two-dimensional space defined by $T$ and $n_H$. The other hyperparameters such as the initial learning rate and the minibatch size were set based on the literature \citep{parmezan2019evaluation} (Table \ref{tab10}). 

\begin{center}
\begin{minipage}{1.0\linewidth}
    \begin{algorithm}[H]
    \caption{Description of the hyperparameter tuning process.}\label{alg2}
    \begin{algorithmic}[1]
        \STATE $Best\_Error = +\infty$
        \STATE $Best\_Config = None$
        \FOR{$T$ in 1 to $T_{max}$}
            \FOR{$n_{H}$ in 1 to $n_{H_{max}}$}
                \STATE Define an empty array $Errors$.
                \FOR{$i$ in 1 to 10}
                    \STATE Train the model with the combination $\{T, n_{H}\}$ using Algorithm \ref{alg1}. 
                    \STATE Compute the $MSE$ of the model on the validation set.
                    \STATE Append this error to the array $Errors$.
                \ENDFOR
                \STATE Compute the average overall $MSE$ values in the array $Errors$.
                \IF{$MSE < Best\_Error$}
                    \STATE $Best\_Error = MSE$ 
                    \STATE $Best\_Config = \{T, n_{H}\}$
                \ENDIF
            \ENDFOR
        \ENDFOR
    \end{algorithmic}
    \end{algorithm}
\end{minipage}
\par
\end{center}

Using each combination $\{T, n_H\}$, the model was trained ten times for a maximum number of 500 epochs in which we save the best set of parameters that minimizes the average MSE over the ten runs in the validation set (Algorithm \ref{alg2}). The objective behind this is to avoid the bias generated by the parameter initialization method and also to select the most stable model’s architecture.

The maximum number $n_{H{max}}$ of evaluated hidden neurons is 10 because we are using univariate time series (Table \ref{tab10}), while the maximum number $T_{max}$ of evaluated time steps is described in Table \ref{tab11} for each behavior.
To define $T_{max}$ used in the chaotic time series, we used the values recommended by \citep{parmezan2019evaluation} since we used their chaotic datasets. 
In the long-memory behavior, $T_{max}$ was selected based on the last relevant peak visualized in the ACF (Auto Correlation Function) graph and the PACF (Partial Auto Correlation Function) graph. 
In the Nonlinear behavior, we already know that the time series were generated with an order equal to 2, but in order to find the best number of time steps, we set $T_{max}=5$ for two reasons: 1) we need to simulate the fact that the model is not aware of the DGP, which is a mandatory condition to use machine learning models. 2) we cannot use the ACF and PACF tests in this case because they can only capture the linear correlations in the data, however, we are using nonlinear data which necessarily contains nonlinear correlations.
In the deterministic behavior, we used $T_{max}=10$ for the T DGP, and $T_{max}=5$ for the remaining DGPs using the seasonality period.
In the random-walk behavior, we used $T_{max}=5$ for the TSRW DGP, $T_{max}=4$ for the SRW DGP, and $T_{max}=10$ for the TRW DGP. 

\begin{table}[H]
\centering
\caption{The values of the hyperparameters used to train the studied RNN models.}
\label{tab10}
\begin{tabular}{lr}
\hline
\textbf{Hyperparameter} & \textbf{Value} \\\hline
Mini-batch size & $100$ \\
Maximum number of epochs & $500$ \\
Learning rate $\alpha$ & $0.01$ \\
Optimization algorithm & Adam \\
Cost function & MSE \\
Horizon $H$ & 1 \\
Maximum number of hidden neurons $n_{H{max}}$ & 10 \\
\hline
\end{tabular}
\end{table}

\begin{table}[H]
\centering
\caption{The $T_{max}$ values used for each DGP of each time series behavior.}
\label{tab11}
\begin{tabular}{llc}
\hline 
\textbf{Behavior} & \textbf{DGP} & \textbf{\makecell{$T_{max}$}} \\\hline
\multirow{5}{*}{\makecell[cl]{Deterministic behavior}} 
 & T & 10 \\
 & SS & 5 \\
 & CS & 5 \\
 & TSS & 5 \\
 & TCS & 5 \\\hline
\multirow{3}{*}{\makecell[cl]{Random-walk behavior}} 
 & TRW & 10 \\
 & SRW & 4 \\
 & TSRW & 5 \\\hline
\multirow{6}{*}{Nonlinear behavior}
 & SAR(2) & 5 \\
 & NMA(2) & 5 \\
 & NAR(2) & 5 \\
 & BL(2) & 5 \\
 & STAR(2) & 5 \\
 & TAR(2) & 5 \\\hline
\multirow{2}{*}{Long-memory behavior} 
 & ARFIMA(2, 0.2, 2) & 20 \\
 & ARFIMA(2, 0.4, 2) & 40 \\\hline
 \multirow{4}{*}{Chaotic behavior} 
 & Mackey-Glass & 7 \\
 & Hénon map & 3 \\
 & Rössler & 14 \\
 & Lorenz & 25 \\
\hline
\end{tabular}
\end{table}

The hyperparameter tuning was performed using the first replicate for each DGP because changing the white noise generator to create these 30 replicates does not change the linear/nonlinear correlations in the time series, it only impacts the randomness of the data. Therefore, the combination ${T, n_H}$ remains appropriate for all the 30s replicates. 
In total, we performed 200 experiments for each model (Table \ref{tab4}) which sums up to 6200 experiments in the whole study.
Once the best architecture of each model was found, the training and the validation sets were merged to form a new training set on which the best model's configuration is retrained for 500 epochs, then evaluated on the test set.

\subsection{Performance evaluation}
\label{Performance evaluation}
To evaluate the forecasting performance of the models, a set of statistical metrics were used. These statistical metrics together provide assistance to compare the models and select the best one. The used metrics are classified into four subsets: 

\begin{itemize}
\setcounter{enumi}{0}
\item Error-based metrics: these metrics only measure the number of errors made by a predictive model by evaluating the goodness-of-fit of the model. Two metrics were used: the Mean Absolute Error (MAE) and the Root Mean Square Error (RMSE). The RMSE measures the standard deviation of residuals, it penalizes large errors. While the MAE measures the average magnitude of the residuals, it is less prone to outliers.

\[RMSE = \sqrt{\frac{1}{|S_{test}|} \sum_{i=1}^{|S_{test}|} (Y^{(i)} - \hat{Y}^{(i)})^{2}} \]

\[MAE = \frac{1}{|S_{test}|} \sum_{i=1}^{|S_{test}|} |Y^{(i)} - \hat{Y}^{(i)}| \]

\item Information criterion-based metrics: these metrics deal with the trade-off between the goodness-of-fit of the model and the simplicity of the model. They penalize complex models by combining the number of errors committed by the model, the number of parameters (i.e., weights and biases) employed by the model to generate the output, and the sample size. They measure the amount of information lost by a model. A low value of these metrics implies less information loss, therefore high model quality. A set of five main statistical criteria were employed: Akaike Information Criterion (AIC) \citep{akaike1969fitting}, Bayesian Information Criterion (BIC) \citep{findley1991counterexamples}, Amemiya Prediction Criterion (APC) \citep{amemiya1980selection}, Hocking's Sp (HSP) \citep{hocking1976biometrics}, and Sawa's Bayesian Information Criterion (SBIC) \citep{sawa1978information}. 

\[AIC = |S_{test}| \ln(\frac{SSE}{|S_{test}|}) + 2K \]

\[BIC = |S_{test}| \ln(\frac{SSE}{|S_{test}|}) + K \ln(|S_{test}|) \]

\[APC = \frac{|S_{test}|+K}{|S_{test}|(|S_{test}|-K)}SSE \]

\[HSP = \frac{SSE}{|S_{test}|(|S_{test}|-K-1)} \]

\[SBIC = |S_{test}| \ln(\frac{SSE}{|S_{test}|}) + 2(K+2)q + 2q^{2} \]

\[q = |S_{test}| \frac{\sigma^{2}}{SSE}\]

Where $SSE$ is the sum square error, $K$ is the number of parameters used by the model (empirical complexity), and $\sigma$ is the standard deviation of the prediction errors.

\[SSE = \sum_{i=1}^{|S_{test}|} (Y^{(i)} - \hat{Y}^{(i)})^{2} \]

\item Naïve-based metric: this metric compares the performance of the predictive model with the naïve model. It is called Theil’s U (TU) coefficient \citep{parmezan2019evaluation}. The naïve model assumes that the best value at time $t+1$ is the value obtained at time $t$. The values of this metric can be interpreted based on different ranges as follows: if $(TU > 1)$, the model's performance is lower than the naïve model. If $(TU = 1)$, the model's performance is the same as the naïve model. If $(TU < 1)$, the model's performance is higher than the naïve model. If $(TU <= 0.55)$, the model is trusted to carry out future predictions.

\[TU = \frac{SSE}{\sum_{i=1}^{|S_{test}|} (Y^{(i)} - Y^{(i-1)})^{2}}\]

\item Direction change-based metric: this metric is called Prediction Of Change In Direction (POCID) \citep{parmezan2019evaluation}. It measures the accuracy in the direction changes. It accumulates, over time, the differences in the direction change between the predicted and the observed values. It penalizes the model when its direction change in the predicted values between two consecutive time steps ($i-1$ and $i$) is different than those of the observed values. This metric is complementary to analyzing the prediction errors. A high value of this metric implies a high similarity degree in the direction changes between the predicted values and the observed values. 

\[POCID = 100 \times \frac{1}{|S_{test}|}\sum_{i=1}^{|S_{test}|} D^{(i)}\]

\[D^{(i)} = \begin{cases}
            1 \;\; if \;\; (\hat{Y}^{(i)} - \hat{Y}^{(i-1)})(Y^{(i)} - Y^{(i-1)})>0\\
            0 \;\; otherwise \;\; 
        \end{cases}\]

\end{itemize}

All the aforementioned metrics were computed by averaging their values over the 30 replicates of the test set of all the DGPs of one specific behavior.

Choosing the best model can be very challenging when different performance metrics are available. To facilitate this task, a new metric was created using the combination of all the aforementioned metrics. We call this metric Multi-Criteria Index Measure (MCIM). 
Unlike the error measures, which generate values that need to be minimized, the POCID index must be maximized. Therefore, a complement of the POCID was created called Error Rate (ER):

\[ER = 100 - POCID\]

Afterward, the MCIM metric was defined by computing the average over all the normalized values of the used metrics. The normalization was used because the performance metrics have different ranges of values. The normalization was performed using the Min-Max method.

\[MCIM(M) = \frac{1}{|S_{PI(M)}|}\sum_{i=1}^{|S_{PI(M)}|} PI^{(n)}_{i}(M)\]

\[S_{PI(M)} = \{MAE(M), RMSE(M), AIC(M), BIC(M), APC(M), HSP(M), SBIC(M), TU(M), ER(M)\}\]

\[PI^{(n)}_{i}(M) = \frac{PI_{i}(M) - PI^{(min)}_{i/S_M}}{PI_{i}(M) - PI^{(max)}_{i/S_M}}\]

Where $S_{PI(M)}$ is the set of values of the Performance Indices of model $M$, $PI_{i}(M)$ is the performance index type $i$ of model $M$, and $PI^{(n)}_{i}(M)$ is its normalized value.
 
\[PI^{(min)}_{i/S_M} = \min\{PI_{i}(M) : M=1,\dots, |S_M|\}\]
\[PI^{(max)}_{i/S_M} = \max\{PI_{i}(M) : M=1,\dots, |S_M|\}\]

$S_M$ is the set of evaluated models, $PI^{(min)}_{i/S_M}$ is the minimum value of the performance index type $i$ over the set of evaluated models $S_M$, and $PI^{(max)}_{i/S_M}$ is the maximum value of the performance index type $i$ over the set of evaluated models $S_M$.

The created MCIM metric was employed to rank all the evaluated RNN architectures and to select the best RNN structure for each specific time series behavior. 
However, relying solely on the MCIM metric may not be statistically reliable to select the best model. To quantify the likelihood of a model being the most performing, to improve our confidence in the models' interpretation and selection, to compare the performance of the RNN models, and to further examine whether any observed difference is statistically significant or it is only due to noise or chance, statistical significance tests should be used. 

The choice of these tests depends on (1) the prediction task (classification or regression), (2) the data distribution, and (3) whether the models are compared on the same data or not. 
In our case, (1) we are solving forecasting tasks which is a type of dynamical regression, (2) we assume that the distribution of our data is unknown, and (3) all the models are tested on the same data. Therefore, a regression non-parametric paired statistical test referred to as Friedman Wilcoxon-Holm signed-rank test \footnote[1]{This statistical test was performed using the framework \href{https://github.com/hfawaz/cd-diagram}{\blue{https://github.com/hfawaz/cd-diagram}} re-adapted to test forecasting models} was used to detect pairwise significance. 

First, the Friedman test \citep{friedman1940comparison} between each pair of models is performed with a significance level of 5\% to reject the null hypothesis (p-value < 0.05). The null hypothesis of this test states that the pair of models are similar.
Then, a pairwise post-hoc analysis based on the Wilcoxon-Holm method (with Holm’s alpha 5\% correction) was used to compare the results \citep{benavoli2016should, wilcoxon1992individual, holm1979simple, garcia2008extension, abdulkarim2016time}.
To visualize this pairwise comparison and to highlight the difference significance, a critical difference diagram proposed by \citep{demvsar2006statistical} was generated where a thick horizontal line groups a set of RNN models that are not significantly different.

\section{Results and discussion} \label{S6}
In this section, we  present the results of the two conducted experiments: (1) The first experiment consists of evaluating and analyzing the role of each component in the LSTM-Vanilla cell with respect to the five time series behaviors. The evaluated architectures were generated by removing (NIG, NFG, NOG, NIAF, NFAF, NOAF, and NCAF), adding (PC and FGR), or substituting (FB1 and CIFG) one cell component. (2) The second experiment aims at evaluating and analyzing the performance of a multitude of RNN cell structures available in the literature (JORDAN, ELMAN, MRNN, SCRN, IRNN, LSTM-Vanilla, GRU, MGU, MUT1, MUT2, MUT3, and 9 SLIM variants) in forecasting the five behaviors. 

\subsection{ \bf Experiment 1: Utility analysis of LSTM cell components in forecasting time series behaviors.} \label{S6-1}

The impact of each component (i.e., input gate, forget gate, output gate, coupled input-forget gate, input activation function, forget activation function, output activation function, candidate activation function, fixing the forget bias to 1, peephole connections, and full gate recurrence) on the performance of LSTM-Vanilla model for predicting the deterministic, random-walk, nonlinear, long-memory, and chaotic behaviors are presented in this section.

Tables \ref{T_D1} to \ref{T_C1} present the average results of the 10 used statistical metrics for each variant of the LSTM-Vanilla model run on the test set of the five types of time series behaviors. It can be noticed that the error-based metrics select models with low residual values, however, the information criterion-based metrics tend to select less complex models with low forecasting errors. Based on the new proposed MCIM metric, the structure CIFG is the most adapted to forecast time series data with deterministic (Table \ref{T_D1}) and random-walk (Table \ref{T_R1}) behaviors. The NOG variant outperforms the other models in forecasting data with nonlinear behavior (Table \ref{T_N1}). The NIG design is more adapted for data with long-memory behavior (Table \ref{T_L1}). The Vanilla structure reveals the higher ability to forecast time series data with chaotic behavior (Table \ref{T_C1}). 

\begin{figure}[H]
     \centering
     \begin{subfigure}[b]{0.5\textwidth}
         \centering
         \includegraphics[height=5.5cm,width=8cm]{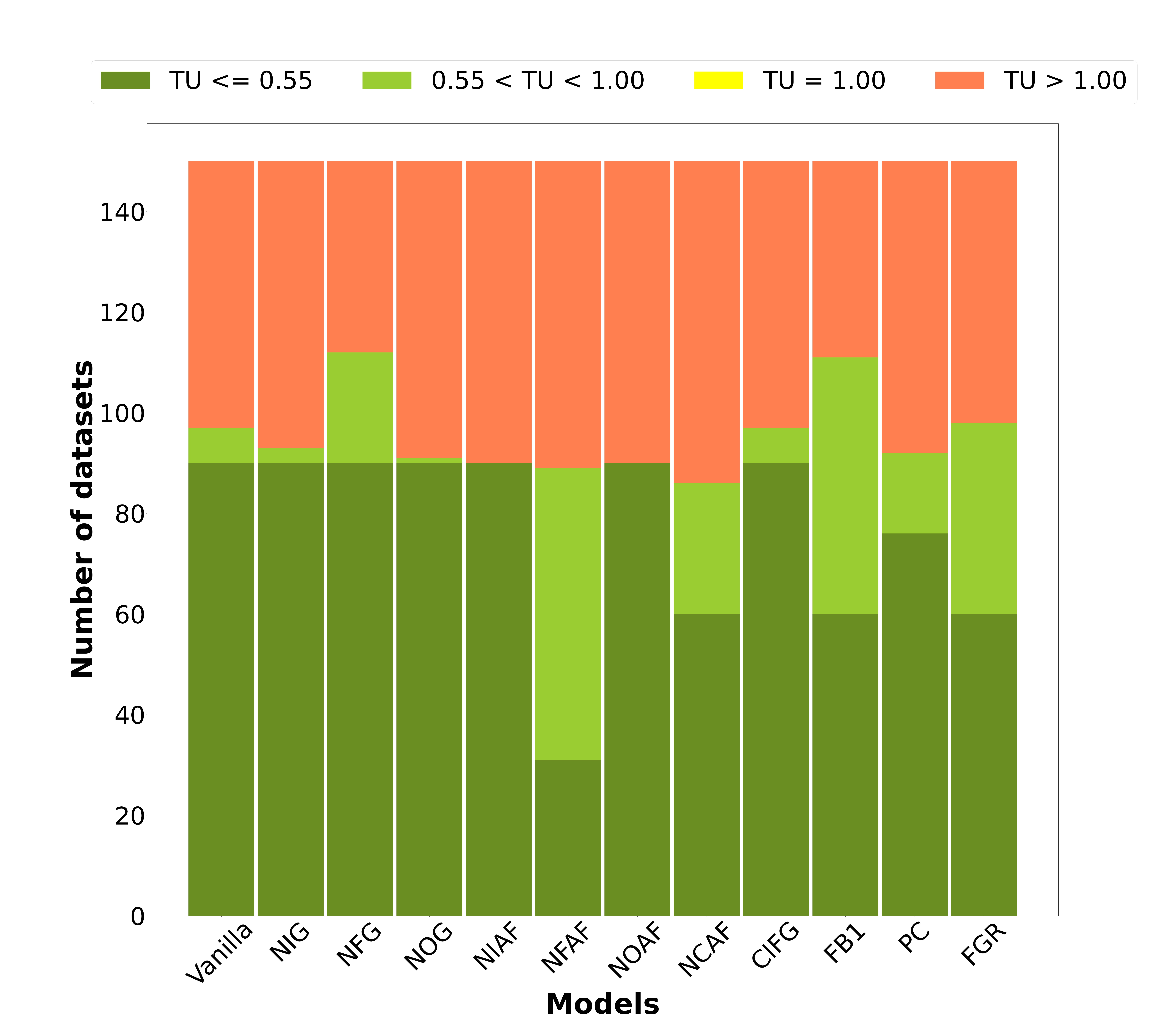}
         \caption{Distribution of four ranges of TU values.}
         \label{F_D1_a}
     \end{subfigure}
     \begin{subfigure}[b]{0.4\textwidth}
         \centering
         \includegraphics[height=5.3cm,width=8cm]{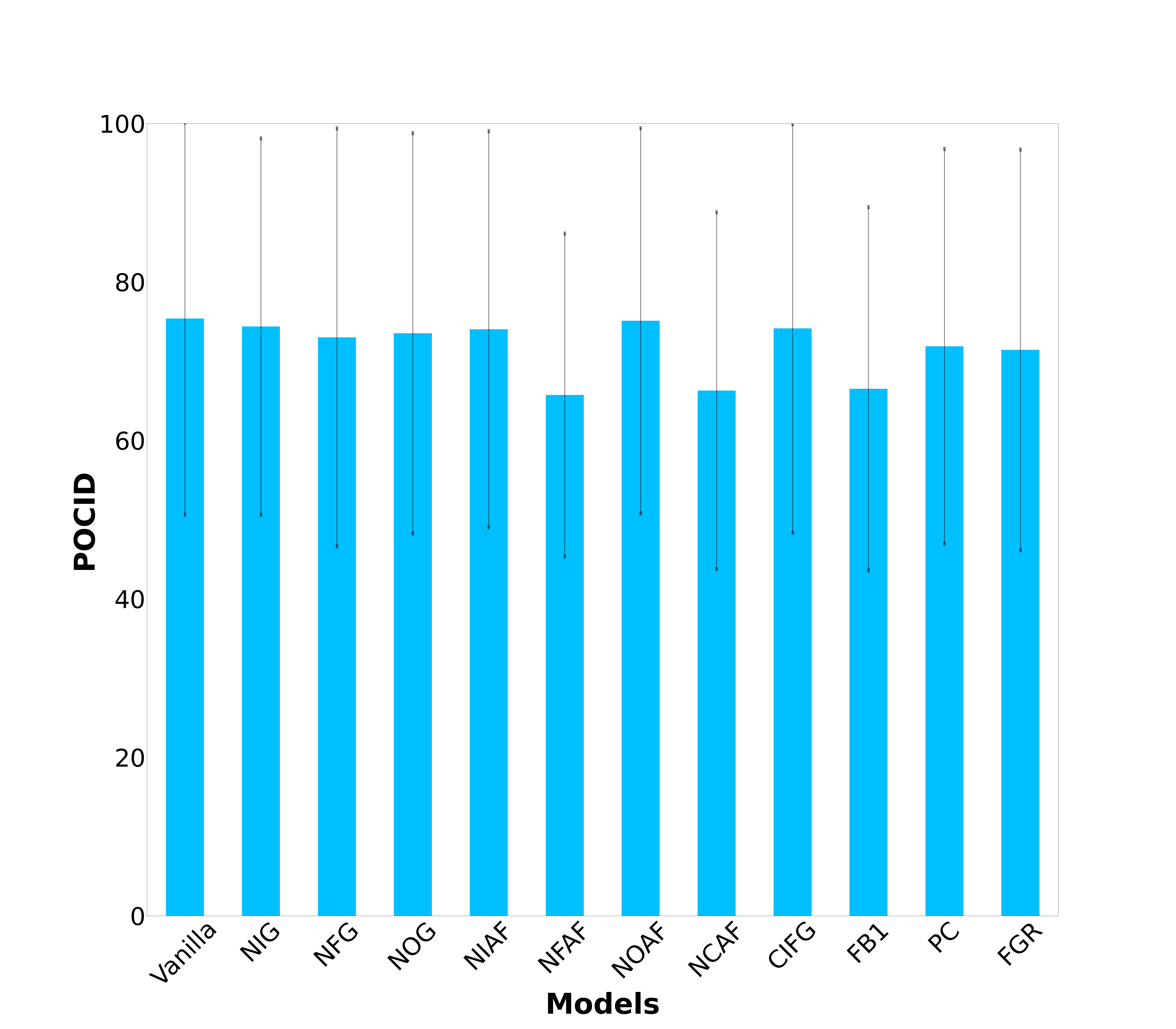}
         \caption{Mean and Std of POCID.}
         \label{F_D1_b}
     \end{subfigure}
    \caption{TU and POCID test results obtained by LSTM-Vanilla structures on deterministic time series.}
    \label{F_D1}
\end{figure}

\begin{figure}[H]
     \centering
     \begin{subfigure}[b]{0.5\textwidth}
         \centering
         \includegraphics[height=5.5cm,width=8cm]{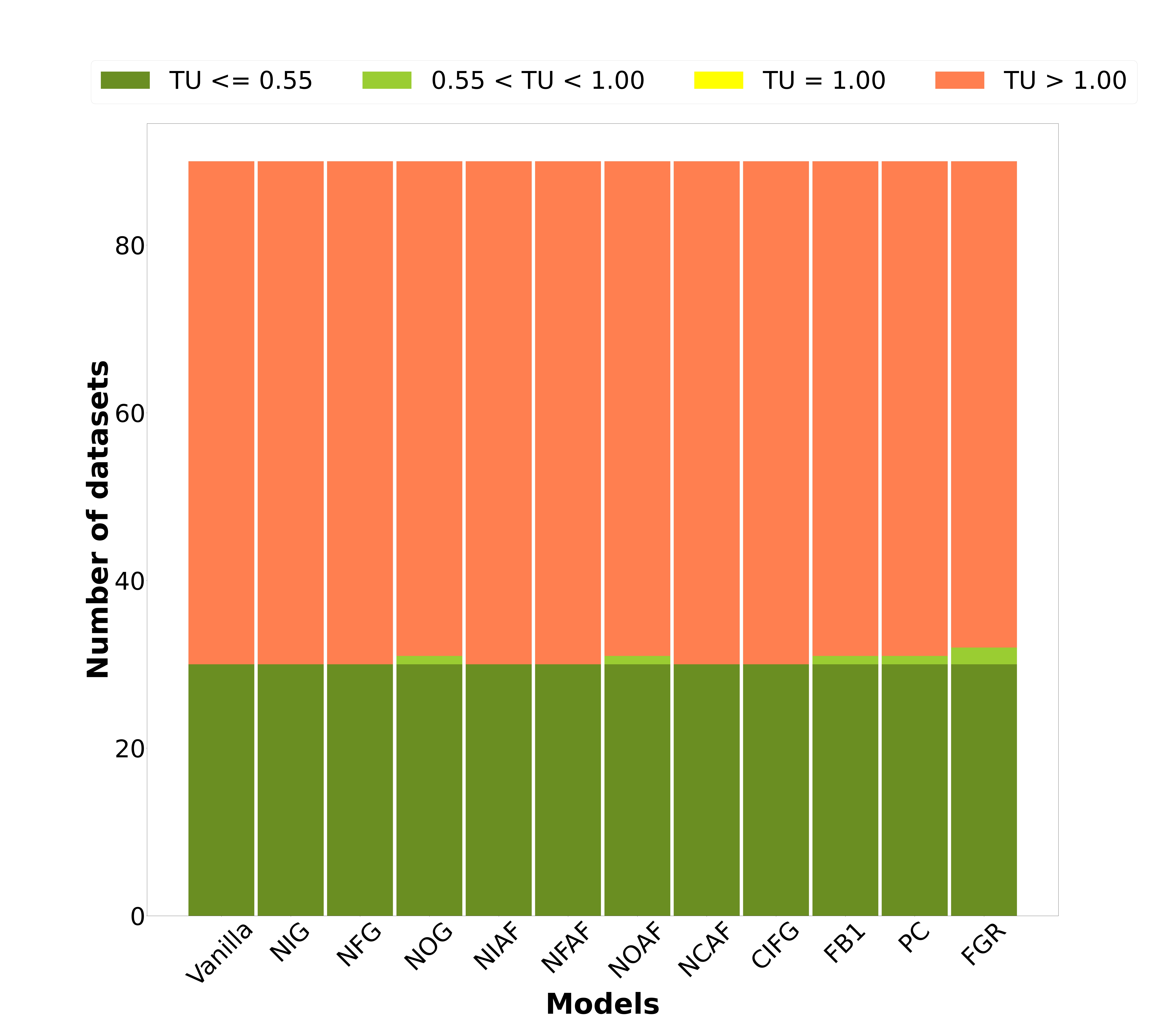}
         \caption{Four ranges of TU values.}
         \label{F_R1_a}
     \end{subfigure}
     \begin{subfigure}[b]{0.4\textwidth}
         \centering
         \includegraphics[height=5.3cm,width=8cm]{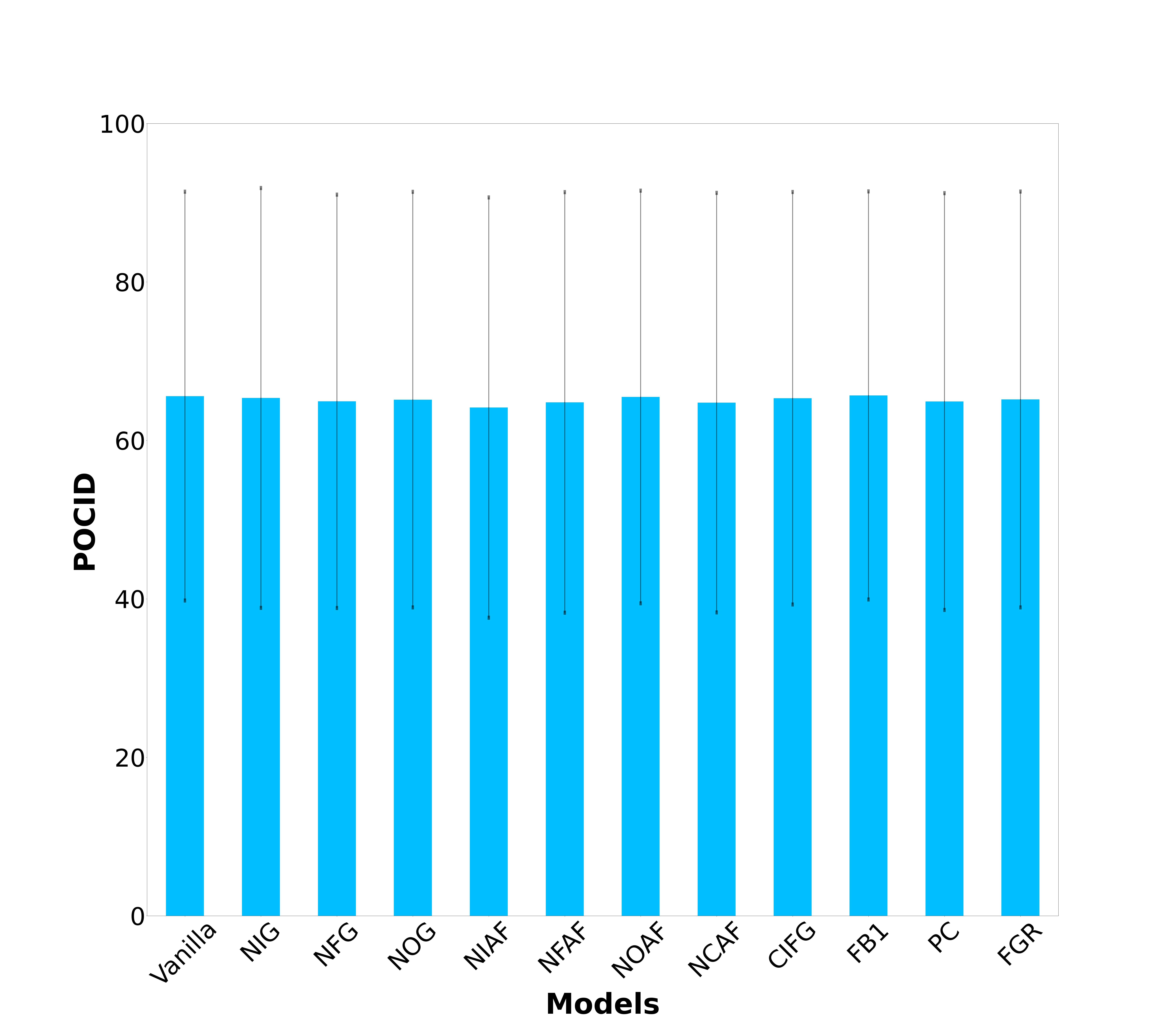}
         \caption{Mean and Std of POCID.}
         \label{F_R1_b}
     \end{subfigure}
    \caption{TU and POCID test results obtained by LSTM-Vanilla structures on random-walk time series}
    \label{F_R1}
\end{figure}

\begin{figure}[H]
     \centering
     \begin{subfigure}[b]{0.5\textwidth}
         \centering
         \includegraphics[height=5.5cm,width=8cm]{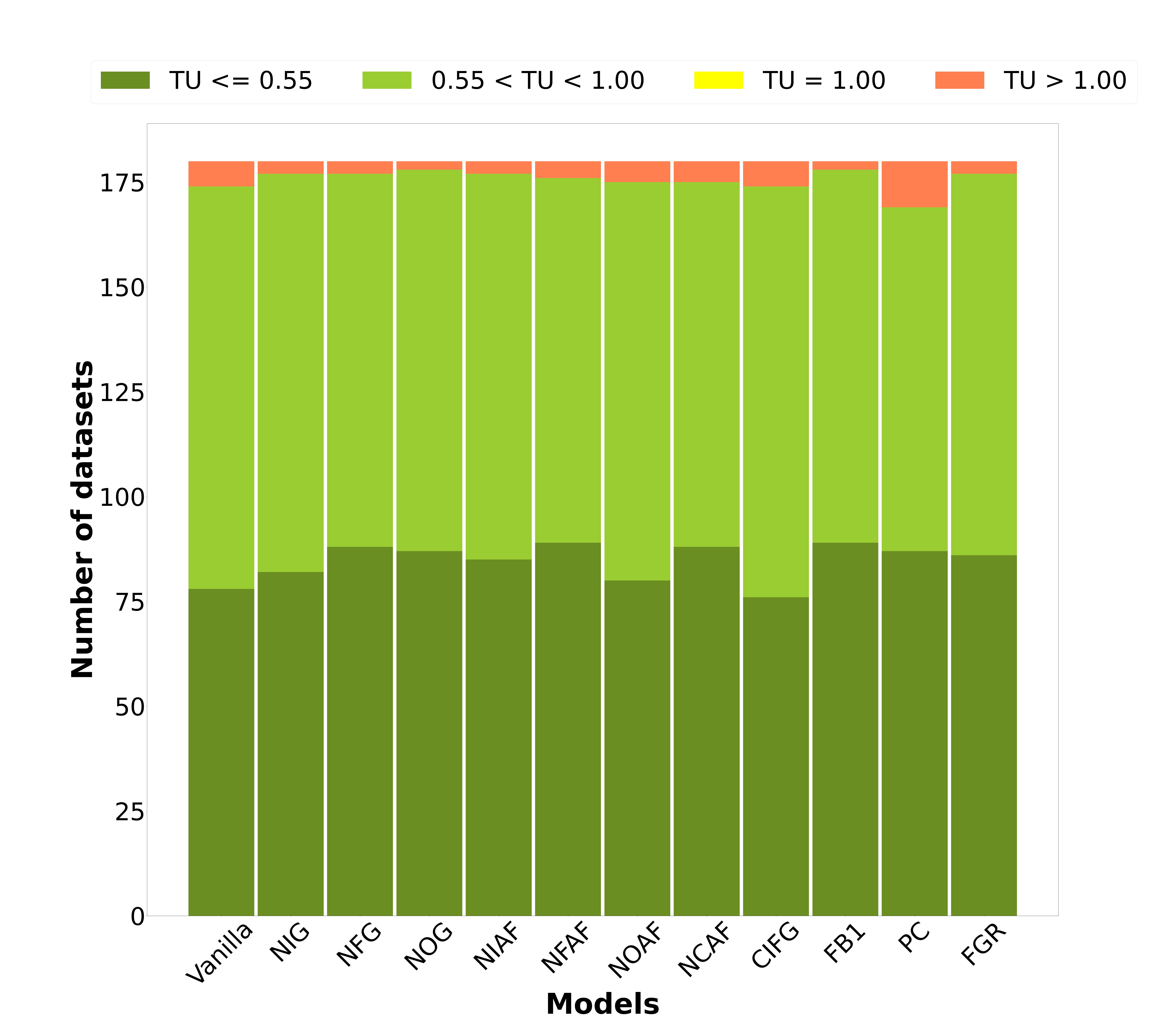}
         \caption{Four ranges of TU values.}
         \label{F_N1_a}
     \end{subfigure}
     \begin{subfigure}[b]{0.4\textwidth}
         \centering
         \includegraphics[height=5.3cm,width=8cm]{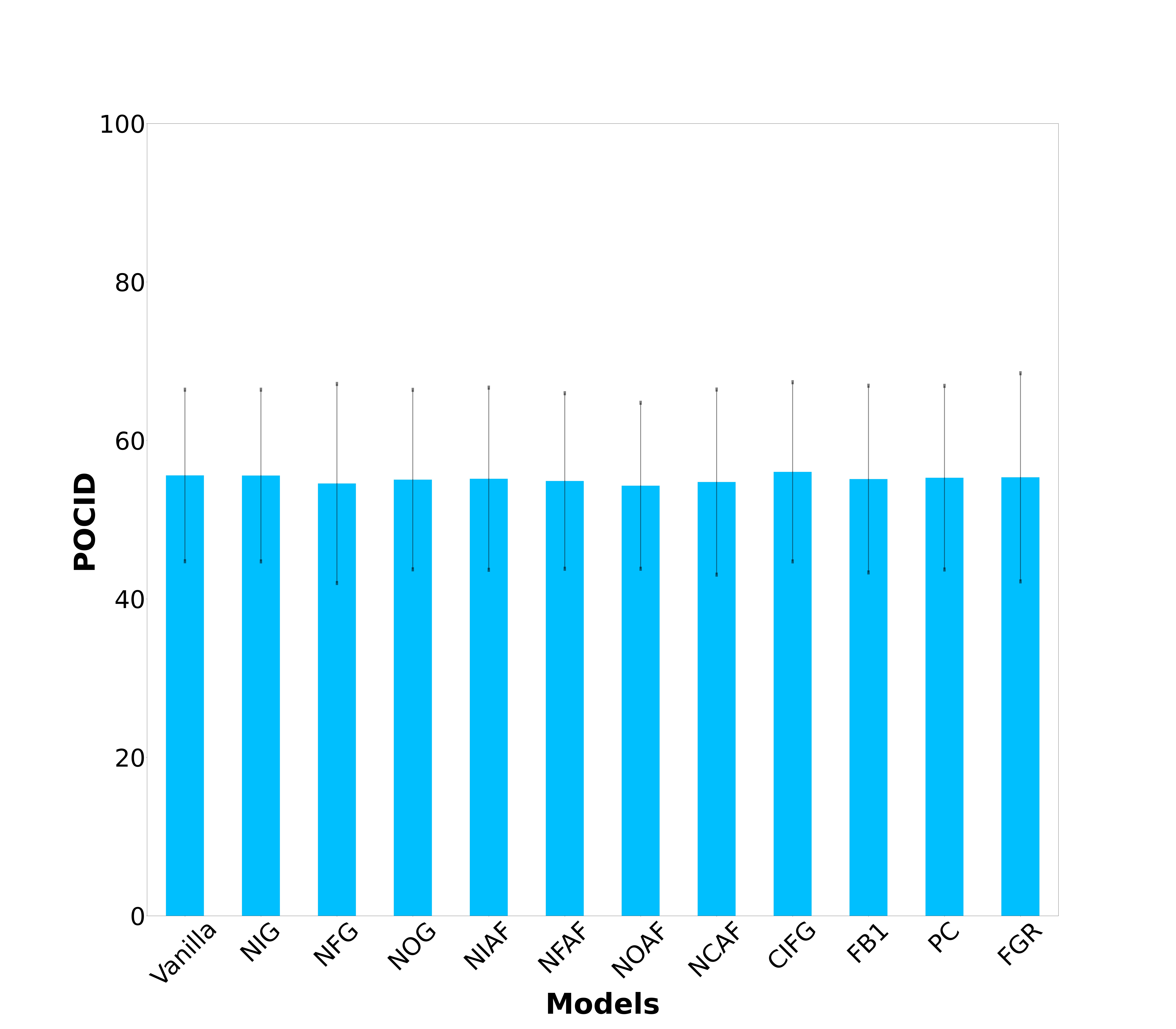}
         \caption{Mean and Std of POCID.}
         \label{F_N1_b}
     \end{subfigure}
    \caption{TU and POCID test results obtained by LSTM-Vanilla structures on nonlinear time series}
    \label{F_N1}
\end{figure}

\begin{figure}[H]
     \centering
     \begin{subfigure}[b]{0.5\textwidth}
         \centering
         \includegraphics[height=5.5cm,width=8cm]{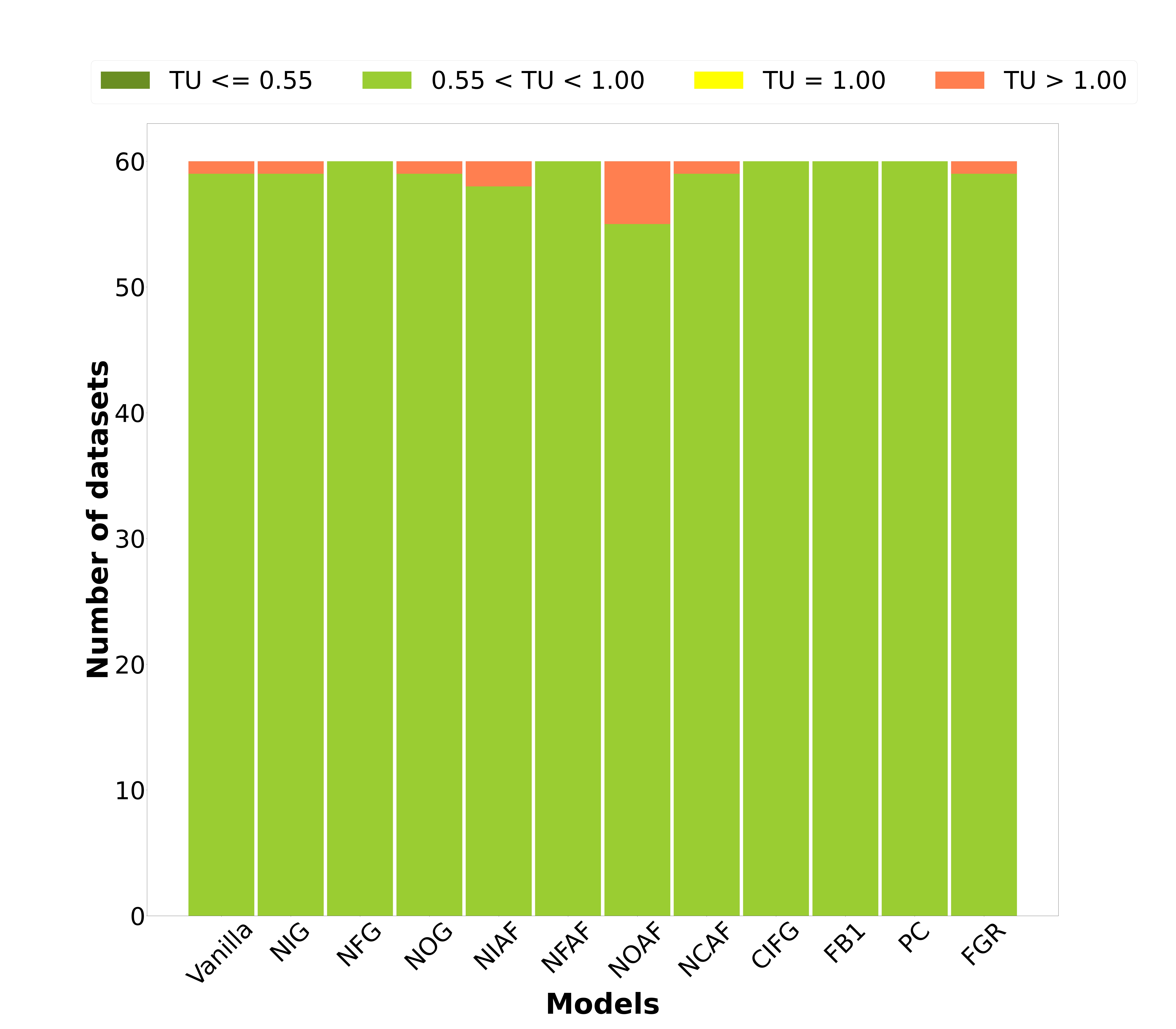}
         \caption{Four ranges of TU values.}
         \label{F_L1_a}
     \end{subfigure}
     \begin{subfigure}[b]{0.4\textwidth}
         \centering
         \includegraphics[height=5.3cm,width=8cm]{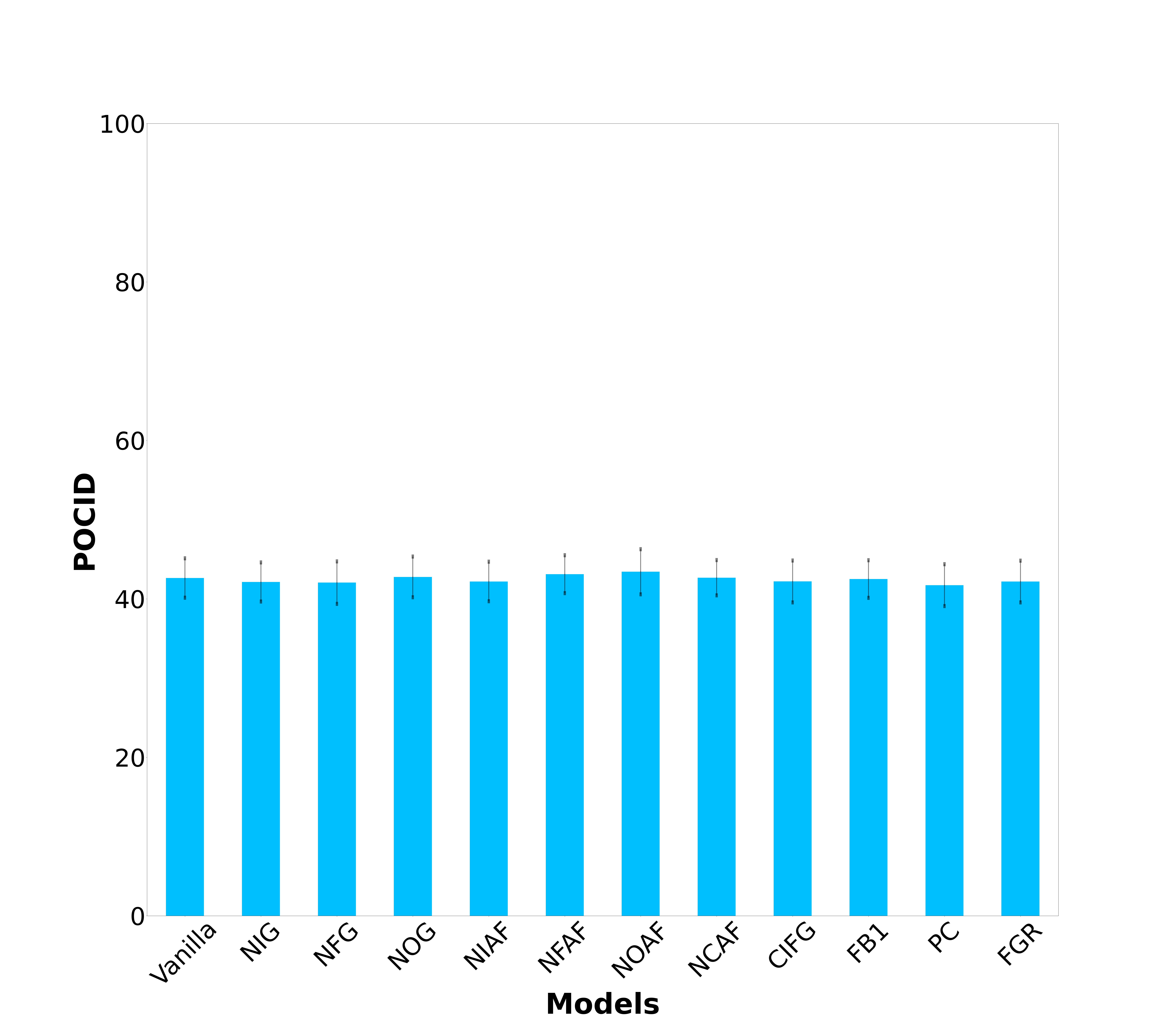}
         \caption{Mean and Std of POCID.}
         \label{F_L1_b}
     \end{subfigure}
    \caption{TU and POCID test results obtained by LSTM-Vanilla structures on long-memory time series}
    \label{F_L1}
\end{figure}

\begin{figure}[H]
     \centering
     \begin{subfigure}[b]{0.5\textwidth}
         \centering
         \includegraphics[height=5.5cm,width=8cm]{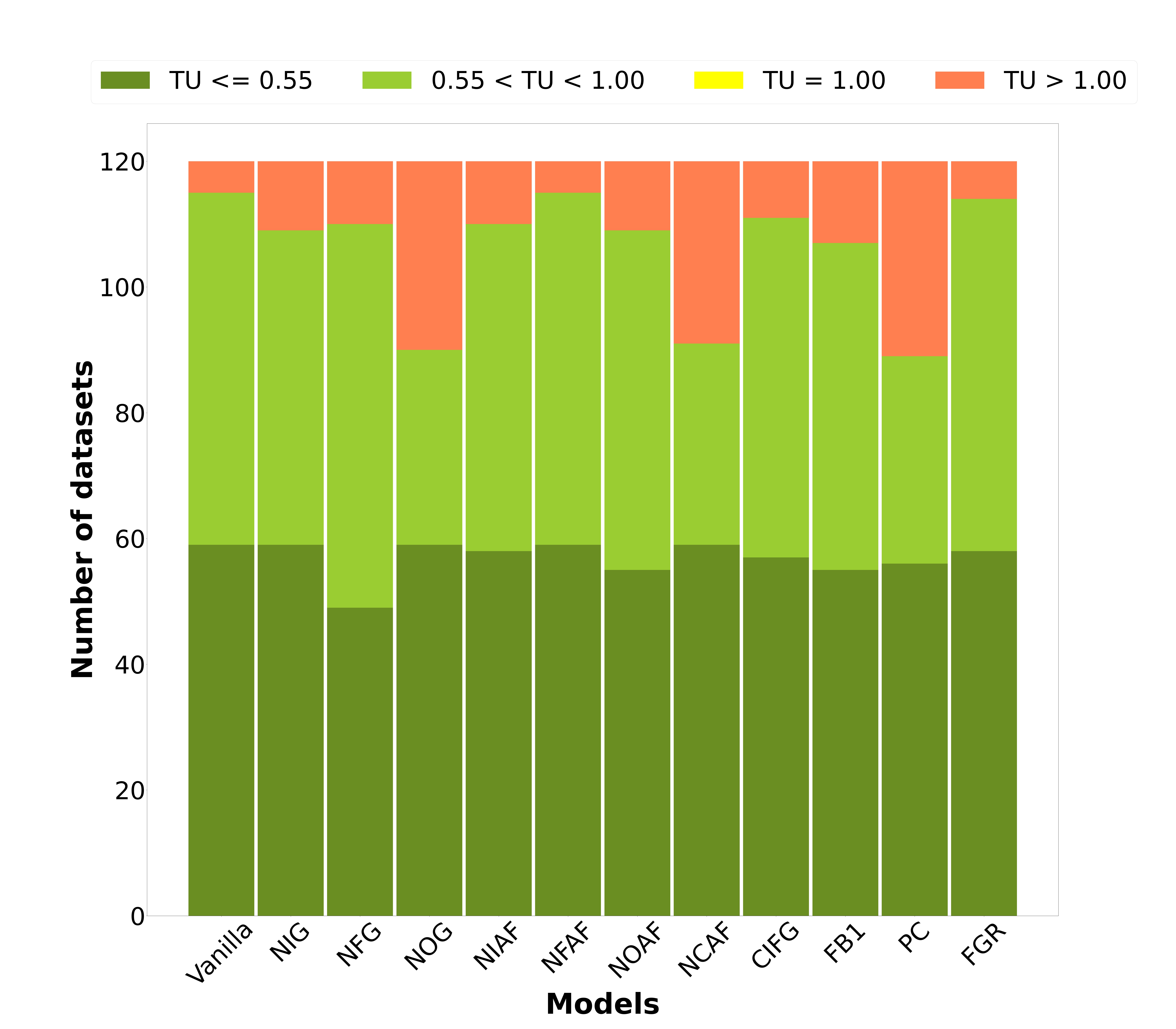}
         \caption{Four ranges of TU values.}
         \label{F_C1_a}
     \end{subfigure}
     \begin{subfigure}[b]{0.4\textwidth}
         \centering
         \includegraphics[height=5.3cm,width=8cm]{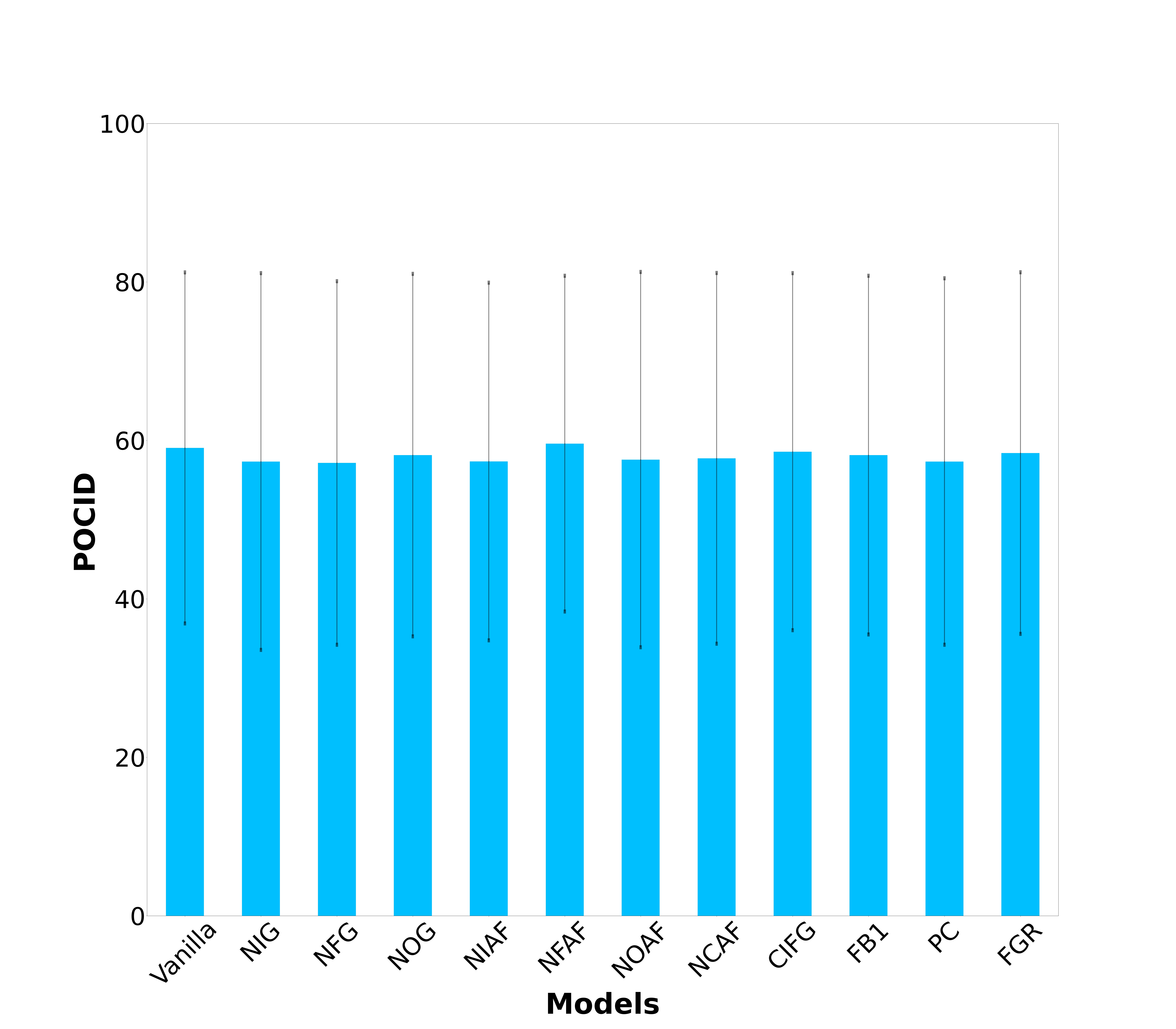}
         \caption{Mean and Std of POCID.}
         \label{F_C1_b}
     \end{subfigure}
        \caption{TU and POCID test results obtained by LSTM-Vanilla structures on chaotic time series}
        \label{F_C1}
\end{figure}

Figures \ref{F_D1} to \ref{F_C1} display the distribution of the TU and the POCID metrics over all the 12 LSTM variants for the five time series behaviors. 
In the deterministic behavior, all the models perform better than the naïve model for more than half of the used data (Figure  \ref{F_D1_a}). The POCID levels of NIAF, NCAF, and FB1 are lower than the remaining models (Figure  \ref{F_D1_b}).
In the random-walk behavior, all the models perform less than the naïve model for more than half of the used time series data (Figure \ref{F_R1_a}). They have approximately the same POCID amounts (Figure \ref{F_R1_b}). 
Within the nonlinear and the long-memory behaviors, the models performed less than the naïve model only for a small number of time series data (Figures \ref{F_N1_a} and \ref{F_L1_a}). The POCID levels of all the models are almost similar (Figures \ref{F_N1_b} and \ref{F_L1_b}).
With respect to the chaotic behavior, for more than half of the data, the models proved that they can be trusted to forecast such type of behavior (Figure \ref{F_C1_a}). Similarly, the POCID values of all the used models are almost equal (Figure \ref{F_C1_b}). 

The direction changes of the predicted values are more similar to the ones of the real values (almost oscillating around 60\%) with the deterministic, random-walk, nonlinear, and chaotic behaviors (Figures \ref{F_D1_b}, \ref{F_R1_b}, \ref{F_N1_b}, and \ref{F_C1_b}) than with the long-memory behavior where the values are only hovering around 40\% (Figure \ref{F_L1_b}).  

\begin{figure}[H]
     \centering
     \begin{subfigure}[b]{\textwidth}
         \centering
         \includegraphics[width=15cm]{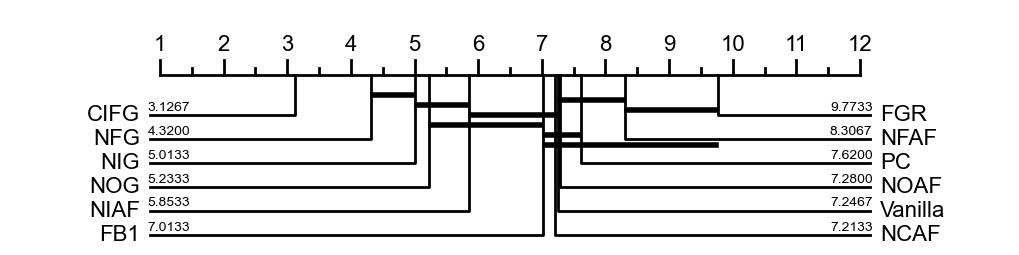}
         \caption{Deterministic behavior.}
         \label{CD1_a}
     \end{subfigure}
     \\
     \begin{subfigure}[b]{\textwidth}
         \centering
         \includegraphics[width=15cm]{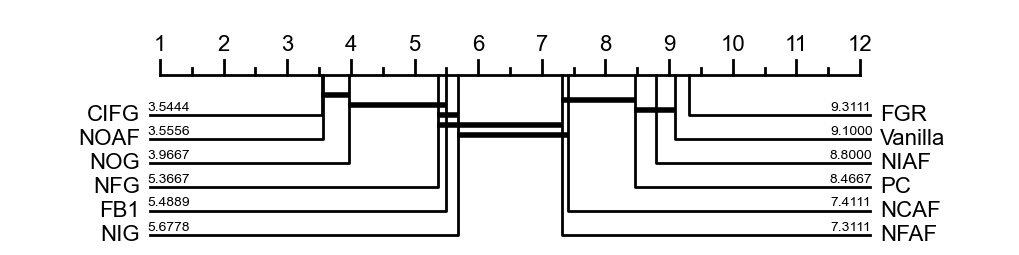}
         \caption{Random-walk behavior.}
         \label{CD1_b}
     \end{subfigure}
     \begin{subfigure}[b]{\textwidth}
         \centering
         \includegraphics[width=15cm]{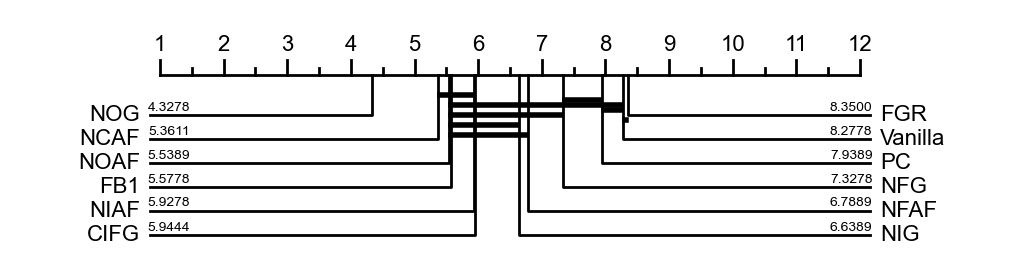}
         \caption{Nonlinear behavior.}
         \label{CD1_c}
     \end{subfigure}
     \begin{subfigure}[b]{\textwidth}
         \centering
         \includegraphics[width=15cm]{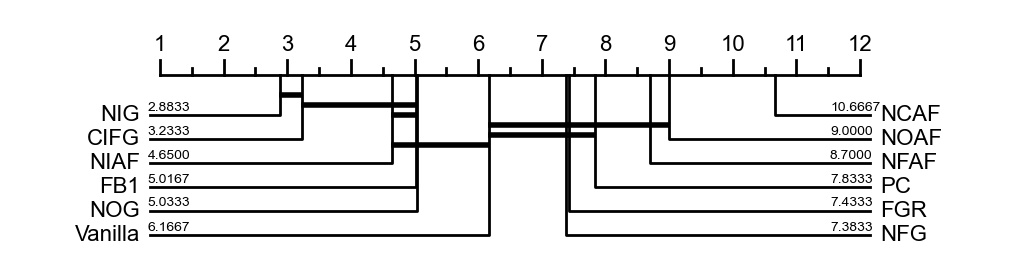}
         \caption{Long-memory behavior.}
         \label{CD1_d}
     \end{subfigure}
     \begin{subfigure}[b]{\textwidth}
         \centering
         \includegraphics[width=15cm]{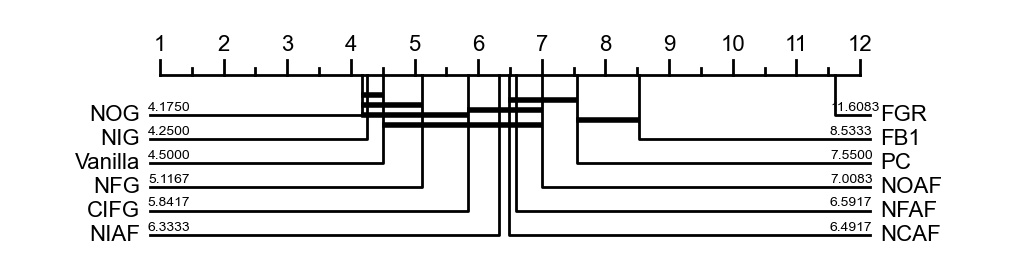}
         \caption{Chaotic behavior.}
         \label{CD1_e}
     \end{subfigure}
    \caption{CD diagrams for the MCIM values comparing the LSTM-Vanilla cell structures on the five time series behaviors.}
    \label{CD1}
\end{figure}

To strengthen our interpretation of the models' performances with respect to the five time series behaviors, Figure \ref{CD1} outlines the results of the statistical significance test between the used models for each data behavior. For the deterministic behavior, the CIFG structure is statistically different than all the other models (Figure \ref{CD1_a}). With the nonlinear behavior, the NOG variant is statistically better than the remaining models (Figure \ref{CD1_c}). With respect to the random-walk behavior, the CIFG and the NOAF have almost the same rank (Figure \ref{CD1_b}). The NIG followed by the CIFG achieved the best results with the long-memory behavior (Figure \ref{CD1_d}). With the chaotic behavior, the NOG followed by the NIG are better ranked than the Vanilla model recommended by the MCIM value (Table \ref{T_C1}), however, there is no statistical difference between them (Figure \ref{CD1_e}). For this type of behavior, the NOG variants are the most recommended by the used statistical test.   

\begin{table}[H]
\centering
\caption{Test results of LSTM-Vanilla variants for time series data with deterministic behavior.}
\label{T_D1}
\resizebox{\textwidth}{!}{%
\begin{tabular}{lllllllllllc}
\hline
\multicolumn{1}{l}{} & \multicolumn{1}{c}{\textbf{MAE}} & \multicolumn{1}{c}{\textbf{RMSE}} & \multicolumn{1}{c}{\textbf{TU}} & \multicolumn{1}{c}{\textbf{AIC}} & \multicolumn{1}{c}{\textbf{BIC}} & \multicolumn{1}{c}{\textbf{APC}} & \multicolumn{1}{c}{\textbf{HSP}} & \multicolumn{1}{c}{\textbf{SBIC}} & \multicolumn{1}{c}{\textbf{POCID}} & \multicolumn{1}{c}{\textbf{MCIM}} & \multicolumn{1}{c}{\textbf{Rank}} \\
\hline
\textbf{Vanilla} & 0,0388 & 0,0481 & 0,8116 & -3150,71 & -2004,37 & 0,0153 & 1,51E-05 & -3188,35 & 75,38 & 0,2977 & 5 \\
\textbf{NIG} & 0,0422 & 0,0520 & 1,0026 & -3400,56 & -2985,30 & 0,0041 & 5,96E-06 & -3420,66 & 74,38 & 0,2402 & 4 \\
\textbf{NFG} & 0,0387 & 0,0476 & 0,5384 & -3523,56 & -3046,95 & 0,0043 & 5,87E-06 & -3534,18 & 73,01 & 0,1474 & 2 \\
\textbf{NOG} & 0,0419 & 0,0522 & 0,7274 & -3450,51 & -3024,54 & 0,0042 & 6,31E-06 & -3470,50 & 73,52 & 0,2034 & 3 \\
\textbf{NIAF} & 0,0458 & 0,0561 & 1,2036 & -3328,43 & -2947,16 & 0,0047 & 6,86E-06 & -3369,75 & 74,02 & 0,3109 & 6 \\
\textbf{NFAF} & 0,0721 & 0,0856 & 1,3043 & -2831,01 & -2095,88 & 0,0195 & 2,51E-05 & -2858,21 & 65,73 & 0,7793 & 12 \\
\textbf{NOAF} & 0,0403 & 0,0498 & 1,0145 & -3213,70 & -2207,87 & 0,0131 & 1,35E-05 & -3283,54 & 75,10 & 0,3122 & 7 \\
\textbf{NCAF} & 0,0676 & 0,0785 & 0,9632 & -2872,99 & -1914,63 & 0,0170 & 2,23E-05 & -2922,57 & 66,29 & 0,6689 & 11 \\
\textbf{CIFG} & \textbf{0,0371} & \textbf{0,0459} & \textbf{0,6321} & \textbf{-3577,16} & \textbf{-3179,44} & \textbf{0,0032} & \textbf{4,76E-06} & \textbf{-3591,33} & \textbf{74,13} & \textbf{0,1210} & \textbf{1} \\
\textbf{FB1} & 0,0660 & 0,0767 & 0,7920 & -2961,69 & -2089,54 & 0,0229 & 2,71E-05 & -2981,35 & 66,52 & 0,6504 & 10 \\
\textbf{PC} & 0,0462 & 0,0573 & 1,2132 & -3126,17 & -2292,59 & 0,0092 & 1,10E-05 & -3162,30 & 71,88 & 0,4111 & 8 \\
\textbf{FGR} & 0,0445 & 0,0549 & 1,0311 & -1900,89 & 1991,09 & -0,0142 & -8,3E-06 & -2152,23 & 71,43 & 0,4990 & 9 \\ \hline
\end{tabular}%
}
\end{table}

\begin{table}[H]
\centering
\caption{Test results of LSTM-Vanilla variants for time series data with random-walk behavior.}
\label{T_R1}
\resizebox{\textwidth}{!}{%
\begin{tabular}{lllllllllllc}
\hline
\multicolumn{1}{l}{} & \multicolumn{1}{c}{\textbf{MAE}} & \multicolumn{1}{c}{\textbf{RMSE}} & \multicolumn{1}{c}{\textbf{TU}} & \multicolumn{1}{c}{\textbf{AIC}} & \multicolumn{1}{c}{\textbf{BIC}} & \multicolumn{1}{c}{\textbf{APC}} & \multicolumn{1}{c}{\textbf{HSP}} & \multicolumn{1}{c}{\textbf{SBIC}} & \multicolumn{1}{c}{\textbf{POCID}} & \multicolumn{1}{c}{\textbf{MCIM}} & \multicolumn{1}{c}{\textbf{Rank}} \\ \hline
\textbf{Vanilla} & 0,0287 & 0,0347 & 35,4234 & -3379,26 & -1698,17 & 0,0088 & 8,84E-06 & -3619,74 & 65,59 & 0,4714 & 11 \\
\textbf{NIG} & 0,0289 & 0,0353 & 24,8229 & -3873,56 & -3219,36 & 0,0044 & 5,05E-06 & -3946,55 & 65,36 & 0,3121 & 5 \\
\textbf{NFG} & 0,0291 & 0,0347 & 18,9766 & -3890,46 & -3194,67 & 0,0034 & 4,15E-06 & -4028,81 & 64,94 & 0,3151 & 6 \\
\textbf{NOG} & 0,0273 & 0,0329 & 20,1432 & -4110,57 & -3687,67 & 0,0026 & 3,71E-06 & -4161,53 & 65,15 & 0,2176 & 4 \\
\textbf{NIAF} & 0,0369 & 0,0438 & 352,6018 & -3399,64 & -2086,53 & 0,0178 & 1,97E-05 & -3610,75 & 64,15 & 0,8970 & 12 \\
\textbf{NFAF} & 0,0279 & 0,0336 & 29,0595 & -3684,21 & -2632,86 & 0,0033 & 3,96E-06 & -3821,16 & 64,80 & 0,3747 & 8 \\
\textbf{NOAF} & 0,0252 & 0,0310 & 13,0473 & -4227,70 & -3977,46 & 0,0015 & 2,34E-06 & -4251,86 & 65,49 & 0,1193 & 2 \\
\textbf{NCAF} & 0,0271 & 0,0326 & 22,9176 & -3659,89 & -2394,30 & 0,0044 & 4,81E-06 & -3816,47 & 64,76 & 0,3833 & 9 \\
\textbf{CIFG} & \textbf{0,0234} & \textbf{0,0285} & \textbf{15,1138} & \textbf{-4290,79} & \textbf{-3993,92} & \textbf{0,0013} & \textbf{1,94E-06} & \textbf{-4324,71} & \textbf{65,33} & \textbf{0,0830} & \textbf{1} \\
\textbf{FB1} & 0,0259 & 0,0314 & 47,4964 & -3954,61 & -3213,19 & 0,0025 & 3,30E-06 & -4044,93 & 65,68 & 0,2090 & 3 \\
\textbf{PC} & 0,0254 & 0,0309 & 28,5131 & -3573,45 & -1977,54 & 0,0086 & 8,24E-06 & -3747,50 & 64,91 & 0,4238 & 10 \\
\textbf{FGR} & 0,0213 & 0,0259 & 13,2219 & -3094,99 & 104,24 & -0,0011 & -2,00E-07 & -3497,19 & 65,17 & 0,3705 & 7 \\ \hline
\end{tabular}%
}
\end{table}

\begin{table}[H]
\centering
\caption{Test results of LSTM-Vanilla variants for time series data with nonlinear behavior.}
\label{T_N1}
\resizebox{\textwidth}{!}{%
\begin{tabular}{lllllllllllc}
\hline
\multicolumn{1}{l}{} & \multicolumn{1}{c}{\textbf{MAE}} & \multicolumn{1}{c}{\textbf{RMSE}} & \multicolumn{1}{c}{\textbf{TU}} & \multicolumn{1}{c}{\textbf{AIC}} & \multicolumn{1}{c}{\textbf{BIC}} & \multicolumn{1}{c}{\textbf{APC}} & \multicolumn{1}{c}{\textbf{HSP}} & \multicolumn{1}{c}{\textbf{SBIC}} & \multicolumn{1}{c}{\textbf{POCID}} & \multicolumn{1}{c}{\textbf{MCIM}} & \multicolumn{1}{c}{\textbf{Rank}} \\ \hline
\textbf{Vanilla} & 0,1147 & 0,1437 & 0,5826 & -1820,62 & -626,42 & 0,0704 & 7,81E-05 & -1852,35 & 55,58 & 0,5189 & 11 \\
\textbf{NIG} & 0,1142 & 0,1430 & 0,5787 & -1977,20 & -1110,66 & 0,0517 & 6,19E-05 & -2003,20 & 55,57 & 0,3706 & 4 \\
\textbf{NFG} & 0,1138 & 0,1424 & 0,5739 & -1897,99 & -845,29 & 0,0505 & 6,06E-05 & -1921,56 & 54,56 & 0,4039 & 6 \\
\textbf{NOG} & \textbf{0,1135} & \textbf{0,1420} & \textbf{0,5677} & \textbf{-2148,56} & \textbf{-1645,13} & \textbf{0,0380} & \textbf{5,01E-05} & \textbf{-2163,75} & \textbf{55,05} & \textbf{0,2187} & \textbf{1} \\
\textbf{NIAF} & 0,1140 & 0,1428 & 0,5756 & -1977,74 & -1104,28 & 0,0586 & 6,76E-05 & -2001,51 & 55,16 & 0,3853 & 5 \\
\textbf{NFAF} & 0,1141 & 0,1428 & 0,5766 & -1952,86 & -1026,12 & 0,0553 & 6,49E-05 & -1981,19 & 54,88 & 0,4139 & 7 \\
\textbf{NOAF} & 0,1159 & 0,1458 & 0,5891 & -2006,59 & -1243,99 & 0,0593 & 6,94E-05 & -2022,09 & 54,28 & 0,6649 & 12 \\
\textbf{NCAF} & 0,1140 & 0,1430 & 0,5812 & -2092,99 & -1492,28 & 0,0392 & 5,14E-05 & -2108,14 & 54,74 & 0,3658 & 3 \\
\textbf{CIFG} & 0,1155 & 0,1448 & 0,5943 & -2045,55 & -1368,08 & 0,0425 & 5,47E-05 & -2059,41 & 56,04 & 0,4845 & 9 \\
\textbf{FB1} & 0,1136 & 0,1422 & 0,57 & -1978,25 & -1101,31 & 0,0415 & 5,30E-05 & -1991,78 & 55,12 & 0,2922 & 2 \\
\textbf{PC} & 0,1141 & 0,1430 & 0,5834 & -1829,74 & -648,40 & 0,0707 & 7,80E-05 & -1858,75 & 55,29 & 0,4878 & 10 \\
\textbf{FGR} & 0,1144 & 0,1430 & 0,5747 & -1034,40 & 1910,12 & -0,0410 & -1,50E-05 & -1110,85 & 55,34 & 0,4773 & 8 \\ \hline
\end{tabular}%
}
\end{table}

\begin{table}[H]
\centering
\caption{Test results of LSTM-Vanilla variants for time series data with long-memory behavior.}
\label{T_L1}
\resizebox{\textwidth}{!}{%
\begin{tabular}{lllllllllllc}
\hline
\multicolumn{1}{l}{} & \multicolumn{1}{c}{\textbf{MAE}} & \multicolumn{1}{c}{\textbf{RMSE}} & \multicolumn{1}{c}{\textbf{TU}} & \multicolumn{1}{c}{\textbf{AIC}} & \multicolumn{1}{c}{\textbf{BIC}} & \multicolumn{1}{c}{\textbf{APC}} & \multicolumn{1}{c}{\textbf{HSP}} & \multicolumn{1}{c}{\textbf{SBIC}} & \multicolumn{1}{c}{\textbf{POCID}} & \multicolumn{1}{c}{\textbf{MCIM}} & \multicolumn{1}{c}{\textbf{Rank}} \\ \hline
\textbf{Vanilla} & 0,0834 & 0,1046 & 0,8042 & -2311,82 & -1517,30 & 0,0312 & 3,66E-05 & -2311,27 & 42,63 & 0,2158 & 8 \\
\textbf{NIG} & \textbf{0,0819} & \textbf{0,1026} & \textbf{0,7767} & \textbf{-2563,29} & \textbf{-2349,46} & \textbf{0,0128} & \textbf{2,04E-05} & \textbf{-2559,39} & \textbf{42,13} & \textbf{0,0851} & \textbf{1} \\
\textbf{NFG} & 0,0825 & 0,1033 & 0,7882 & -2299,99 & -1378,39 & 0,0251 & 3,04E-05 & -2304,08 & 42,05 & 0,2593 & 10 \\
\textbf{NOG} & 0,0826 & 0,1035 & 0,7880 & -2417,51 & -1981,24 & 0,0168 & 2,42E-05 & -2415,43 & 42,77 & 0,1298 & 4 \\
\textbf{NIAF} & 0,0834 & 0,1075 & 0,9296 & -2470,59 & -2174,63 & 0,0158 & 2,47E-05 & -2467,95 & 42,20 & 0,1412 & 5 \\
\textbf{NFAF} & 0,0831 & 0,1042 & 0,8014 & -2250,91 & -1382,01 & 0,0228 & 2,92E-05 & -2252,96 & 43,11 & 0,2147 & 7 \\
\textbf{NOAF} & 0,1002 & 0,2860 & 139,2791 & -2059,28 & -1192,65 & 3,2133 & 4,24E-03 & -2061,37 & 43,42 & 0,8308 & 12 \\
\textbf{NCAF} & 0,0832 & 0,1045 & 0,8031 & -2016,45 & -503,06 & 0,0866 & 8,33E-05 & -2023,55 & 42,67 & 0,3962 & 11 \\
\textbf{CIFG} & 0,0819 & 0,1026 & 0,7775 & -2561,35 & -2259,12 & 0,0133 & 2,06E-05 & -2557,83 & 42,20 & 0,0868 & 2 \\
\textbf{FB1} & 0,0819 & 0,1026 & 0,7759 & -2466,48 & -2043,20 & 0,0152 & 2,24E-05 & -2463,62 & 42,51 & 0,1179 & 3 \\
\textbf{PC} & 0,0831 & 0,1044 & 0,8033 & -2355,37 & -1579,13 & 0,0256 & 3,11E-05 & -2354,66 & 41,73 & 0,2516 & 9 \\
\textbf{FGR} & 0,0823 & 0,1032 & 0,7862 & -2380,71 & -1746,97 & 0,0182 & 2,49E-05 & -2381,48 & 42,19 & 0,1947 & 6 \\ \hline
\end{tabular}%
}
\end{table}

\begin{table}[H]
\centering
\caption{Test results of LSTM-Vanilla variants for time series data with chaotic behavior.}
\label{T_C1}
\resizebox{\textwidth}{!}{%
\begin{tabular}{lllllllllllc}
\hline
\multicolumn{1}{l}{} & \multicolumn{1}{c}{\textbf{MAE}} & \multicolumn{1}{c}{\textbf{RMSE}} & \multicolumn{1}{c}{\textbf{TU}} & \multicolumn{1}{c}{\textbf{AIC}} & \multicolumn{1}{c}{\textbf{BIC}} & \multicolumn{1}{c}{\textbf{APC}} & \multicolumn{1}{c}{\textbf{HSP}} & \multicolumn{1}{c}{\textbf{SBIC}} & \multicolumn{1}{c}{\textbf{POCID}} & \multicolumn{1}{c}{\textbf{MCIM}} & \multicolumn{1}{c}{\textbf{Rank}} \\ \hline
\textbf{Vanilla} & \textbf{0,0532} & \textbf{0,0671} & \textbf{0,4772} & \textbf{-3475,04} & \textbf{-2677,35} & \textbf{0,0506} & \textbf{4,94E-05} & \textbf{-3486,80} & \textbf{59,06} & \textbf{0,2131} & \textbf{1} \\
\textbf{NIG} & 0,0530 & 0,0669 & 0,5168 & -3576,95 & -3005,20 & 0,0089 & 1,38E-05 & -3598,77 & 57,33 & 0,2299 & 2 \\
\textbf{NFG} & 0,0536 & 0,0677 & 0,5595 & -3505,32 & -2986,05 & 0,0139 & 1,81E-05 & -3519,26 & 57,15 & 0,3227 & 7 \\
\textbf{NOG} & 0,0536 & 0,0676 & 0,5264 & -3580,20 & -3025,62 & 0,0178 & 2,15E-05 & -3588,31 & 58,15 & 0,2355 & 3 \\
\textbf{NIAF} & 0,0622 & 0,0777 & 0,5529 & -3497,35 & -2960,94 & 0,0194 & 2,49E-05 & -3510,89 & 57,34 & 0,5260 & 10 \\
\textbf{NFAF} & 0,0543 & 0,0695 & 0,5207 & -3337,37 & -2202,66 & 0,0622 & 6,01E-05 & -3365,92 & 59,59 & 0,3263 & 8 \\
\textbf{NOAF} & 0,0529 & 0,0668 & 0,5287 & -3300,57 & -2173,95 & 0,0159 & 1,96E-05 & -3363,36 & 57,58 & 0,2937 & 4 \\
\textbf{NCAF} & 0,0533 & 0,0673 & 0,5460 & -3370,83 & -2296,95 & 0,0168 & 2,05E-05 & -3416,97 & 57,74 & 0,3071 & 5 \\
\textbf{CIFG} & 0,0561 & 0,0709 & 0,5346 & -3446,18 & -2694,20 & 0,0205 & 2,43E-05 & -3466,81 & 58,58 & 0,3183 & 6 \\
\textbf{FB1} & 0,0602 & 0,0752 & 0,5673 & -3166,73 & -1891,64 & 0,0280 & 3,16E-05 & -3216,24 & 58,15 & 0,5341 & 11 \\
\textbf{PC} & 0,0602 & 0,0752 & 0,5740 & -3274,88 & -2191,16 & 0,0808 & 7,64E-05 & -3327,35 & 57,33 & 0,6630 & 12 \\
\textbf{FGR} & 0,0531 & 0,0669 & 0,4978 & -1807,35 & 2662,05 & -0,0312 & -2E-05 & -1992,26 & 58,40 & 0,4144 & 9 \\ \hline
\end{tabular}%
}
\end{table}

\subsection{\bf Experiment 2: Performance analysis of different RNN cell structures in forecasting time series behaviors.} \label{S6-2}
The performance of each RNN cell structure (JORDAN, ELMAN, MRNN, SCRN, IRNN, LSTM-Vanilla, GRU, MGU, MUT1, MUT2, MUT3, and 9 SLIM variants) for predicting the deterministic behavior, random-walk behavior, nonlinear behavior, long-memory behavior, and chaotic behavior is provided in this section.

Tables \ref{T_D2} to \ref{T_C2} highlight the outcomes of the statistical metrics for each RNN model run on the five types of time series behaviors. Based on our MCIM metric, the model MGU-SLIM3 outperforms the other models with respect to the deterministic (Table \ref{T_D2}) and the nonlinear (Table \ref{T_N2}) behaviors. MGU-SLIM2 achieves better results than the remaining models when applied to data with random-walk (Table \ref{T_R2}) and long-memory (Table \ref{T_L2}) behaviors. Eventually, LSTM-SLIM3 exhibits higher abilities to forecast time series data with chaotic behavior (Table \ref{T_C2}).

Figures \ref{F_D2} to \ref{F_C2} show the distribution of the TU and the POCID metrics over all the 20 RNN structures for the five time series behaviors. 
In the deterministic behavior, all the models perform better than the naïve model for more than half of the used data except the JORDAN model (Figure  \ref{F_D2_a}). The models display different POCID levels but generally oscillate around 60\% (Figure  \ref{F_D2_b}).
In the random-walk behavior, all the models perform less than the naïve model for more than half of the used time series data (Figure \ref{F_R2_a}). The basic RNN variants (ELMAN, JORDAN, MRNN, and IRNN) have smaller POCID values (less than 60\%) compared to the other models (Figure \ref{F_R2_b}). 
With regard to the nonlinear and the long-memory behaviors, the models are performing widely better than the naïve model (Figures \ref{F_N2_a} and \ref{F_L2_a}). These models proved that they can be trusted to forecast nonlinear data for more than half of the time series. There are small variations in the POCID levels of these RNN models (Figures \ref{F_N2_b} and \ref{F_L2_b}).
For the chaotic behavior, the RNN models are better than the naïve model for more than half of the time series data (Figure \ref{F_C2_a}). Similarly, there are small variations in the POCID levels of the RNN models (Figure \ref{F_C2_b}). 

The direction changes of the predicted values are less similar to the ones of the real values for the long-memory behavior (oscillating around 40\%) (Figure \ref{F_L2_b}), however, they are more similar to the remaining behaviors with a percentage more than 50\% (Figures \ref{F_D2_b}, \ref{F_R2_b}, \ref{F_N2_b}, and \ref{F_C2_b}).

\begin{figure}[H]
     \centering
     \begin{subfigure}[b]{\textwidth}
         \centering
         \includegraphics[width=15cm]{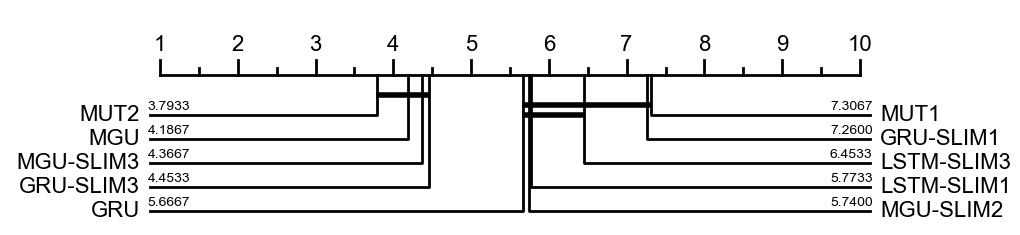}
         \caption{Deterministic behavior.}
         \label{CD2_a}
     \end{subfigure}
     \\
     \begin{subfigure}[b]{\textwidth}
         \centering
         \includegraphics[width=15cm]{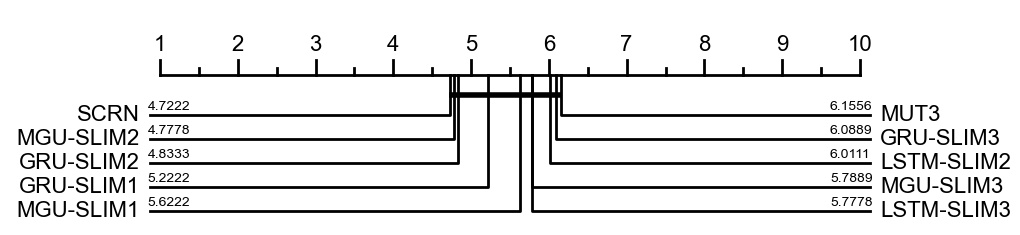}
         \caption{Random-walk behavior.}
         \label{CD2_b}
     \end{subfigure}
     \begin{subfigure}[b]{\textwidth}
         \centering
         \includegraphics[width=15cm]{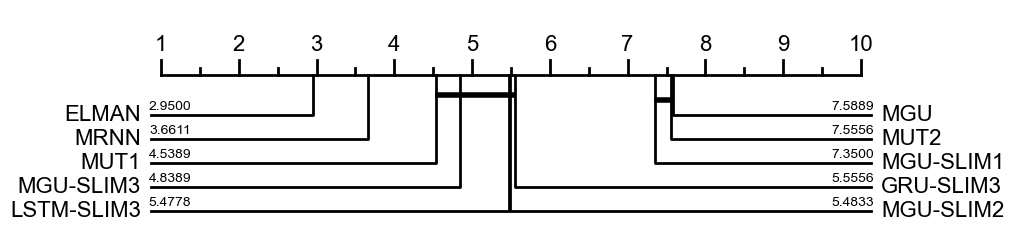}
         \caption{Nonlinear behavior.}
         \label{CD2_c}
     \end{subfigure}
     \begin{subfigure}[b]{\textwidth}
         \centering
         \includegraphics[width=15cm]{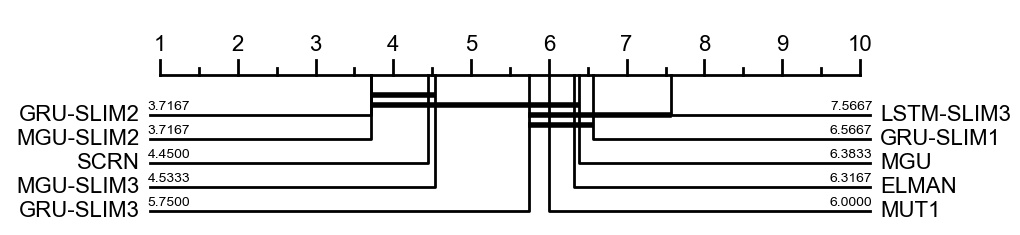}
         \caption{Long-memory behavior.}
         \label{CD2_d}
     \end{subfigure}
     \begin{subfigure}[b]{\textwidth}
         \centering
         \includegraphics[width=15cm]{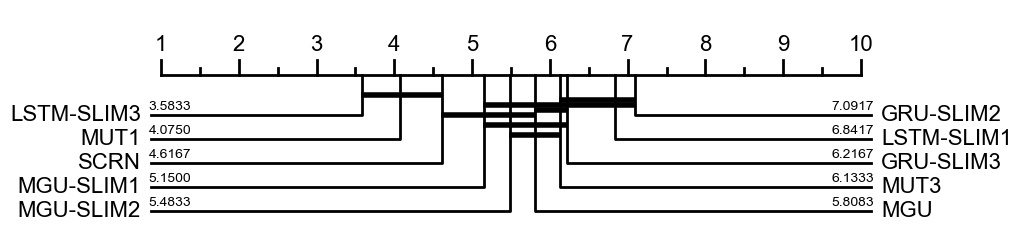}
         \caption{Chaotic behavior.}
         \label{CD2_e}
     \end{subfigure}
        \caption{CD diagrams for the MCIM values comparing the top 10 RNN cell structures on the five time series behaviors.}
        \label{CD2}
\end{figure}

Figure \ref{CD2} displays the statistical significance test results for top the 10 ranked models with respect to the five behaviors. With the deterministic behavior, the MUT2 followed by MGU, MGU-SLIM3, and GRU-SLIM3 are statistically similar such that the MUT2 variant is better ranked than the other models (Figure \ref{CD2_a}).
With the random-walk behavior, all the 10 models are statistically similar such that SCRN followed by MGU-SLIM2 have the higher ranks (Figure \ref{CD2_b}).  
For the nonlinear behavior, ELMAN is the top one ranked model that is significantly different than the remaining models (Figure \ref{CD2_c}. This outcome appears to be different than the one presented in Table \ref{T_N2} where this model has the rank of five. 
For the long-memory behavior, MGU-SLIM2 and GRU-SLIM2 have similar ranks (Figure \ref{CD2_d}, while LSTM-SLIM3 is the best-ranked model for the chaotic behavior (Figure \ref{CD2_e}.

\begin{table}[H]
\centering
\caption{Test results of RNN variants for time series data with deterministic behavior.}
\label{T_D2}
\resizebox{\textwidth}{!}{%
\begin{tabular}{lllllllllllc}
\hline
\multicolumn{1}{l}{} & \multicolumn{1}{c}{\textbf{MAE}} & \multicolumn{1}{c}{\textbf{RMSE}} & \multicolumn{1}{c}{\textbf{TU}} & \multicolumn{1}{c}{\textbf{AIC}} & \multicolumn{1}{c}{\textbf{BIC}} & \multicolumn{1}{c}{\textbf{APC}} & \multicolumn{1}{c}{\textbf{HSP}} & \multicolumn{1}{c}{\textbf{SBIC}} & \multicolumn{1}{c}{\textbf{POCID}} & \multicolumn{1}{c}{\textbf{MCIM}} & \multicolumn{1}{c}{\textbf{Rank}} \\ \hline
\textbf{ELMAN} & 0,0777 & 0,0929 & 2,0477 & -2983,50 & -2935,19 & 0,0115 & 1,90E-05 & -2979,17 & 66,07 & 0,7339 & 18 \\
\textbf{JORDAN} & 0,0794 & 0,0922 & 1,0851 & -3089,27 & -3021,59 & 0,0134 & 2,14E-05 & -3085,68 & 61,10 & 0,6794 & 17 \\
\textbf{MRNN} & 0,0594 & 0,0717 & 1,0088 & -3262,12 & -3209,41 & 0,0066 & 1,09E-05 & -3258,36 & 65,70 & 0,3971 & 14 \\
\textbf{IRNN} & 0,0880 & 0,1015 & 0,9632 & -2951,90 & -2734,79 & 0,0189 & 2,80E-05 & -2951,70 & 57,48 & 0,8617 & 20 \\
\textbf{SCRN} & 0,0699 & 0,0827 & 0,7725 & -3226,38 & -3061,21 & 0,0125 & 1,93E-05 & -3225,90 & 61,95 & 0,5492 & 16 \\
\textbf{MUT1} & 0,0403 & 0,0499 & 0,9056 & -3403,41 & -2826,57 & 0,0045 & 6,20E-06 & -3436,33 & 71,53 & 0,2116 & 9 \\
\textbf{MUT2} & 0,0376 & 0,0465 & 0,6510 & -3679,50 & -3480,23 & 0,0029 & 4,64E-06 & -3689,98 & 72,17 & 0,0276 & 3 \\
\textbf{MUT3} & 0,0438 & 0,0539 & 0,8292 & -3359,13 & -2631,38 & 0,0069 & 9,30E-06 & -3405,58 & 68,53 & 0,2974 & 11 \\
\textbf{MGU} & 0,0373 & 0,0463 & 0,7275 & -3651,80 & -3441,96 & 0,0031 & 4,61E-06 & -3655,85 & 73,17 & 0,0391 & 4 \\
\textbf{MGU-SLIM1} & 0,0714 & 0,0849 & 1,0354 & -2984,17 & -2428,10 & 0,0172 & 2,34E-05 & -2991,50 & 63,83 & 0,7370 & 19 \\
\textbf{MGU-SLIM2} & 0,0373 & 0,0463 & 0,7616 & -3522,46 & -2994,61 & 0,0037 & 5,22E-06 & -3558,84 & 74,67 & 0,1081 & 5 \\
\textbf{MGU-SLIM3} & \textbf{0,0368} & \textbf{0,0455} & \textbf{0,7064} & \textbf{-3667,28} & \textbf{-3410,10} & \textbf{0,0031} & \textbf{4,62E-06} & \textbf{-3675,18} & \textbf{74,63} & \textbf{0,0232} & \textbf{1} \\
\textbf{GRU} & 0,0413 & 0,0512 & 0,8014 & -3517,85 & -3102,46 & 0,0067 & 8,54E-06 & -3531,20 & 70,88 & 0,1865 & 6 \\
\textbf{GRU-SLIM1} & 0,0381 & 0,0475 & 0,6157 & -3390,80 & -2634,37 & 0,0055 & 6,83E-06 & -3423,60 & 72,53 & 0,2018 & 8 \\
\textbf{GRU-SLIM2} & 0,0486 & 0,0602 & 1,1913 & -3371,85 & -3088,15 & 0,0066 & 9,52E-06 & -3381,21 & 69,68 & 0,3075 & 12 \\
\textbf{GRU-SLIM3} & 0,0368 & 0,0456 & 0,6022 & -3676,72 & -3403,73 & 0,0031 & 4,75E-06 & -3689,67 & 72,39 & 0,0270 & 2 \\
\textbf{LSTM-Vanilla} & 0,0388 & 0,0481 & 0,8116 & -3150,71 & -2004,37 & 0,0153 & 1,51E-05 & -3188,35 & 75,38 & 0,4291 & 15 \\
\textbf{LSTM-SLIM1} & 0,0424 & 0,0529 & 0,8197 & -3433,15 & -3024,88 & 0,0066 & 8,48E-06 & -3441,72 & 72,09 & 0,2172 & 10 \\
\textbf{LSTM-SLIM2} & 0,0499 & 0,0606 & 0,8998 & -3291,78 & -2807,97 & 0,0060 & 9,19E-06 & -3320,51 & 69,46 & 0,3268 & 13 \\
\textbf{LSTM-SLIM3} & 0,0428 & 0,0531 & 1,1304 & -3447,89 & -3153,68 & 0,0041 & 6,22E-06 & -3458,40 & 73,50 & 0,1910 & 7 \\ \hline
\end{tabular}%
}
\end{table}

\begin{table}[H]
\centering
\caption{Test results of RNN variants for time series data with random-walk behavior.}
\label{T_R2}
\resizebox{\textwidth}{!}{%
\begin{tabular}{lllllllllllc}
\hline
\multicolumn{1}{l}{} & \multicolumn{1}{c}{\textbf{MAE}} & \multicolumn{1}{c}{\textbf{RMSE}} & \multicolumn{1}{c}{\textbf{TU}} & \multicolumn{1}{c}{\textbf{AIC}} & \multicolumn{1}{c}{\textbf{BIC}} & \multicolumn{1}{c}{\textbf{APC}} & \multicolumn{1}{c}{\textbf{HSP}} & \multicolumn{1}{c}{\textbf{SBIC}} & \multicolumn{1}{c}{\textbf{POCID}} & \multicolumn{1}{c}{\textbf{MCIM}} & \multicolumn{1}{c}{\textbf{Rank}} \\ \hline
\textbf{ELMAN} & 0,0928 & 0,1097 & 21,3471 & -3285,91 & -3246,38 & 0,0250 & 4,15E-05 & -3287,33 & 56,51 & 0,7287 & 18 \\
\textbf{JORDAN} & 0,1020 & 0,1184 & 14,0806 & -3320,03 & -3274,63 & 0,0309 & 5,14E-05 & -3321,53 & 48,45 & 0,8188 & 19 \\
\textbf{MRNN} & 0,0931 & 0,1105 & 66,4363 & -3308,72 & -3203,50 & 0,0296 & 4,82E-05 & -3316,88 & 56,69 & 0,8471 & 20 \\
\textbf{IRNN} & 0,0839 & 0,0997 & 10,9240 & -3549,6 & -3430,98 & 0,0244 & 3,82E-05 & -3560,61 & 56,14 & 0,6121 & 17 \\
\textbf{SCRN} & 0,0262 & 0,0314 & 17,7274 & -4300,43 & -4178,95 & 0,0015 & 2,35E-06 & -4315,80 & 68,48 & 0,0361 & 2 \\
\textbf{MUT1} & 0,0333 & 0,0395 & 29,6909 & -3824,18 & -3339,95 & 0,0032 & 4,44E-06 & -3895,82 & 65,42 & 0,2377 & 15 \\
\textbf{MUT2} & 0,0257 & 0,0308 & 15,3731 & -4031,30 & -3328,85 & 0,0027 & 3,36E-06 & -4114,86 & 65,62 & 0,1395 & 12 \\
\textbf{MUT3} & 0,0245 & 0,0293 & 24,1815 & -4098,09 & -3570,16 & 0,0021 & 2,7E-06 & -4148,32 & 65,94 & 0,1266 & 10 \\
\textbf{MGU} & 0,0263 & 0,0317 & 16,2492 & -4005,74 & -3476,69 & 0,0022 & 2,92E-06 & -4062,89 & 66,98 & 0,1340 & 11 \\
\textbf{MGU-SLIM1} & 0,0236 & 0,0285 & 10,9444 & -4227,87 & -3840,13 & 0,0015 & 2,07E-06 & -4265,43 & 65,07 & 0,0615 & 3 \\
\textbf{MGU-SLIM2} & \textbf{0,0235} & \textbf{0,0282} & \textbf{13,5910} & \textbf{-4352,39} & \textbf{-4070,08} & \textbf{0,0013} & \textbf{2,03E-06} & \textbf{-4382,87} & \textbf{65,59} & \textbf{0,0276} & \textbf{ 1} \\
\textbf{MGU-SLIM3} & 0,0262 & 0,0315 & 11,6589 & -4185,93 & -3995,77 & 0,0015 & 2,31E-06 & -4208,46 & 65,44 & 0,0722 & 5 \\
\textbf{GRU} & 0,0235 & 0,0284 & 17,0903 & -3958,84 & -2981,02 & 0,0029 & 3,36E-06 & -4062,29 & 65,15 & 0,1686 & 13 \\
\textbf{GRU-SLIM1} & 0,0242 & 0,0294 & 16,5218 & -4250,87 & -3986,04 & 0,0013 & 2,02E-06 & -4286,01 & 64,89 & 0,0639 & 4 \\
\textbf{GRU-SLIM2} & 0,0252 & 0,0303 & 27,3054 & -4219,19 & -3933,82 & 0,0014 & 2,15E-06 & -4260,82 & 66,44 & 0,0880 & 7 \\
\textbf{GRU-SLIM3} & 0,0250 & 0,03 & 10,2584 & -4174,49 & -3886,43 & 0,0014 & 2,08E-06 & -4205,25 & 65,17 & 0,0729 & 6 \\
\textbf{LSTM-Vanilla} & 0,0287 & 0,0347 & 35,4234 & -3379,26 & -1698,17 & 0,0088 & 8,84E-06 & -3619,74 & 65,59 & 0,4147 & 16 \\
\textbf{LSTM-SLIM1} & 0,0258 & 0,0315 & 12,9594 & -3756,65 & -2601,97 & 0,0056 & 5,77E-06 & -3900,54 & 65,35 & 0,2364 & 14 \\
\textbf{LSTM-SLIM2} & 0,0255 & 0,0307 & 21,5779 & -4090,69 & -3601,71 & 0,0016 & 2,30E-06 & -4160,11 & 65,66 & 0,1216 & 9 \\
\textbf{LSTM-SLIM3} & 0,0269 & 0,0322 & 19,3579 & -4164,58 & -3898,25 & 0,0017 & 2,56E-06 & -4194,03 & 65,48 & 0,0984 & 8 \\ \hline
\end{tabular}%
}
\end{table}

\begin{table}[H]
\centering
\caption{Test results of RNN variants for time series data with nonlinear behavior.}
\label{T_N2}
\resizebox{\textwidth}{!}{%
\begin{tabular}{lllllllllllc}
\hline
\multicolumn{1}{l}{} & \multicolumn{1}{c}{\textbf{MAE}} & \multicolumn{1}{c}{\textbf{RMSE}} & \multicolumn{1}{c}{\textbf{TU}} & \multicolumn{1}{c}{\textbf{AIC}} & \multicolumn{1}{c}{\textbf{BIC}} & \multicolumn{1}{c}{\textbf{APC}} & \multicolumn{1}{c}{\textbf{HSP}} & \multicolumn{1}{c}{\textbf{SBIC}} & \multicolumn{1}{c}{\textbf{POCID}} & \multicolumn{1}{c}{\textbf{MCIM}} & \multicolumn{1}{c}{\textbf{Rank}} \\ \hline
\textbf{ELMAN} & 0,1149 & 0,1437 & 0,5812 & -2340,32 & -2281,75 & 0,0231 & 3,80E-05 & -2336,61 & 52,67 & 0,1635 & 5 \\
\textbf{JORDAN} & 0,1178 & 0,1480 & 0,6117 & -2279,52 & -2187,23 & 0,0248 & 4,01E-05 & -2277,59 & 48,36 & 0,4874 & 18 \\
\textbf{MRNN} & 0,1146 & 0,1436 & 0,5928 & -2311,45 & -2205,32 & 0,0241 & 3,87E-05 & -2309,68 & 54,07 & 0,1830 & 7 \\
\textbf{IRNN} & 0,1151 & 0,1442 & 0,5844 & -2268,78 & -2054,90 & 0,0273 & 4,14E-05 & -2272,43 & 51,28 & 0,2699 & 12 \\
\textbf{SCRN} & 0,1156 & 0,1441 & 0,6068 & -2260,90 & -2025,20 & 0,0275 & 4,22E-05 & -2263,39 & 54,61 & 0,2932 & 14 \\
\textbf{MUT1} & 0,1133 & 0,1417 & 0,5671 & -2260,10 & -1992,21 & 0,0274 & 4,10E-05 & -2265,44 & 55,08 & 0,0912 & 2 \\
\textbf{MUT2} & 0,1135 & 0,1420 & 0,5684 & -2131,19 & -1579,99 & 0,0362 & 4,86E-05 & -2143,82 & 55,42 & 0,2234 & 8 \\
\textbf{MUT3} & 0,1145 & 0,1432 & 0,5783 & -2056,78 & -1359,81 & 0,0523 & 6,25E-05 & -2080,26 & 53,34 & 0,4447 & 16 \\
\textbf{MGU} & 0,114 & 0,1428 & 0,5800 & -2149,52 & -1668,03 & 0,0311 & 4,45E-05 & -2157,48 & 55,08 & 0,2463 & 10 \\
\textbf{MGU-SLIM1} & 0,1136 & 0,1422 & 0,5716 & -2138,20 & -1603,70 & 0,0320 & 4,50E-05 & -2148,80 & 53,48 & 0,2433 & 9 \\
\textbf{MGU-SLIM2} & 0,1136 & 0,1422 & 0,5707 & -2229,20 & -1904,99 & 0,0296 & 4,31E-05 & -2236,35 & 54,53 & 0,1549 & 4 \\
\textbf{MGU-SLIM3} & \textbf{0,1130} & \textbf{0,1415} & \textbf{0,5662} & \textbf{-2261,98} & \textbf{-1997,78} & \textbf{0,0262} & \textbf{4E-05} & \textbf{-2264,52} & \textbf{54,55} & \textbf{0,0813} & \textbf{1} \\
\textbf{GRU} & 0,1140 & 0,1428 & 0,5805 & -1966,17 & -1085,15 & 0,0420 & 5,36E-05 & -1978,43 & 55,60 & 0,4098 & 15 \\
\textbf{GRU-SLIM1} & 0,1143 & 0,1430 & 0,5794 & -2166,31 & -1725,81 & 0,0313 & 4,47E-05 & -2174,54 & 54,75 & 0,2506 & 11 \\
\textbf{GRU-SLIM2} & 0,1147 & 0,1436 & 0,5786 & -2207,72 & -1860,93 & 0,0339 & 4,72E-05 & -2217,78 & 52,45 & 0,2890 & 13 \\
\textbf{GRU-SLIM3} & 0,1134 & 0,1418 & 0,5681 & -2251,09 & -1971,54 & 0,0262 & 4,01E-05 & -2252,97 & 54,31 & 0,1115 & 3 \\
\textbf{LSTM-Vanilla} & 0,1147 & 0,1437 & 0,5826 & -1820,62 & -626,42 & 0,0704 & 7,81E-05 & -1852,35 & 55,58 & 0,6735 & 20 \\
\textbf{LSTM-SLIM1} & 0,1137 & 0,1424 & 0,5756 & -1925,86 & -942,15 & 0,0547 & 6,42E-05 & -1945,69 & 54,76 & 0,4806 & 17 \\
\textbf{LSTM-SLIM2} & 0,1140 & 0,1425 & 0,5699 & -1890,03 & -822,17 & 0,0688 & 7,62E-05 & -1919,24 & 54,37 & 0,5702 & 19 \\
\textbf{LSTM-SLIM3} & 0,1138 & 0,1424 & 0,5678 & -2277,20 & -2064,31 & 0,0262 & 4,03E-05 & -2279,11 & 50,25 & 0,1730 & 6 \\ \hline
\end{tabular}%
}
\end{table}

\begin{table}[H]
\centering
\caption{Test results of RNN variants for time series data with long-memory behavior.}
\label{T_L2}
\resizebox{\textwidth}{!}{%
\begin{tabular}{lllllllllllc}
\hline
\multicolumn{1}{l}{} & \multicolumn{1}{c}{\textbf{MAE}} & \multicolumn{1}{c}{\textbf{RMSE}} & \multicolumn{1}{c}{\textbf{TU}} & \multicolumn{1}{c}{\textbf{AIC}} & \multicolumn{1}{c}{\textbf{BIC}} & \multicolumn{1}{c}{\textbf{APC}} & \multicolumn{1}{c}{\textbf{HSP}} & \multicolumn{1}{c}{\textbf{SBIC}} & \multicolumn{1}{c}{\textbf{POCID}} & \multicolumn{1}{c}{\textbf{MCIM}} & \multicolumn{1}{c}{\textbf{Rank}} \\ \hline
\textbf{ELMAN} & 0,0831 & 0,1041 & 0,7994 & -2613,9 & -2376,66 & 0,0139 & 2,10E-05 & -2609,74 & 40,36 & 0,1422 & 10 \\
\textbf{JORDAN} & 0,0879 & 0,1100 & 0,8836 & -2606,64 & -2540,76 & 0,0134 & 2,19E-05 & -2601,62 & 36,50 & 0,2931 & 14 \\
\textbf{MRNN} & 0,0952 & 0,1189 & 1,2506 & -2445,10 & -2219,43 & 0,0178 & 2,75E-05 & -2441,99 & 43,74 & 0,5397 & 19 \\
\textbf{IRNN} & 0,0875 & 0,1095 & 0,8701 & -2597,71 & -2435,18 & 0,0139 & 2,21E-05 & -2593,91 & 37,10 & 0,2889 & 13 \\
\textbf{SCRN} & 0,0829 & 0,1039 & 0,7958 & -2666,62 & -2548,06 & 0,0121 & 1,95E-05 & -2661,75 & 40,95 & 0,0738 & 4 \\
\textbf{MUT1} & 0,0834 & 0,1043 & 0,8052 & -2618,77 & -2526,96 & 0,0120 & 1,99E-05 & -2613,97 & 39,67 & 0,1259 & 8 \\
\textbf{MUT2} & 0,0829 & 0,1038 & 0,7943 & -2573,52 & -2239,69 & 0,0140 & 2,10E-05 & -2570,23 & 40,41 & 0,1632 & 11 \\
\textbf{MUT3} & 0,0844 & 0,1064 & 0,8395 & -2142,53 & -1053,80 & 0,0288 & 3,46E-05 & -2144,50 & 43,41 & 0,5989 & 20 \\
\textbf{MGU} & 0,0818 & 0,1025 & 0,7749 & -2575,53 & -2427,40 & 0,0122 & 1,99E-05 & -2570,85 & 42,02 & 0,0850 & 5 \\
\textbf{MGU-SLIM1} & 0,0823 & 0,1032 & 0,7870 & -2488,92 & -1997,73 & 0,0158 & 2,27E-05 & -2487,78 & 42,37 & 0,1973 & 12 \\
\textbf{MGU-SLIM2} & \textbf{0,0819} & \textbf{0,1026} & \textbf{0,7772} & \textbf{-2642,25} & \textbf{-2519,68} & \textbf{0,0119} & \textbf{1,94E-05} & \textbf{-2637,50} & \textbf{42,18} & \textbf{0,0438} & \textbf{1} \\
\textbf{MGU-SLIM3} & 0,0817 & 0,1024 & 0,7738 & -2627,20 & -2445,44 & 0,0122 & 1,95E-05 & -2622,88 & 42,08 & 0,0566 & 2 \\
\textbf{GRU} & 0,0830 & 0,1039 & 0,7970 & -2400,06 & -1683,90 & 0,0226 & 2,82E-05 & -2398,57 & 40,06 & 0,3815 & 15 \\
\textbf{GRU-SLIM1} & 0,0823 & 0,1030 & 0,7820 & -2579,70 & -2435,88 & 0,0122 & 2E-05 & -2574,98 & 40,96 & 0,1085 & 6 \\
\textbf{GRU-SLIM2} & 0,0826 & 0,1035 & 0,7907 & -2647,31 & -2564,15 & 0,0118 & 1,94E-05 & -2642,32 & 41,07 & 0,0693 & 3 \\
\textbf{GRU-SLIM3} & 0,0819 & 0,1026 & 0,7759 & -2565,52 & -2232,18 & 0,0150 & 2,19E-05 & -2562,48 & 42,25 & 0,1294 & 9 \\
\textbf{LSTM-Vanilla} & 0,0834 & 0,1046 & 0,8042 & -2311,82 & -1517,30 & 0,0312 & 3,66E-05 & -2311,27 & 42,63 & 0,5036 & 18 \\
\textbf{LSTM-SLIM1} & 0,0826 & 0,1035 & 0,7890 & -2252,23 & -1304,66 & 0,0270 & 3,25E-05 & -2254,49 & 41,19 & 0,4992 & 17 \\
\textbf{LSTM-SLIM2} & 0,0823 & 0,1032 & 0,7847 & -2241,02 & -1373,81 & 0,0260 & 3,17E-05 & -2242,06 & 42,81 & 0,4575 & 16 \\
\textbf{LSTM-SLIM3} & 0,0816 & 0,1023 & 0,7719 & -2535,86 & -2287,80 & 0,0129 & 2,04E-05 & -2532,17 & 42,25 & 0,1125 & 7 \\ \hline
\end{tabular}%
}
\end{table}

\begin{table}[H]
\centering
\caption{Test results of RNN variants for time series data with Chaotic behavior.}
\label{T_C2}
\resizebox{\textwidth}{!}{%
\begin{tabular}{lllllllllllc}
\hline
\multicolumn{1}{l}{} & \multicolumn{1}{c}{\textbf{MAE}} & \multicolumn{1}{c}{\textbf{RMSE}} & \multicolumn{1}{c}{\textbf{TU}} & \multicolumn{1}{c}{\textbf{AIC}} & \multicolumn{1}{c}{\textbf{BIC}} & \multicolumn{1}{c}{\textbf{APC}} & \multicolumn{1}{c}{\textbf{HSP}} & \multicolumn{1}{c}{\textbf{SBIC}} & \multicolumn{1}{c}{\textbf{POCID}} & \multicolumn{1}{c}{\textbf{MCIM}} & \multicolumn{1}{c}{\textbf{Rank}} \\ \hline
\textbf{ELMAN} & 0,0611 & 0,0768 & 2,8365 & -3555,56 & -3445,75 & 0,0105 & 1,66E-05 & -3553,07 & 56,78 & 0,3312 & 13 \\
\textbf{JORDAN} & 0,0614 & 0,0776 & 0,8107 & -3542,07 & -3449,85 & 0,0102 & 1,67E-05 & -3540,03 & 55,82 & 0,3344 & 14 \\
\textbf{MRNN} & 0,0606 & 0,0767 & 1,4364 & -3449,87 & -3293,96 & 0,0112 & 1,70E-05 & -3449,98 & 55,98 & 0,3963 & 16 \\
\textbf{IRNN} & 0,0680 & 0,0842 & 0,6610 & -3477,67 & -3112,90 & 0,0161 & 2,37E-05 & -3484,68 & 54,71 & 0,5457 & 20 \\
\textbf{SCRN} & 0,0556 & 0,0699 & 0,6424 & -3637,80 & -3422,59 & 0,0094 & 1,45E-05 & -3639,88 & 55,80 & 0,1798 & 5 \\
\textbf{MUT1} & 0,0576 & 0,0726 & 0,5842 & -3693,89 & -3584,17 & 0,0093 & 1,51E-05 & -3694,28 & 56,45 & 0,1521 & 2 \\
\textbf{MUT2} & 0,0609 & 0,0760 & 0,5364 & -3514,34 & -2980,48 & 0,0160 & 2,17E-05 & -3527,04 & 57,72 & 0,3474 & 15 \\
\textbf{MUT3} & 0,0533 & 0,0671 & 0,5245 & -3513,64 & -2833,53 & 0,0187 & 2,21E-05 & -3540,32 & 57,19 & 0,2659 & 6 \\
\textbf{MGU} & 0,0539 & 0,0679 & 0,5391 & -3602,20 & -3153,47 & 0,0117 & 1,62E-05 & -3607,21 & 57,16 & 0,1701 & 3 \\
\textbf{MGU-SLIM1} & 0,0606 & 0,0757 & 0,5782 & -3613,21 & -3341,08 & 0,0122 & 1,83E-05 & -3620,21 & 56,19 & 0,2794 & 8 \\
\textbf{MGU-SLIM2} & 0,0540 & 0,0679 & 0,5748 & -3560,06 & -3021,49 & 0,0105 & 1,52E-05 & -3579,54 & 58,07 & 0,1729 & 4 \\
\textbf{MGU-SLIM3} & 0,0600 & 0,0752 & 7,0372 & -3529,90 & -3120,72 & 0,0143 & 2,07E-05 & -3541,90 & 55,87 & 0,4650 & 18 \\
\textbf{GRU} & 0,0536 & 0,0676 & 0,4793 & -3406,40 & -2458,13 & 0,0189 & 2,25E-05 & -3433,64 & 59,22 & 0,3104 & 11 \\
\textbf{GRU-SLIM1} & 0,0603 & 0,0753 & 0,5461 & -3549,65 & -3086,89 & 0,0122 & 1,83E-05 & -3570,81 & 56,66 & 0,3129 & 12 \\
\textbf{GRU-SLIM2} & 0,0587 & 0,0740 & 0,5543 & -3519,91 & -2989,82 & 0,0115 & 1,73E-05 & -3545,10 & 57,80 & 0,2836 & 9 \\
\textbf{GRU-SLIM3} & 0,0602 & 0,0752 & 0,5664 & -3593,38 & -3290,79 & 0,0127 & 1,87E-05 & -3602,37 & 56,40 & 0,2851 & 10 \\
\textbf{LSTM-Vanilla} & 0,0532 & 0,0671 & 0,4772 & -3475,04 & -2677,35 & 0,0506 & 4,94E-05 & -3486,80 & 59,06 & 0,4290 & 17 \\
\textbf{LSTM-SLIM1} & 0,0527 & 0,0664 & 0,4974 & -3434,01 & -2545,32 & 0,0122 & 1,65E-05 & -3476,53 & 57,73 & 0,2696 & 7 \\
\textbf{LSTM-SLIM2} & 0,0533 & 0,0674 & 0,5045 & -3290,49 & -2110,00 & 0,0320 & 3,36E-05 & -3330,35 & 58,74 & 0,4784 & 19 \\
\textbf{LSTM-SLIM3} & \textbf{0,0532} & \textbf{0,0670} & \textbf{0,4927} & \textbf{-3701,04} & \textbf{-3370,02} & \textbf{0,0102} & \textbf{1,49E-05} & \textbf{-3708,28} & \textbf{57,05} & \textbf{0,0804} & \textbf{1} \\ \hline
\end{tabular}%
}
\end{table}

\begin{figure}[H]
     \centering
     \begin{subfigure}[b]{0.5\textwidth}
         \centering
         \includegraphics[height=8.5cm,width=8.5cm]{TU_Exp2_Deterministic_bhv.jpg}
         \caption{Distribution of four ranges of TU values.}
         \label{F_D2_a}
     \end{subfigure}
     \begin{subfigure}[b]{0.4\textwidth}
         \centering
         \includegraphics[height=8.3cm,width=8.5cm]{POCID_Exp2_Deterministic_bhv.jpg}
         \caption{Mean and Std of POCID.}
         \label{F_D2_b}
     \end{subfigure}
    \caption{TU and POCID test results obtained by RNN cell structures on deterministic time series.}
    \label{F_D2}
\end{figure}

\begin{figure}[H]
     \centering
     \begin{subfigure}[b]{0.5\textwidth}
         \centering
         \includegraphics[height=8.5cm,width=8.5cm]{TU_Exp2_Unit_root_bhv.jpg}
         \caption{Four ranges of TU values.}
         \label{F_R2_a}
     \end{subfigure}
     \begin{subfigure}[b]{0.4\textwidth}
         \centering
         \includegraphics[height=8.3cm,width=8.5cm]{POCID_Exp2_Unit_root_bhv.jpg}
         \caption{Mean and Std of POCID.}
         \label{F_R2_b}
     \end{subfigure}
    \caption{TU and POCID test results obtained by RNN cell structures on random-walk time series}
    \label{F_R2}
\end{figure}

\begin{figure}[H]
     \centering
     \begin{subfigure}[b]{0.5\textwidth}
         \centering
         \includegraphics[height=8.5cm,width=8.5cm]{TU_Exp2_NonLinear_bhv.jpg}
         \caption{Four ranges of TU values.}
         \label{F_N2_a}
     \end{subfigure}
     \begin{subfigure}[b]{0.4\textwidth}
         \centering
         \includegraphics[height=8.3cm,width=8.5cm]{POCID_Exp2_NonLinear_bhv.jpg}
         \caption{Mean and Std of POCID.}
         \label{F_N2_b}
     \end{subfigure}
    \caption{TU and POCID test results obtained by RNN cell structures on nonlinear time series}
    \label{F_N2}
\end{figure}

\begin{figure}[H]
     \centering
     \begin{subfigure}[b]{0.5\textwidth}
         \centering
         \includegraphics[height=8.5cm,width=8.5cm]{TU_Exp2_LongMemory_bhv.jpg}
         \caption{Four ranges of TU values.}
         \label{F_L2_a}
     \end{subfigure}
     \begin{subfigure}[b]{0.4\textwidth}
         \centering
         \includegraphics[height=8.3cm,width=8.5cm]{POCID_Exp2_LongMemory_bhv.jpg}
         \caption{Mean and Std of POCID.}
         \label{F_L2_b}
     \end{subfigure}
    \caption{TU and POCID test results obtained by RNN cell structures on long-memory time series}
    \label{F_L2}
\end{figure}

\begin{figure}[H]
     \centering
     \begin{subfigure}[b]{0.5\textwidth}
         \centering
         \includegraphics[height=8.5cm,width=8.5cm]{TU_Exp2_Chaotic_bhv.jpg}
         \caption{Four ranges of TU values.}
         \label{F_C2_a}
     \end{subfigure}
     \begin{subfigure}[b]{0.4\textwidth}
         \centering
         \includegraphics[height=8.3cm,width=8.5cm]{POCID_Exp2_Chaotic_bhv.jpg}
         \caption{Mean and Std of POCID.}
         \label{F_C2_b}
     \end{subfigure}
        \caption{TU and POCID test results obtained by RNN cell structures on chaotic time series}
        \label{F_C2}
\end{figure}

The results of the two experiments are summarized in Table \ref{final}. The MUT2, the SCRN, and the ELMAN models are the most recommended RNNs to forecast time series data with deterministic, random-walk, and nonlinear behaviors, respectively. Whereas, the MGU-SLIM2 and the LSTM-SLIM3 are the most recommended for the long-memory and chaotic behaviors, respectively.

\begin{table}[H]
\centering
\caption{Summary of the test results for the two experiments.}
\label{final}
\begin{tabular}{llrlrl}\hline
\multicolumn{1}{c}{\multirow{2}{*}{Behaviors}} & \multicolumn{2}{c}{Experiment 1} & \multicolumn{2}{c}{Experiment 2} & \multicolumn{1}{c}{\multirow{2}{*}{Recommended cell}} \\
\multicolumn{1}{c}{} & Best cell & MCIM & Best cell & MCIM & \multicolumn{1}{c}{} \\\hline
Deterministic & CIFG & 0,1210 & MUT2 & 0,0276 & MUT2 \\
Random-walk & CIFG & 0,0830 & SCRN & 0,0361 & SCRN \\
Nonlinear & NOG & 0,2187 & ELMAN & 0,1635 & ELMAN \\
Long-memory & NIG & 0,0851 & MGU-SLIM2 & 0,0438 & MGU-SLIM2 \\
Chaotic & NOG & 0,2355 & LSTM-SLIM3 & 0,0804 & LSTM-SLIM3 \\\hline
\end{tabular}
\end{table}

\section{Conclusions}
In this paper, we proposed a comprehensive taxonomy of all possible time series behaviors, which are: deterministic, random-walk, nonlinear, long-memory, and chaotic. Then, we conducted two experiments to show the best RNN cell structure for each behavior. In the first experiment, we evaluated the LSTM-Vanilla model and 11 of its variants created based on one alteration in its basic architecture that consists in (1) removing (NIG, NFG, NOG, NIAF, NFAF, NOAF, and NCAF), (2) adding (PC and FGR), or (3) substituting (FB1 and CIFG) one cell component. While, in the second experiment, we evaluated LSTM-Vanilla along with a set of 19 RNN models based on other recurrent cell structures (JORDAN, ELMAN, MRNN, SCRN, IRNN, GRU, MGU, MUT1, MUT2, MUT3, and 9 SLIM variants). 

To evaluate, compare, and select the best model, different statistical metrics were used: error-based metrics (MAE and RMSE), information criterion-based metrics (AIC, BIC, APC, HSP, and SBIC), naïve-based metric (TU), and direction change-based metric (POCID). To facilitate the task of the best model selection, a new statistical metric was proposed (MCIM). Further to improve our confidence in the models’ interpretation and selection, Friedman Wilcoxon-Holm signed-rank test was used.  

In the first experiment, We showed that the CIFG model is the most suitable for non-stationary time series due to the existence of deterministic behavior or random-walk behavior. We also, experimentally, proved that the NOG model is the most performing for the nonlinear and chaotic behaviors, while the NIG model outperformed the other models with respect to long-memory behavior. 
In the second experiment, over the 20 evaluated RNN models, the best forecasting results were achieved by the new parameter-reduced variants of MGU (MGU-SLIM2) and LSTM (LSTM-SLIM3) for the long-memory and chaotic behaviors, respectively. For the deterministic behavior, the best significant model is MUT2. In the case of the random-walk behavior, the most significantly better model is SCRN. Finally, for the nonlinear behavior, the ELMAN model is the best significant model.   

Based on the outcomes of both experiments, we arrived to demonstrate that the SLIM3 version of the LSTM cell has the highest ability to increase the performance of the RNN model in forecasting chaotic behavior. While the SLIM2 version of the MGU model is recommended in the case of time series with long-memory behavior. Finally, the MUT2, SCRN, and ELMAN variants are the strongly advocated models to forecast time series data with deterministic, random-walk, and nonlinear behaviors, respectively.     
The outcomes of our study are limited to time series with a single behavior. However, in real-world problems, combined behaviors (i.e., more than one behavior in the same time series) can also occur. In future work, evaluating the best RNN cell with respect to such types of time series can complement the guidelines provided by this study.

\section*{Acknowledgements}
This work was partially supported by DETECTOR (A-RNM-256-UGR18 Universidad de Granada/FEDER), LifeWatch SmartEcomountains (LifeWatch-2019-10-UGR-01 Ministerio de Ciencia e Innovación/Universidad de Granada/FEDER),  DeepL-ISCO (A-TIC-458-UGR18 Ministerio de Ciencia e Innovación/FEDER),  BigDDL-CET (P18-FR-4961 Ministerio de Ciencia e Innovación/Universidad de Granada/FEDER).

\bibliography{sample}

\pagebreak

\appendix
\section{Architectures of the studied RNN cells }
In this section, we provide the cell structures of the different RNN models along with their cellular calculations. To understand the calculations, Table \ref{Nom} presents the deception of the mathematical notations.  
To better understand the components inside the LSTM-Vanilla cell, we present below the role of the main elements:
\begin{itemize}
\setcounter{enumi}{0}
\item Input state $x_t$: it contains the data features at time step $t$. 
\item Output state $y_t$: it contains the output of the model at time step $t$.
\item Hidden state $h_t$: it represents the short-term memory of the cell at time step $t$.
\item Cell state $c_t$: it represents the long-term memory of the cell at time step $t$. 
\item Candidate cell state $\tilde{c_t}$: it contains the new information we can use to update the cell state at time step $t$.
\item Input gate $\Gamma_{i_t}$: it filters from the current candidate cell state the information that should be used to update the current cell state.  
\item Forget gate $\Gamma_{f_t}$: it filters from the previous cell state the information that should be used to update the current cell state.
\item Output gate $\Gamma_{o_t}$: it filters from the current cell state the information that should be exposed to the external network (the next time step and the next hidden and/or output layer).
\end{itemize}

\begin{longtable}[c]{ll}
\caption{Nomenclature}
\label{Nom}\\ \hline
Symbol & Significance \\ \hline
\endhead
$x_t$ & the input state at time step t.\\
$h_t$ & the hidden state at time step t. \\
$y_t$ & the output state at time step t. \\
$\Gamma_{i_t}$ & the input gate at time step t.\\
$\Gamma_{f_t}$ & the forget gate at time step t.\\
$\Gamma_{o_t}$ & the output gate at time step t.\\
$\Gamma_{u_t}$ & the update gate at time step t.\\
$\Gamma_{r_t}$ & the relevance gate at time step t.\\
$c_t$ & the cell sate at time step t.\\
$\tilde{c}_t$ & the candidate cell sate at time step t.\\
$W_{xh}$ & the weight matrix between the input and the hidden states. \\
$W_{hh}$ & the weight matrix between the previous and the current hidden states. \\\hline
$W_{hy}$ & the weight matrix between the hidden and the output states. \\
$W_{yh}$ & the weight matrix between the output and the hidden states. \\
$W_{xs}$ & the weight matrix between the input and the context states. \\
$W_{sh}$ & the weight matrix between the context and the hidden states. \\
$W_{sy}$ & the weight matrix between the context and the output states. \\
$W_{xi}$ & the weight matrix between the input state and the input gate. \\
$W_{hi}$ & the weight matrix between the hidden state and the input gate. \\
$W_{xo}$ & the weight matrix between the input state and the output gate. \\
$W_{ho}$ & the weight matrix between the hidden state and the output gate. \\
$W_{xf}$ & the weight matrix between the input state and the forget gate. \\
$W_{hf}$ & the weight matrix between the hidden state and the forget gate.\\
$W_{x \tilde{c}}$ & the weight matrix between the input state and the candidate cell state.\\
$W_{h \tilde{c}}$ & the weight matrix between the hidden state and the candidate cell state.\\
$W_{x \tilde{h}}$ & the weight matrix between the input state and the candidate hidden state. \\
$W_{h \tilde{h}}$ & the weight matrix between the hidden state and the candidate hidden state. \\
$W_{ci}$ & the weight matrix between the cell state and the input gate. \\
$W_{cf}$ & the weight matrix between the cell state and the forget gate. \\
$W_{co}$ & the weight matrix between the cell state and the output gate. \\
$W_{xu}$ & the weight matrix between the input state and the update gate. \\
$W_{hu}$  & the weight matrix between the hidden state and the update gate. \\
$W_{xr}$ & the weight matrix between the input state and the relevance gate. \\
$W_{hr}$ & the weight matrix between the hidden state and the relevance gate. \\
$W_{ii}$ & the weight matrix between the previous and the current input gates. \\
$W_{ff}$ & the weight matrix between the previous and the current forget gates. \\
$W_{oo}$ & the weight matrix between the previous and the current output gates. \\
$W_{if}$ & the weight matrix between the previous input gate and the current forget gate. \\
$W_{io}$ & the weight matrix between the previous input gate and the current output gate. \\
$W_{fi}$ & the weight matrix between the previous forget gate and the current input gate. \\
$W_{fo}$ & the weight matrix between the previous forget gate and the current output gate. \\
$W_{oi}$ & the weight matrix between the previous output gate and the current input gate. \\
$W_{of}$ & the weight matrix between the previous output gate and the current forget gate. \\
$b_{h}$ & the bias related to the hidden state. \\
$b_{y}$ & the bias related to the output state. \\
$b_{\tilde{c}}$ & the bias related to the candidate cell state. \\\hline
$b_{\tilde{h}}$ & the bias related to the candidate hidden state. \\
$b_{i}$ & the bias related to the input gate. \\
$b_{o}$ & the bias related to the output gate. \\
$b_{f}$ & the bias related to the forget gate. \\
$b_{u}$ & the bias related to the update gate. \\
$b_{r}$ & the bias related to the relevance gate. \\
$g$ & the activation function of the output state (the identity function). \\
$\otimes$ & the point-wise multiplication (Hadamard product) presented in the figures as $\odot$. \\
$\sigma$ & the sigmoid activation function. \\
\hline
\end{longtable}

\begin{longtable}[H]{c >{\centering\arraybackslash} m{6cm} c}
\caption{Description of different LSTM-Vanilla cell structures created based on one change in its architecture.}
\label{tab1}
\\\hline
\multicolumn{1}{c}{Cell name} & \multicolumn{1}{c}{Cell architecture} & \multicolumn{1}{c}{Cell computations} \\ \hline\hline

\endfirsthead
\caption{ -- \textit{Continued from previous page}} \\\hline
\multicolumn{1}{c}{Cell name} & \multicolumn{1}{c}{Cell architecture} & \multicolumn{1}{c}{Cell computations} \\ \hline\hline
\endhead
\hline \\
\endfoot
\hline
\endlastfoot
    
\multirow{3}{*}{LSTM-Vanilla} & \multirow{3}{*}{\includegraphics[height=3.8cm,width=5cm]{LSTM-V.jpg}} 
    & $ \tilde{c_t} =  \tanh(x_t.W_{x\tilde{c}} + h_{t-1}.W_{h\tilde{c}} + b_{\tilde{c}}) $ \\  
 &  & $ \Gamma_{i_t} = \sigma(x_t.W_{xi} + h_{t-1}.W_{hi} + b_i) $ \\
 &  & $ \Gamma_{f_t} = \sigma(x_t.W_{xf} + h_{t-1}.W_{hf} + b_f) $ \\ 
 &  & $ \Gamma_{o_t} = \sigma(x_t.W_{xo} + h_{t-1}.W_{ho} + b_o) $ \\
 &  & $ c_t =\Gamma_{f_t} \otimes c_{t-1}+\Gamma_{i_t} \otimes \tilde{c_t} $ \\
 &  & $ h_t = \Gamma_{o_t} \otimes \tanh(c_t) $
\\\hline

\multirow{3}{*}{LSTM-NIG} & \multirow{3}{*}{\includegraphics[height=3.8cm,width=5cm]{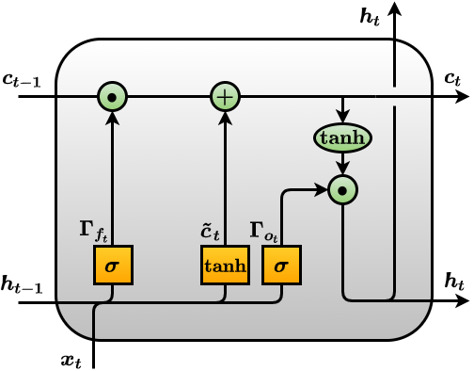}} 
    & $ \tilde{c_t} =  \tanh(x_t.W_{x\tilde{c}} + h_{t-1}.W_{h\tilde{c}} + b_{\tilde{c}}) $ \\  
 &  & $ \Gamma_{f_t} = \sigma(x_t.W_{xf} + h_{t-1}.W_{hf} + b_f) $ \\ 
 &  & $ \Gamma_{o_t} = \sigma(x_t.W_{xo} + h_{t-1}.W_{ho} + b_o) $ \\
 &  & $ c_t =\Gamma_{f_t} \otimes c_{t-1}+ \tilde{c_t} $ \\
 &  & $ h_t = \Gamma_{o_t} \otimes \tanh(c_t) $ \\
 &  &
\\\hline

\multirow{3}{*}{LSTM-NFG} & \multirow{3}{*}{\includegraphics[height=3.8cm,width=5cm]{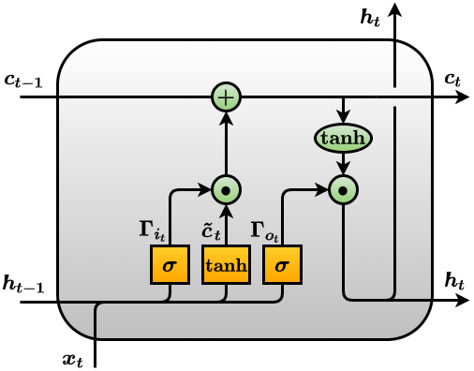}} 
    & $ \tilde{c_t} =  \tanh(x_t.W_{x\tilde{c}} + h_{t-1}.W_{h\tilde{c}} + b_{\tilde{c}}) $ \\  
 &  & $ \Gamma_{i_t} = \sigma(x_t.W_{xi} + h_{t-1}.W_{hi} + b_i) $ \\
 &  & $ \Gamma_{o_t} = \sigma(x_t.W_{xo} + h_{t-1}.W_{ho} + b_o) $ \\
 &  & $ c_t = c_{t-1}+\Gamma_{i_t} \otimes \tilde{c_t} $ \\
 &  & $ h_t = \Gamma_{o_t} \otimes \tanh(c_t) $ \\
  &  &
\\\hline

\pagebreak

\multirow{3}{*}{LSTM-NOG} & \multirow{3}{*}{\includegraphics[height=3.8cm,width=5cm]{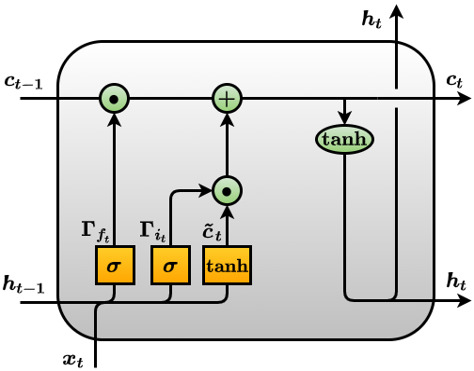}} 
    & $ \tilde{c_t} =  \tanh(x_t.W_{x\tilde{c}} + h_{t-1}.W_{h\tilde{c}} + b_{\tilde{c}}) $ \\  
 &  & $ \Gamma_{i_t} = \sigma(x_t.W_{xi} + h_{t-1}.W_{hi} + b_i) $ \\
 &  & $ \Gamma_{f_t} = \sigma(x_t.W_{xf} + h_{t-1}.W_{hf} + b_f) $ \\ 
 &  & $ c_t =\Gamma_{f_t} \otimes c_{t-1}+\Gamma_{i_t} \otimes \tilde{c_t} $ \\
 &  & $ h_t = \tanh(c_t) $ \\
  &  &
\\\hline

\multirow{3}{*}{LSTM-NIAF} & \multirow{3}{*}{\includegraphics[height=3.8cm,width=5cm]{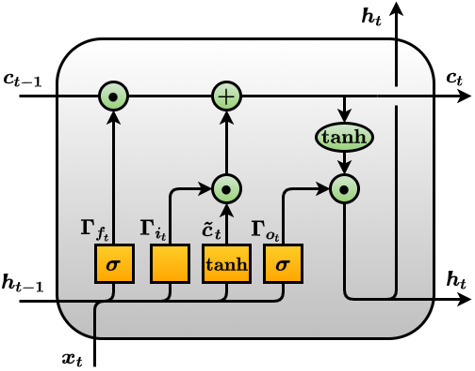}} 
    & $ \tilde{c_t} =  \tanh(x_t.W_{x\tilde{c}} + h_{t-1}.W_{h\tilde{c}} + b_{\tilde{c}}) $ \\  
 &  & $ \Gamma_{i_t} = x_t.W_{xi} + h_{t-1}.W_{hi} + b_i $ \\
 &  & $ \Gamma_{f_t} = \sigma(x_t.W_{xf} + h_{t-1}.W_{hf} + b_f) $ \\ 
 &  & $ \Gamma_{o_t} = \sigma(x_t.W_{xo} + h_{t-1}.W_{ho} + b_o) $ \\
 &  & $ c_t =\Gamma_{f_t} \otimes c_{t-1}+\Gamma_{i_t} \otimes \tilde{c_t} $ \\
 &  & $ h_t = \Gamma_{o_t} \otimes \tanh(c_t) $
\\\hline

\multirow{3}{*}{LSTM-NFAF} & \multirow{3}{*}{\includegraphics[height=3.8cm,width=5cm]{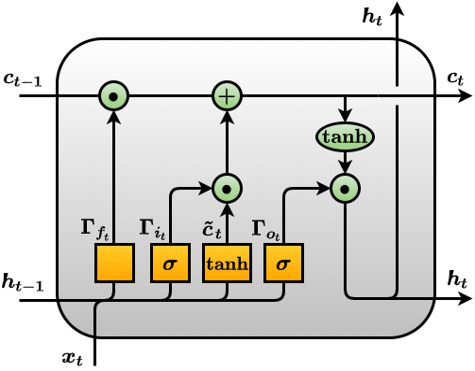}} 
    & $ \tilde{c_t} =  \tanh(x_t.W_{x\tilde{c}} + h_{t-1}.W_{h\tilde{c}} + b_{\tilde{c}}) $ \\  
 &  & $ \Gamma_{i_t} = \sigma(x_t.W_{xi} + h_{t-1}.W_{hi} + b_i) $ \\
 &  & $ \Gamma_{f_t} = x_t.W_{xf} + h_{t-1}.W_{hf} + b_f $ \\ 
 &  & $ \Gamma_{o_t} = \sigma(x_t.W_{xo} + h_{t-1}.W_{ho} + b_o) $ \\
 &  & $ c_t =\Gamma_{f_t} \otimes c_{t-1}+\Gamma_{i_t} \otimes \tilde{c_t} $ \\
 &  & $ h_t = \Gamma_{o_t} \otimes \tanh(c_t) $
\\\hline

\multirow{3}{*}{LSTM-NOAF} & \multirow{3}{*}{\includegraphics[height=3.8cm,width=5cm]{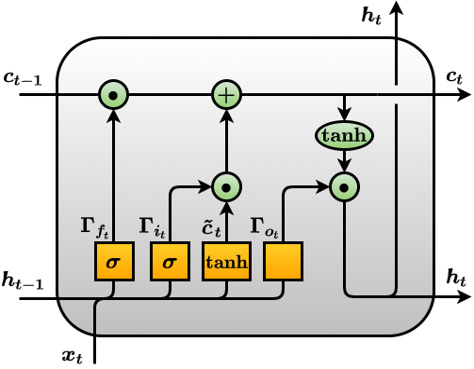}} 
    & $ \tilde{c_t} =  \tanh(x_t.W_{x\tilde{c}} + h_{t-1}.W_{h\tilde{c}} + b_{\tilde{c}}) $ \\  
 &  & $ \Gamma_{i_t} = \sigma(x_t.W_{xi} + h_{t-1}.W_{hi} + b_i) $ \\
 &  & $ \Gamma_{f_t} = \sigma(x_t.W_{xf} + h_{t-1}.W_{hf} + b_f) $ \\ 
 &  & $ \Gamma_{o_t} = x_t.W_{xo} + h_{t-1}.W_{ho} + b_o $ \\
 &  & $ c_t =\Gamma_{f_t} \otimes c_{t-1}+\Gamma_{i_t} \otimes \tilde{c_t} $ \\
 &  & $ h_t = \Gamma_{o_t} \otimes \tanh(c_t) $
\\\hline

\multirow{3}{*}{LSTM-NCAF} & \multirow{3}{*}{\includegraphics[height=3.8cm,width=5cm]{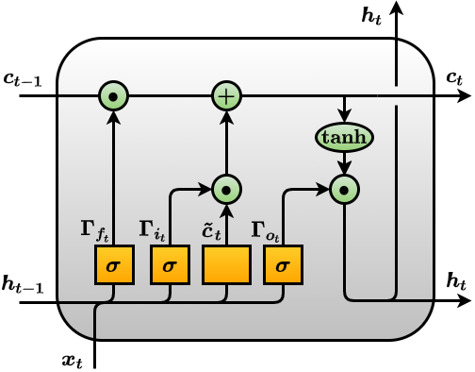}} 
    & $ \tilde{c_t} =  x_t.W_{x\tilde{c}} + h_{t-1}.W_{h\tilde{c}} + b_{\tilde{c}} $ \\  
 &  & $ \Gamma_{i_t} = \sigma(x_t.W_{xi} + h_{t-1}.W_{hi} + b_i) $ \\
 &  & $ \Gamma_{f_t} = \sigma(x_t.W_{xf} + h_{t-1}.W_{hf} + b_f) $ \\ 
 &  & $ \Gamma_{o_t} = \sigma(x_t.W_{xo} + h_{t-1}.W_{ho} + b_o) $ \\
 &  & $ c_t =\Gamma_{f_t} \otimes c_{t-1}+\Gamma_{i_t} \otimes \tilde{c_t} $ \\
 &  & $ h_t = \Gamma_{o_t} \otimes \tanh(c_t) $
\\\hline
\pagebreak
\multirow{3}{*}{LSTM-FB1} & \multirow{3}{*}{\includegraphics[height=3.8cm,width=5cm]{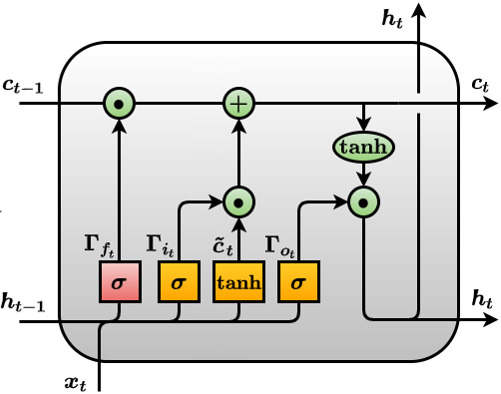}} 
    & $ \tilde{c_t} =  \tanh(x_t.W_{x\tilde{c}} + h_{t-1}.W_{h\tilde{c}} + b_{\tilde{c}}) $ \\  
 &  & $ \Gamma_{i_t} = \sigma(x_t.W_{xi} + h_{t-1}.W_{hi} + b_i) $ \\
 &  & $ b_f = 1 $ \\ 
 &  & $ \Gamma_{f_t} = \sigma(x_t.W_{xf} + h_{t-1}.W_{hf} + b_f) $ \\ 
 &  & $ \Gamma_{o_t} = \sigma(x_t.W_{xo} + h_{t-1}.W_{ho} + b_o) $ \\
 &  & $ c_t =\Gamma_{f_t} \otimes c_{t-1}+\Gamma_{i_t} \otimes \tilde{c_t} $ \\
 &  & $ h_t = \Gamma_{o_t} \otimes \tanh(c_t) $
\\\hline

\multirow{3}{*}{LSTM-CIFG} & \multirow{3}{*}{\includegraphics[height=3.8cm,width=5cm]{LSTM-CIFG.jpg}} 
    & $ \tilde{c_t} =  \tanh(x_t.W_{x\tilde{c}} + h_{t-1}.W_{h\tilde{c}} + b_{\tilde{c}}) $ \\  
 &  & $ \Gamma_{i_t} = \sigma(x_t.W_{xi} + h_{t-1}.W_{hi} + b_i) $ \\
 &  & $ \Gamma_{o_t} = \sigma(x_t.W_{xo} + h_{t-1}.W_{ho} + b_o) $ \\
 &  & $ c_t =(1-\Gamma_{i_t}) \otimes c_{t-1}+\Gamma_{i_t} \otimes \tilde{c_t} $ \\
 &  & $ h_t = \Gamma_{o_t} \otimes \tanh(c_t) $ \\
 &  &  
\\\hline

\multirow{3}{*}{LSTM-PC} & \multirow{3}{*}{\includegraphics[height=3.8cm,width=5cm]{LSTM-PC.jpg}} 
    & $ \tilde{c_t} =  \tanh(x_t.W_{x\tilde{c}} + h_{t-1}.W_{h\tilde{c}} + b_{\tilde{c}}) $ \\  
 &  & $ \Gamma_{i_t} = \sigma(x_t.W_{xi} + h_{t-1}.W_{hi} + c_{t-1}.W_{ci} + b_i) $ \\
 &  & $ \Gamma_{f_t} = \sigma(x_t.W_{xf} + h_{t-1}.W_{hf} + c_{t-1}.W_{cf} + b_f) $ \\ 
  &  & $ \Gamma_{o_t} = \sigma(x_t.W_{xo} + h_{t-1}.W_{ho} + c_{t}.W_{co} + b_o) $ \\
 &  & $ c_t =\Gamma_{f_t} \otimes c_{t-1}+\Gamma_{i_t} \otimes \tilde{c_t} $ \\
 &  & $ h_t = \Gamma_{o_t} \otimes \tanh(c_t) $
\\\hline

\multirow{3}{*}{LSTM-FGR} & \multirow{3}{*}{\includegraphics[height=4.5cm,width=6cm]{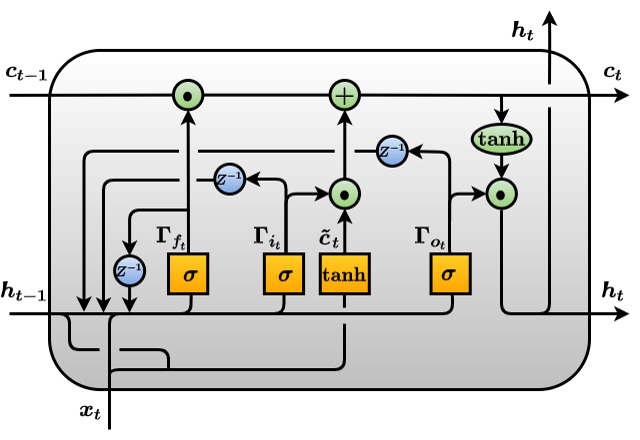}}
    & $ \tilde{c_t} =  \tanh(x_t.W_{x\tilde{c}} + h_{t-1}.W_{h\tilde{c}} + b_{\tilde{c}}) $ \\  
 &  & 
 $\!\begin{aligned}[t] 
 &\Gamma_{i_t} = \sigma(x_t.W_{xi} + h_{t-1}.W_{hi} + \Gamma_{i_{(t-1)}}.W_{ii} \\ 
 &+ \Gamma_{f_{(t-1)}}.W_{fi} + \Gamma_{o_{(t-1)}}.W_{oi} + b_i) 
 \end{aligned}$ \\
 
 &  & 
 $ \!\begin{aligned}[t] 
 &\Gamma_{f_t} = \sigma(x_t.W_{xf} + h_{t-1}.W_{hf} + \Gamma_{i_{(t-1)}}.W_{if} \\ 
 &+ \Gamma_{f_{(t-1)}}.W_{ff} + \Gamma_{o_{(t-1)}}.W_{of} + b_f) 
 \end{aligned}$ \\ 
 
 &  & 
 $ \!\begin{aligned}[t] 
 &\Gamma_{o_t} = \sigma(x_t.W_{xo} + h_{t-1}.W_{ho} + \Gamma_{i_{(t-1)}}.W_{io} \\  
 &+ \Gamma_{f_{(t-1)}}.W_{fo} + \Gamma_{o_{(t-1)}}.W_{oo} + b_o) 
 \end{aligned}$ \\
 
 &  & $ c_t =\Gamma_{f_t} \otimes c_{t-1}+\Gamma_{i_t} \otimes \tilde{c_t} $ \\
 &  & $ h_t = \Gamma_{o_t} \otimes \tanh(c_t)$\\
\hline

\end{longtable}

\pagebreak

\begin{longtable}[H]{c >{\centering\arraybackslash} m{6cm} c}
\caption{Description of different architectures of RNN cells.}
\label{tab2}
\\\hline
\multicolumn{1}{c}{Cell name} & \multicolumn{1}{c}{Cell architecture} & \multicolumn{1}{c}{Cell computations} \\ \hline\hline

\endfirsthead
\caption{ -- \textit{Continued from previous page}} \\\hline
\multicolumn{1}{c}{Cell name} & \multicolumn{1}{c}{Cell architecture} & \multicolumn{1}{c}{Cell computations} \\ \hline\hline
\endhead
\hline \\
\endfoot
\hline
\endlastfoot
    
\multirow{3}{*}{LSTM-SLIM1} & \multirow{3}{*}{\includegraphics[height=3.8cm,width=5cm]{LSTM-SLIM1.jpg}} 
    & $ \tilde{c_t} =  \tanh(x_t.W_{x\tilde{c}} + h_{t-1}.W_{h\tilde{c}} + b_{\tilde{c}}) $ \\  
 &  & $ \Gamma_{i_t} = \sigma(h_{t-1}.W_{hi} + b_i) $ \\
 &  & $ \Gamma_{f_t} = \sigma(h_{t-1}.W_{hf} + b_f) $ \\ 
 &  & $ \Gamma_{o_t} = \sigma(h_{t-1}.W_{ho} + b_o) $ \\
 &  & $ c_t =\Gamma_{f_t} \otimes c_{t-1}+\Gamma_{i_t} \otimes \tilde{c_t} $ \\
 &  & $ h_t = \Gamma_{o_t} \otimes \tanh(c_t) $
\\\hline

\multirow{3}{*}{LSTM-SLIM2} & \multirow{3}{*}{\includegraphics[height=3.8cm,width=5cm]{LSTM-SLIM2.jpg}} 
    & $ \tilde{c_t} =  \tanh(x_t.W_{x\tilde{c}} + h_{t-1}.W_{h\tilde{c}} + b_{\tilde{c}}) $ \\  
 &  & $ \Gamma_{i_t} = \sigma(h_{t-1}.W_{hi}) $ \\
 &  & $ \Gamma_{f_t} = \sigma(h_{t-1}.W_{hf}) $ \\ 
 &  & $ \Gamma_{o_t} = \sigma(h_{t-1}.W_{ho}) $ \\
 &  & $ c_t =\Gamma_{f_t} \otimes c_{t-1}+\Gamma_{i_t} \otimes \tilde{c_t} $ \\
 &  & $ h_t = \Gamma_{o_t} \otimes \tanh(c_t) $
\\\hline

\multirow{3}{*}{LSTM-SLIM3} & \multirow{3}{*}{\includegraphics[height=3.8cm,width=5cm]{LSTM-SLIM3.jpg}} 
    & $ \tilde{c_t} =  \tanh(x_t.W_{x\tilde{c}} + h_{t-1}.W_{h\tilde{c}} + b_{\tilde{c}}) $ \\  
 &  & $ \Gamma_{i_t} = \sigma(b_i) $ \\
 &  & $ \Gamma_{f_t} = \sigma(b_f) $ \\ 
 &  & $ \Gamma_{o_t} = \sigma(b_o) $ \\
 &  & $ c_t =\Gamma_{f_t} \otimes c_{t-1}+\Gamma_{i_t} \otimes \tilde{c_t} $ \\
 &  & $ h_t = \Gamma_{o_t} \otimes \tanh(c_t) $
\\\hline

\multirow{3}{*}{GRU} & \multirow{3}{*}{\includegraphics[height=3.8cm,width=5cm]{GRU.jpg}} 
   &  \\ 
 &  & $ \Gamma_{u_t} = \sigma(x_t.W_{xu} + h_{t-1}.W_{hu} + b_u) $ \\
 &  & $ \Gamma_{r_t} = \sigma(x_t.W_{xr} + h_{t-1}.W_{hr} + b_r) $ \\ 
 &  & $ \tilde{h_t} =  \tanh(x_t.W_{x\tilde{h}} + (\Gamma_{r_t} \otimes h_{t-1}).W_{h\tilde{h}} + b_{\tilde{h}}) $ \\  
 &  & $ h_t =\Gamma_{u_t} \otimes \tilde{h_t} + (1-\Gamma_{u_t}) \otimes h_{t-1} $ \\
  &  &
\\\hline

\multirow{3}{*}{MUT1} & \multirow{3}{*}{\includegraphics[height=3.8cm,width=5cm]{MUT1.jpg}}
   &  \\ 
 &  & $ \Gamma_{u_t} = \sigma(x_t.W_{xu} + b_u) $ \\
 &  & $ \Gamma_{r_t} = \sigma(x_t.W_{xr} + h_{t-1}.W_{hr} + b_r) $ \\ 
 &  & $ \tilde{h_t} =  \tanh(\tanh(x_t) + (\Gamma_{r_t} \otimes h_{t-1}).W_{h\tilde{h}} + b_{\tilde{h}}) $ \\  
 &  & $ h_t =\Gamma_{u_t} \otimes \tilde{h_t} + (1-\Gamma_{u_t}) \otimes h_{t-1} $ \\
  &  &
\\\hline
\pagebreak
\multirow{3}{*}{MUT2} & \multirow{3}{*}{\includegraphics[height=3.8cm,width=5cm]{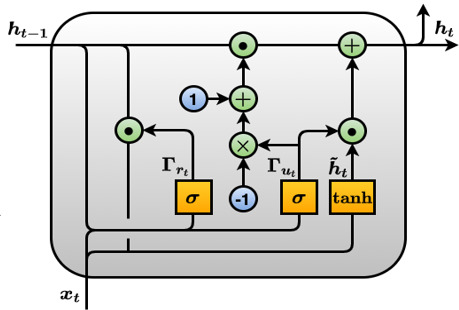}} 
   &  \\ 
 &  & $ \Gamma_{u_t} = \sigma(x_t.W_{xu} + h_{t-1}.W_{hu} + b_u) $ \\
 &  & $ \Gamma_{r_t} = \sigma(x_t + h_{t-1}.W_{hr} + b_r) $ \\ 
 &  & $ \tilde{h_t} =  \tanh(x_t.W_{x\tilde{h}} + (\Gamma_{r_t} \otimes h_{t-1}).W_{h\tilde{h}} + b_{\tilde{h}}) $ \\  
 &  & $ h_t =\Gamma_{u_t} \otimes \tilde{h_t} + (1-\Gamma_{u_t}) \otimes h_{t-1} $ \\
 &  & \\
\\\hline

\multirow{3}{*}{MUT3} & \multirow{3}{*}{\includegraphics[height=3.8cm,width=5cm]{MUT3.jpg}} 
   &  \\ 
 &  & $ \Gamma_{u_t} = \sigma(x_t.W_{xu} + \tanh(h_{t-1}).W_{hu} + b_u) $ \\
 &  & $ \Gamma_{r_t} = \sigma(x_t.W_{xr} + h_{t-1}.W_{hr} + b_r) $ \\ 
 &  & $ \tilde{h_t} =  \tanh(x_t.W_{x\tilde{h}} + (\Gamma_{r_t} \otimes h_{t-1}).W_{h\tilde{h}} + b_{\tilde{h}}) $ \\  
 &  & $ h_t =\Gamma_{u_t} \otimes \tilde{h_t} + (1-\Gamma_{u_t}) \otimes h_{t-1} $ \\
  &  &
\\\hline

\multirow{3}{*}{GRU-SLIM1} & \multirow{3}{*}{\includegraphics[height=3.8cm,width=5cm]{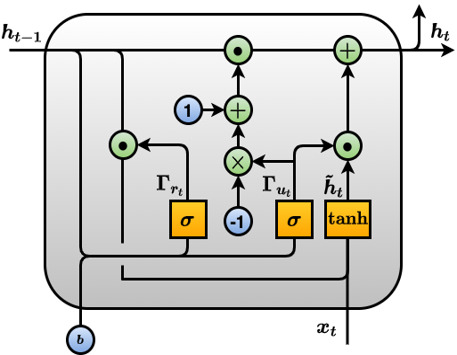}} 
   &  \\ 
 &  & $ \Gamma_{u_t} = \sigma(h_{t-1}.W_{hu} + b_u) $ \\
 &  & $ \Gamma_{r_t} = \sigma(h_{t-1}.W_{hr} + b_r) $ \\ 
 &  & $ \tilde{h_t} =  \tanh(x_t.W_{x\tilde{h}} + (\Gamma_{r_t} \otimes h_{t-1}).W_{h\tilde{h}} + b_{\tilde{h}}) $ \\  
 &  & $ h_t =\Gamma_{u_t} \otimes \tilde{h_t} + (1-\Gamma_{u_t}) \otimes h_{t-1} $ \\
  &  &
\\\hline

\multirow{3}{*}{GRU-SLIM2} & \multirow{3}{*}{\includegraphics[height=3.8cm,width=5cm]{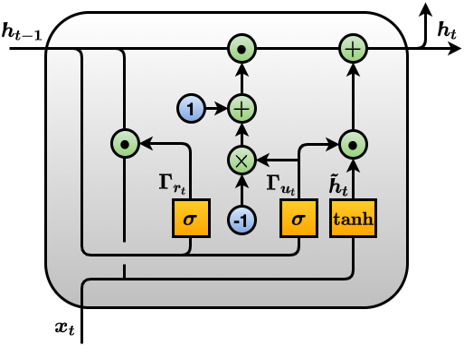}} 
   &  \\ 
 &  & $ \Gamma_{u_t} = \sigma(h_{t-1}.W_{hu}) $ \\
 &  & $ \Gamma_{r_t} = \sigma(h_{t-1}.W_{hr}) $ \\ 
 &  & $ \tilde{h_t} =  \tanh(x_t.W_{x\tilde{h}} + (\Gamma_{r_t} \otimes h_{t-1}).W_{h\tilde{h}} + b_{\tilde{h}}) $ \\  
 &  & $ h_t =\Gamma_{u_t} \otimes \tilde{h_t} + (1-\Gamma_{u_t}) \otimes h_{t-1} $ \\
  &  &
\\\hline

\multirow{3}{*}{GRU-SLIM3} & \multirow{3}{*}{\includegraphics[height=3.8cm,width=5cm]{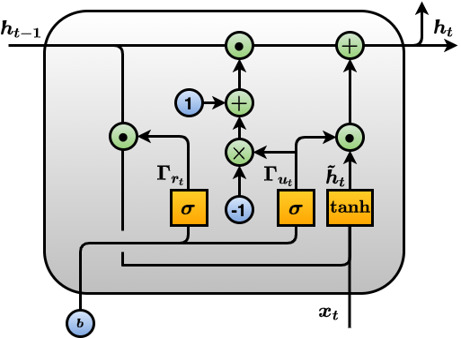}} 
   &  \\ 
 &  & $ \Gamma_{u_t} = \sigma(b_u) $ \\
 &  & $ \Gamma_{r_t} = \sigma(b_r) $ \\ 
 &  & $ \tilde{h_t} =  \tanh(x_t.W_{x\tilde{h}} + (\Gamma_{r_t} \otimes h_{t-1}).W_{h\tilde{h}} + b_{\tilde{h}}) $ \\  
 &  & $ h_t =\Gamma_{u_t} \otimes \tilde{h_t} + (1-\Gamma_{u_t}) \otimes h_{t-1} $ \\
  &  &
\\\hline

\pagebreak

\multirow{3}{*}{MGU} & \multirow{3}{*}{\includegraphics[height=3.8cm,width=5cm]{MGU.jpg}} 
   &  \\ 
 &  & $ \Gamma_{f_t} = \sigma(x_t.W_{xf} + h_{t-1}.W_{hf} + b_f) $ \\
 &  & $ \tilde{h_t} =  \tanh(x_t.W_{x\tilde{h}} + (\Gamma_{f_t} \otimes h_{t-1}).W_{h\tilde{h}} + b_{\tilde{h}}) $ \\  
 &  & $ h_t =\Gamma_{f_t} \otimes \tilde{h_t} + (1-\Gamma_{f_t}) \otimes h_{t-1} $ \\
 &  & \\
  &  &
\\\hline

\multirow{3}{*}{MGU-SLIM1} & \multirow{3}{*}{\includegraphics[height=3.8cm,width=5cm]{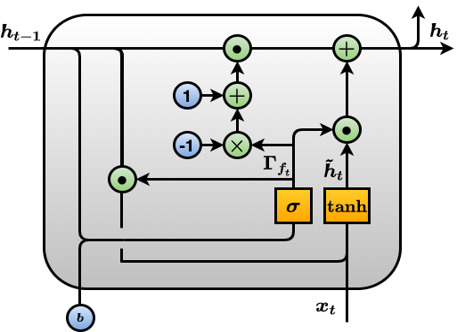}} 
   &  \\ 
 &  & $ \Gamma_{f_t} = \sigma(h_{t-1}.W_{hf} + b_f) $ \\
 &  & $ \tilde{h_t} =  \tanh(x_t.W_{x\tilde{h}} + (\Gamma_{f_t} \otimes h_{t-1}).W_{h\tilde{h}} + b_{\tilde{h}}) $ \\  
 &  & $ h_t =\Gamma_{f_t} \otimes \tilde{h_t} + (1-\Gamma_{f_t}) \otimes h_{t-1} $ \\
 &  & \\
  &  &
\\\hline

\multirow{3}{*}{MGU-SLIM2} & \multirow{3}{*}{\includegraphics[height=3.8cm,width=5cm]{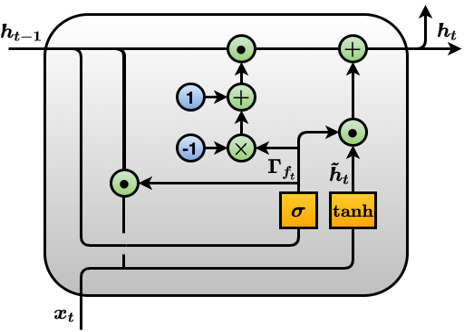}}
   &  \\ 
 &  & $ \Gamma_{f_t} = \sigma(h_{t-1}.W_{hf}) $ \\
 &  & $ \tilde{h_t} =  \tanh(x_t.W_{x\tilde{h}} + (\Gamma_{f_t} \otimes h_{t-1}).W_{h\tilde{h}} + b_{\tilde{h}}) $ \\  
 &  & $ h_t =\Gamma_{f_t} \otimes \tilde{h_t} + (1-\Gamma_{f_t}) \otimes h_{t-1} $ \\
 &  & \\
 &  &
\\\hline

\multirow{3}{*}{MGU-SLIM3} & \multirow{3}{*}{\includegraphics[height=3.8cm,width=5cm]{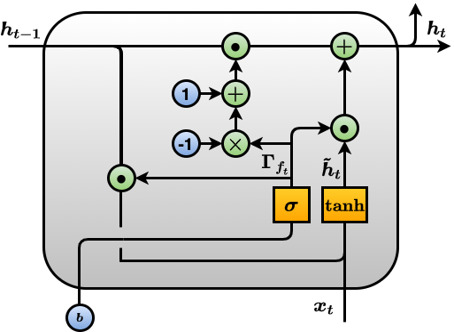}}
    &  \\ 
 &  & $ \Gamma_{f_t} = \sigma(b_f) $ \\
 &  & $ \tilde{h_t} =  \tanh(x_t.W_{x\tilde{h}} + (\Gamma_{f_t} \otimes h_{t-1}).W_{h\tilde{h}} + b_{\tilde{h}}) $ \\  
 &  & $ h_t =\Gamma_{f_t} \otimes \tilde{h_t} + (1-\Gamma_{f_t}) \otimes h_{t-1} $ \\
 &  & \\
 &  &
\\\hline

\multirow{3}{*}{SCRN} & \multirow{3}{*}{\includegraphics[height=3.8cm,width=4.8cm]{SCRN.jpg}} 
   &  \\ 
  &  & \\ 
  &  &  $ s_t = (1-\alpha)x_t.W_{xs} + \alpha.s_{t-1} $\\  
  &  & $\alpha \in [0, 1]$ \\ 
  &  & $ h_t = \tanh(x_t.W_{xh} + h_{t-1}.W_{hh} + s_{t-1}.W_{sh} + b_{h}) $ \\
 &  & 
\\\hline

\pagebreak

\multirow{3}{*}{MRNN} & \multirow{3}{*}{\includegraphics[height=3.8cm,width=4.8cm]{MRNN.jpg}} 
   &  \\ 
 &  & \\
 &  & $ h_t =  \tanh(x_t.W_{xh} + h_{t-1}.W_{hh} + \hat{y}_{t-1}.W_{yh} + b_{h}) $ \\  
 &  & \\
 &  &  \\
 &  & 
\\\hline

\multirow{3}{*}{JORDAN} & \multirow{3}{*}{\includegraphics[height=3.8cm,width=4.5cm]{JORDAN.jpg}} 
   &  \\ 
 &  & \\
 &  & $ h_t =  \tanh(x_t.W_{xh} + \hat{y}_{t-1}.W_{yh} + b_{h}) $ \\
 &  & \\
 &  &  \\
 &  & 
\\\hline

\multirow{3}{*}{IRNN} & \multirow{3}{*}{\includegraphics[height=3.8cm,width=4.8cm]{IRNN.jpg}} 
   &  \\ 
 &  & \\
 &  &  $ h_t =  ReLu(x_t.W_{xh} + h_{t-1}.W_{hh} + b_{h}) $ \\
 &  & \\
 &  &  \\
 &  & 
\\\hline

\multirow{3}{*}{ELMAN} & \multirow{3}{*}{\includegraphics[height=3.8cm,width=4.8cm]{ELMAN.jpg}} 
   &  \\ 
 &  & \\
 &  &  \\ 
 &  & $ h_t =  \tanh(x_t.W_{xh} + h_{t-1}.W_{hh} + b_{h}) $ \\
 &  &  \\
 &  & 
\\\hline

\end{longtable}

\end{document}